\documentclass[10pt,a4paper]{article}
\usepackage{wechat-techreport}
\usepackage{listings}
\usepackage{xcolor}
\usepackage{tcolorbox}
\usepackage{amsmath}    % \operatorname
\usepackage{booktabs}   % \toprule, \midrule, \bottomrule, \cmidrule
\usepackage{multirow}   % \multirow
\usepackage{graphicx}   % \resizebox, \includegraphics
\usepackage{tabularx}   % tabularx
\usepackage{array}
\usepackage{xcolor}
\usepackage{colortbl}   % \rowcolor

\definecolor{codebg}{RGB}{248,248,248}
\definecolor{codecomment}{RGB}{100,100,100}

\definecolor{WeChatGreen}{RGB}{3,193,96}
\colorlet{OursRowColor}{WeChatGreen!12}

\definecolor{BestCellColor}{HTML}{1FAA63}
\definecolor{SecondCellColor}{HTML}{6ABA85}
\definecolor{ThirdCellColor}{HTML}{CDEFD7}

\colorlet{BestCellColorFade}{BestCellColor!60}
\colorlet{SecondCellColorFade}{SecondCellColor!60}
\colorlet{ThirdCellColorFade}{ThirdCellColor!70}

\newcommand{\hlbest}[1]{\setlength{\fboxsep}{2pt}\colorbox{BestCellColorFade}{$\vphantom{0}\raisebox{-0.18ex}{#1}$}}
\newcommand{\hlsecond}[1]{\setlength{\fboxsep}{2pt}\colorbox{SecondCellColorFade}{$\vphantom{0}\raisebox{-0.18ex}{#1}$}}
\newcommand{\hlthird}[1]{\setlength{\fboxsep}{2pt}\colorbox{ThirdCellColorFade}{$\vphantom{0}\raisebox{-0.18ex}{#1}$}}

\lstdefinelanguage{json}{
    basicstyle=\ttfamily\small,
    numbers=none,
    breaklines=true,
    frame=none,
    backgroundcolor=\color{codebg},
    showstringspaces=false,
    morestring=[b]",
    stringstyle=\color{WeChatInk},
    literate=
     *{:}{{{\color{gray}:}}}{1}
      {,}{{{\color{gray},}}}{1}
}

\newtcblisting{jsonbox}{
    listing only,
    listing options={
        language=json,
        basicstyle=\ttfamily\scriptsize,
        breaklines=true,
        breakatwhitespace=false,
        columns=fullflexible,
        keepspaces=true,
        showstringspaces=false
    },
    colback=codebg,
    colframe=black!15,
    boxrule=0.5pt,
    arc=2pt,
    left=5pt,
    right=5pt,
    top=5pt,
    bottom=5pt
}

\newcommand{\ModelName}{WeLM-Image}
\newcommand{\ModelNameTurbo}{WeLM-Image-Turbo}

\newcommand{\ModelNameTurboeight}{WeLM-Image-Turbo-8step}

\newcommand{\UNDPath}{UND path}
\newcommand{\GENPath}{GEN path}

\newcommand{\ReportTitle}{\ModelName: A Unified Image Generation Model towards Visual Intelligence}
\newcommand{\ReportOrganization}{WeChat}
\newcommand{\ReportAuthors}{WeChat Vision, Tencent Inc.}
\newcommand{\ReportVersion}{Technical Report}

\newcommand{\TeaserAsset}{assets/teaser.pdf}

\hypersetup{
  pdftitle={\ReportTitle},
  pdfauthor={\ReportAuthors},
  pdfsubject={Unified image foundation model technical report},
  pdfkeywords={image generation, image editing, multimodal understanding, MoT, MoE}
}

\begin{document}

\pagenumbering{roman}
\thispagestyle{empty}

% -----------------------------------------------------------------------------
% Front-page visual hierarchy
%   1. WeChat is the publisher / brand mark -> small, quiet, top-left.
%   2. WeGen-Image is the product -> dominant hero mark.
%   3. Paper title and authors sit directly below the product mark.
%   4. Abstract is visually grouped, so the teaser can start higher on the page.
%
% Both PNGs contain large transparent / white margins. The trim values below
% preserve a little safety area for the soft glow while removing most dead space.
% -----------------------------------------------------------------------------

\vspace*{-10mm}

% --- Publisher / metadata bar -------------------------------------------------
\noindent
\begin{minipage}[c]{0.36\textwidth}
  \raggedright
  \IfFileExists{assets/wechat-logo.png}{%
    \includegraphics[
      height=6.5mm,
      trim=105 105 107 105,
      clip
    ]{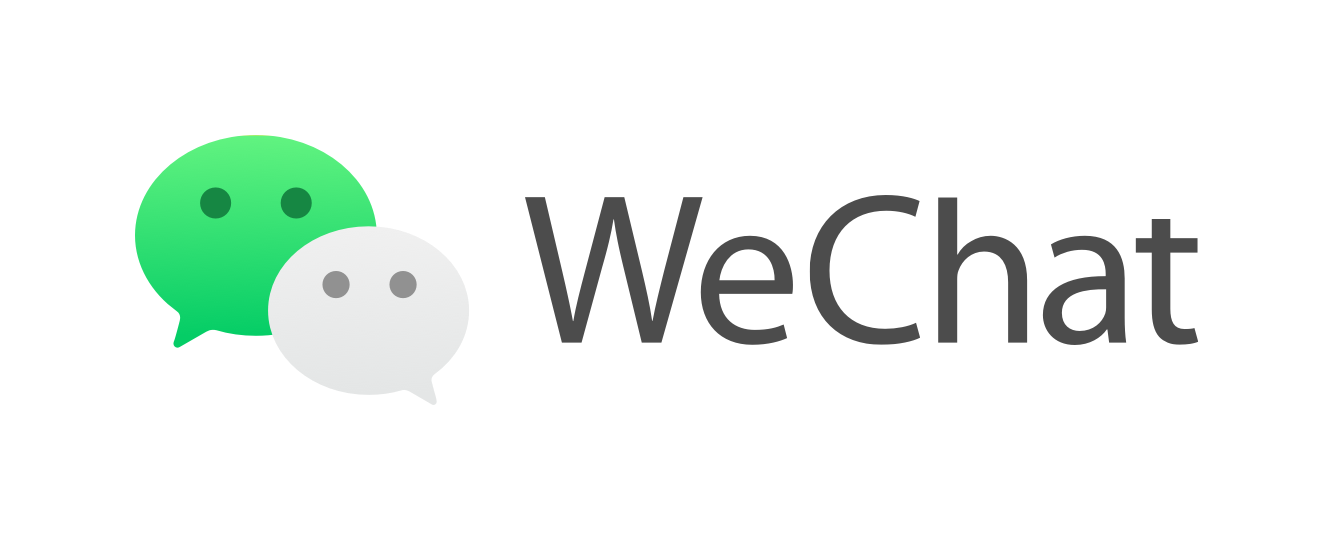}%
  }{}%
\end{minipage}%
\hfill
\begin{minipage}[c]{0.60\textwidth}
  \raggedleft
  {\scriptsize\bfseries\color{WeChatInk}TECHNICAL REPORT}\par
\end{minipage}

\vspace{1.8mm}
\noindent{\color{WeChatGreen}\rule{\textwidth}{1.2pt}\par}

% --- Product hero -------------------------------------------------------------
% Large hero retained; vertical space is recovered from implicit center/float spacing and dead image margins.
\vspace{2mm}
\begingroup
  \centering
  \IfFileExists{assets/welmimage.png}{%
    % The source PNG has generous white margins. Crop only the dead space
    % (while preserving the soft glow), then give the product mark enough
    % visual weight to balance the title below.
    \includegraphics[
      width=0.9\textwidth,
      trim=40 140 32 140,
      clip
    ]{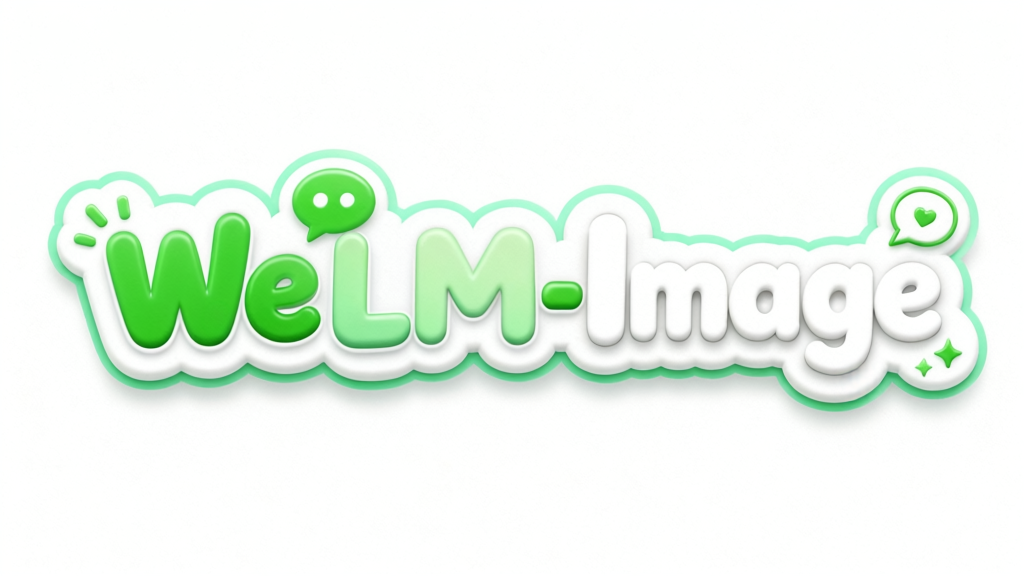}\par
  }{}%

  \vspace{3mm}
  \begin{minipage}{0.92\textwidth}
    \centering
    {\fontsize{22}{25.5}\selectfont\bfseries\color{WeChatInk}
      \ReportTitle\par}
  \end{minipage}

  \vspace{3mm}
  {\small\color{WeChatGray}\ReportAuthors\par}
  \vspace{3mm}
\endgroup

\vspace{2mm}

% --- Abstract card ------------------------------------------------------------
% A soft container gives the text a clear role without competing with the logo.
\begin{tcolorbox}[
  colback=WeChatGreen!3,
  colframe=WeChatGreen!38,
  boxrule=.55pt,
  arc=1.6mm,
  left=4.2mm,
  right=4.2mm,
  top=2.2mm,
  bottom=2.2mm,
  before skip=0pt,
  after skip=0pt,
  fontupper=\normalsize
]

  \noindent\makebox[\linewidth][c]{\textbf{\color{WeChatGreen}Abstract}}\par
  \vspace{2mm}
  \color{WeChatInk}

We present \textbf{\ModelName{}}, a unified image foundation model
that integrates multimodal understanding into text-to-image generation,
instruction-guided editing, and multi-reference synthesis. Designed for
single-GPU deployment in WeChat's large-scale consumer applications, the model
processes text, reference images, and target latents as one interleaved sequence.
A Mixture-of-Transformers (MoT) backbone combines shared attention with
independent understanding and generation parameters to limit optimization
interference. Equipping each path with Mixture-of-Experts (MoE) modules
provides sparse capacity, with 31.7B stored parameters and approximately 5.4B
generation parameters activated per target latent token. A generation-oriented
data engine combines a knowledge-grounded visual vocabulary with large-scale
visual clustering to balance semantic coverage and editing tasks, while
training-time evaluation guides iterative data refinement. Bilingual
annotations and verified editing supervision support progressive training up
to native 1.5K resolution. We optimize generation and editing separately with
rubric- and checklist-based rewards, then consolidate the specialized policies
through noise-aware on-policy distillation. 
In the reported overall comparisons, the 1536-resolution model ranks first among the
evaluated open-source models on PRISM-Bench, Qwen-Image-Bench, and UnicBench.
MeanFlow-based few-step distillation yields \ModelNameTurbo{}, which generates
a $1024\times1024$ image in 4.98\,s on a single NVIDIA H20 GPU using eight
guidance-free steps.

% \par\vspace{0.8mm}
% {\scriptsize
% \textbf{GitHub: }\hspace{2.2em}
% \textbf{HuggingFace Model:}}
% \par}

\end{tcolorbox}

\vspace{10mm}

% --- figure for performance --- added by yinggzhang ---- %
{
\setlength{\intextsep}{3pt}
\setlength{\textfloatsep}{3pt}
\begin{figure}[!ht]
  \centering
  % Keep the benchmark plots close to their original size for readability.
  % Most of the vertical saving is taken from the hero/header area instead.
  \begin{minipage}[c]{0.35\linewidth}
    \centering
    \includegraphics[width=\linewidth,trim=0 0 0 0,clip]{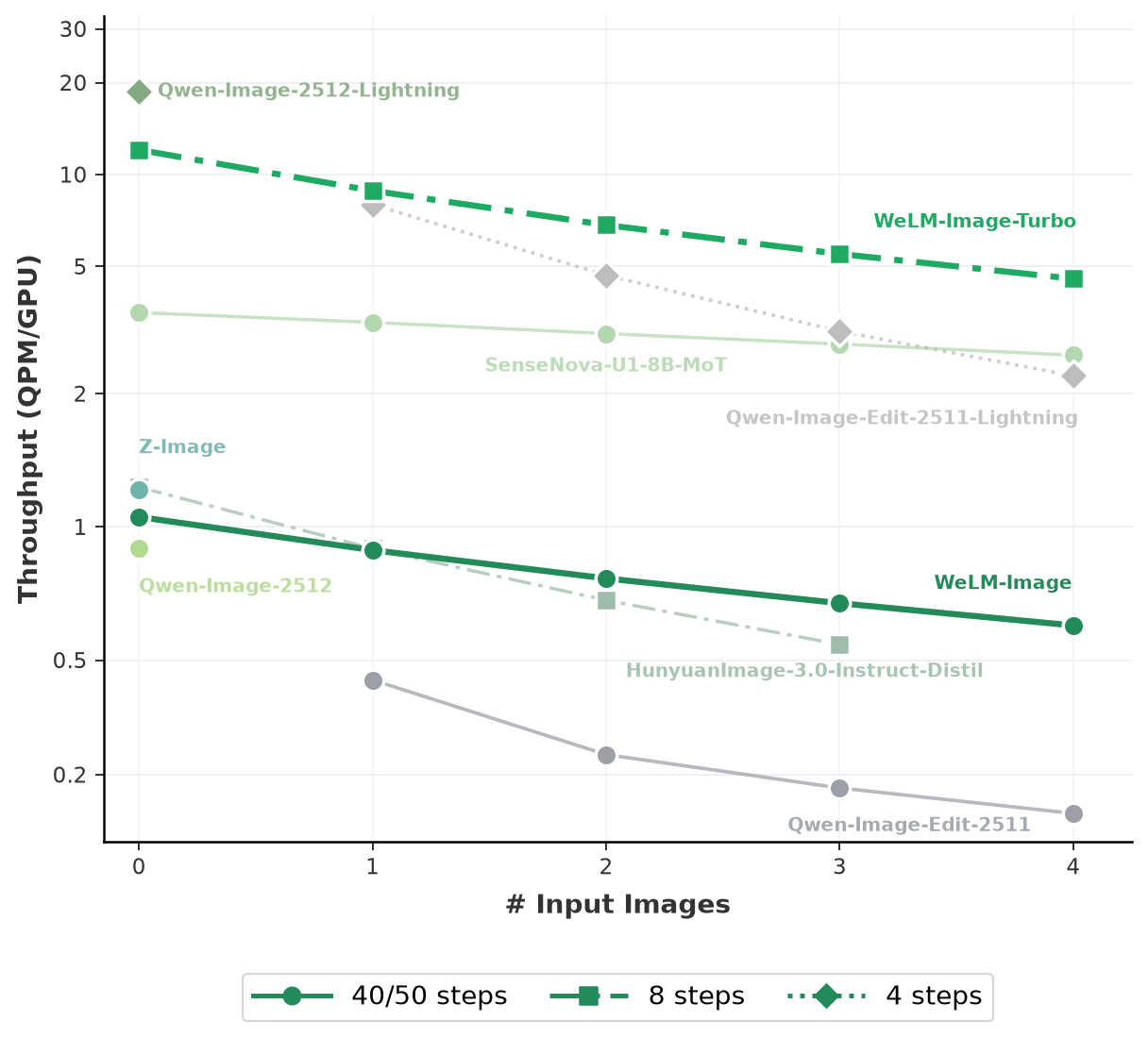}
  \end{minipage}%
  \hspace{0.005\linewidth}
  \begin{minipage}[c]{0.30\linewidth}
    \centering
    \includegraphics[width=\linewidth,trim=0 0 0 0,clip]{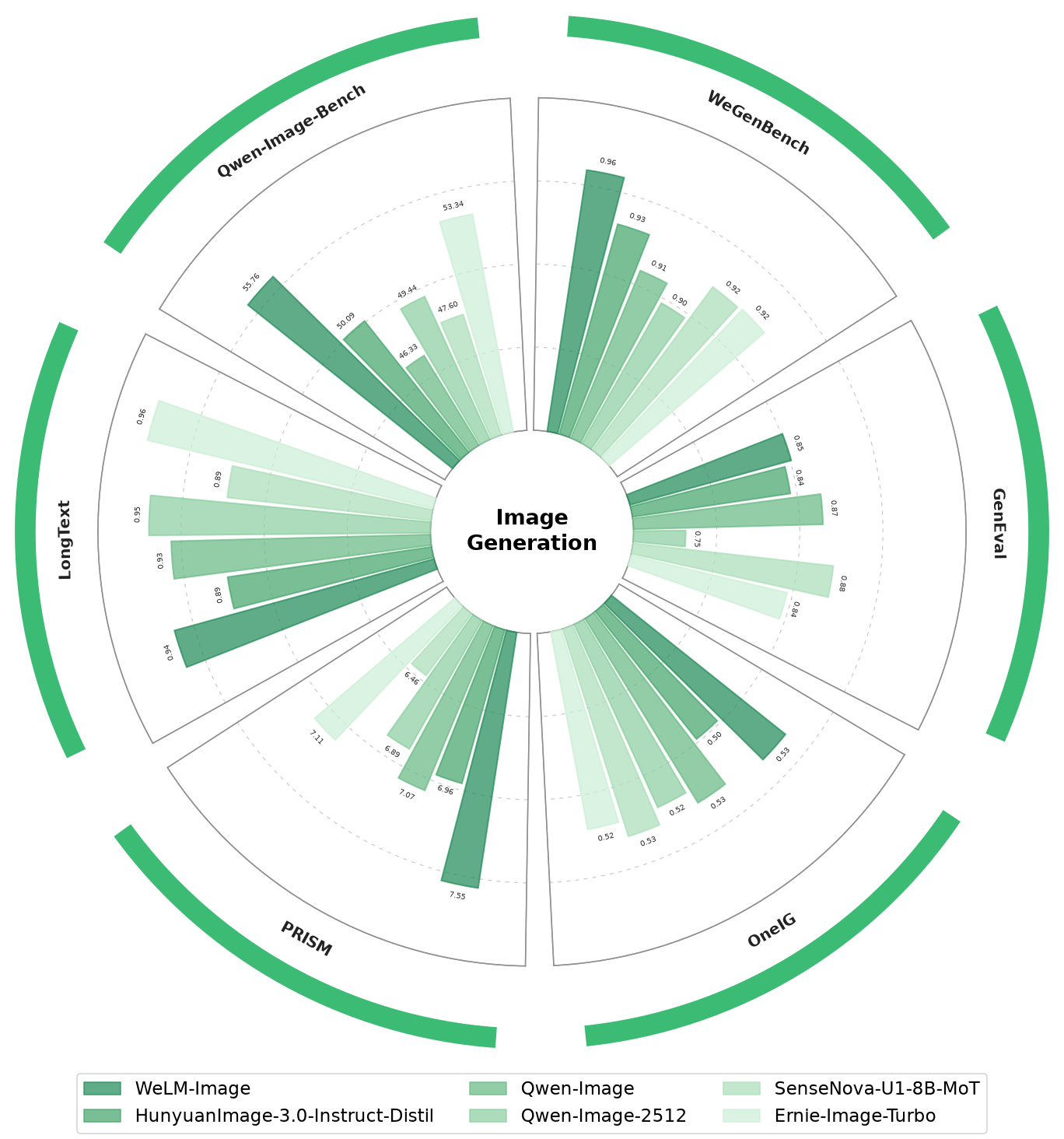}
  \end{minipage}%
  \hspace{0.005\linewidth}
  \begin{minipage}[c]{0.30\linewidth}
    \centering
    \includegraphics[width=\linewidth,trim=0 0 0 0,clip]{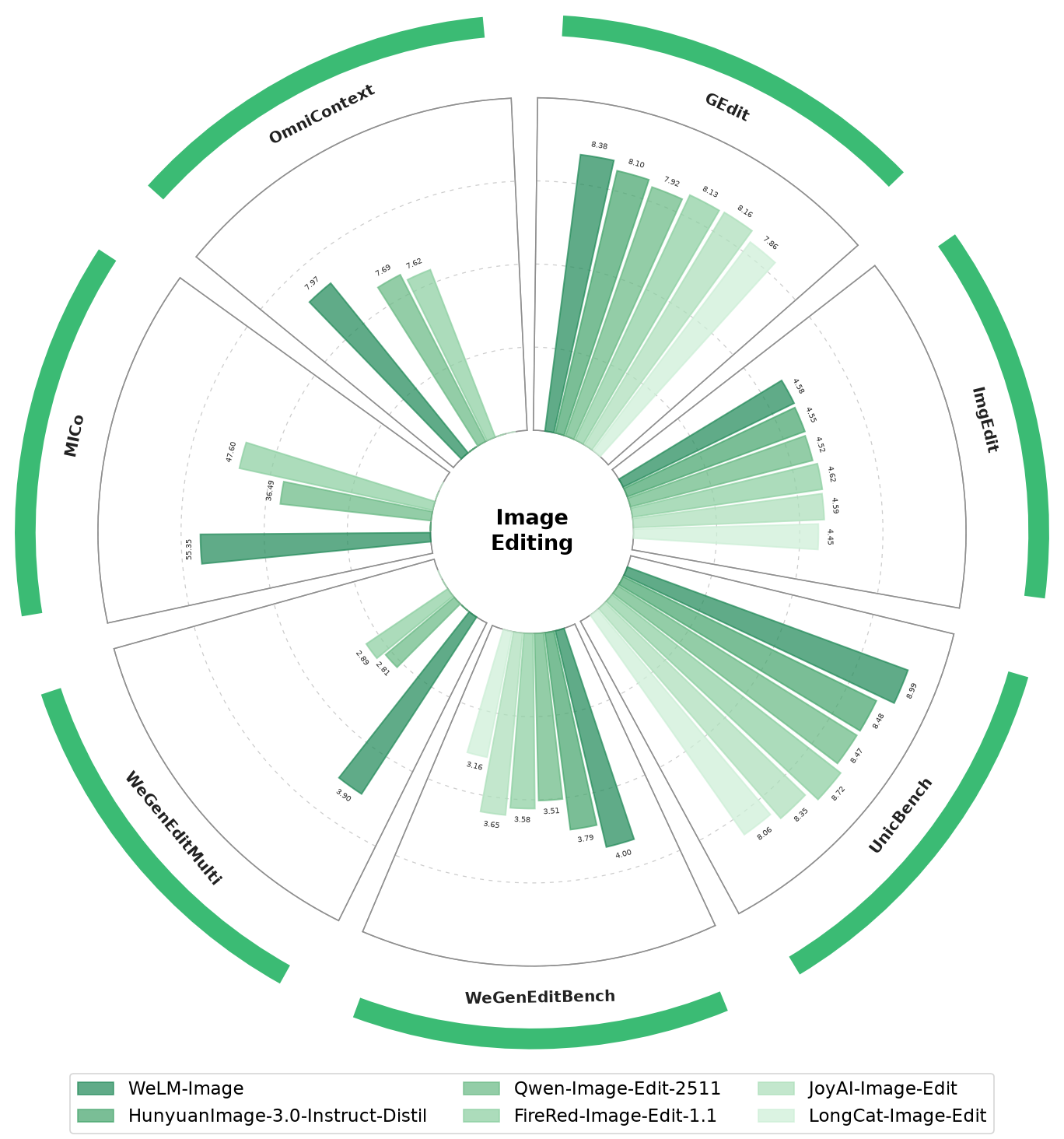}
  \end{minipage}
  \setlength{\abovecaptionskip}{2pt}
  \setlength{\belowcaptionskip}{0pt}
  \vspace{10mm}
  \caption{\small Performance and efficiency overview of \textbf{\ModelName{}}. \textbf{Left:} Throughput (QPM/GPU) compared with representative baselines as the number of input images increases. \textbf{Middle:} Aggregate performance on a suite of image-generation benchmarks. \textbf{Right:} Aggregate performance on a suite of image-editing benchmarks.}
  \label{fig:overall_performance_efficiency}
\end{figure}
}

% \begin{figure}
%   \centering  \includegraphics[width=0.8\linewidth]{assets/gen_hist.png}
%   \centering  \includegraphics[width=0.8\linewidth]{assets/edit_hist.png}
%   \caption{Performance and efficiency overview of \textbf{\ModelName{}}. \textbf{Left:}}
%   \label{fig:overallperformace}
% \end{figure}

% --- Teaser / system overview -------------------------------------------------
% Keep the teaser broad: it is the first content image readers should inspect.

\begin{comment}
\IfFileExists{\TeaserAsset}{
  \begin{center}
    \includegraphics[
      width=\textwidth,
      height=88mm,
      keepaspectratio
    ]{\TeaserAsset}
  \end{center}
}{
  \begin{tcolorbox}[
    height=88mm,
    colback=WeChatGreen!2,
    colframe=WeChatLine,
    boxrule=.65pt,
    arc=1.5mm,
    valign=center
  ]
    %\centering
    {\small\bfseries\color{WeChatGray}TEASER / SYSTEM OVERVIEW}\par
    \vspace{1mm}
    {\footnotesize\color{WeChatGray}Set \texttt{\string\TeaserAsset} in metadata.tex}
  \end{tcolorbox}
}
\vfill

\end{comment}

\clearpage
\thispagestyle{empty}

% Capability showcase placed between the front page and the table of contents.
% The two teasers share a 16:9 media box, but the editing board has a wide
% white margin while the generation collage runs nearly to the crop. Trim both
% to the measured ink so they occupy the same visual width. The height cap is
% only a guard that keeps both panels and their captions on a single page.
% Captions are set by hand rather than with \caption so that this front-matter
% page does not consume figure numbers from the body sections.
\vfill

\begin{center}
  \includegraphics[
    width=\textwidth,
    height=.42\textheight,
    keepaspectratio,
    trim=0 0 14.5 0,
    clip
  ]{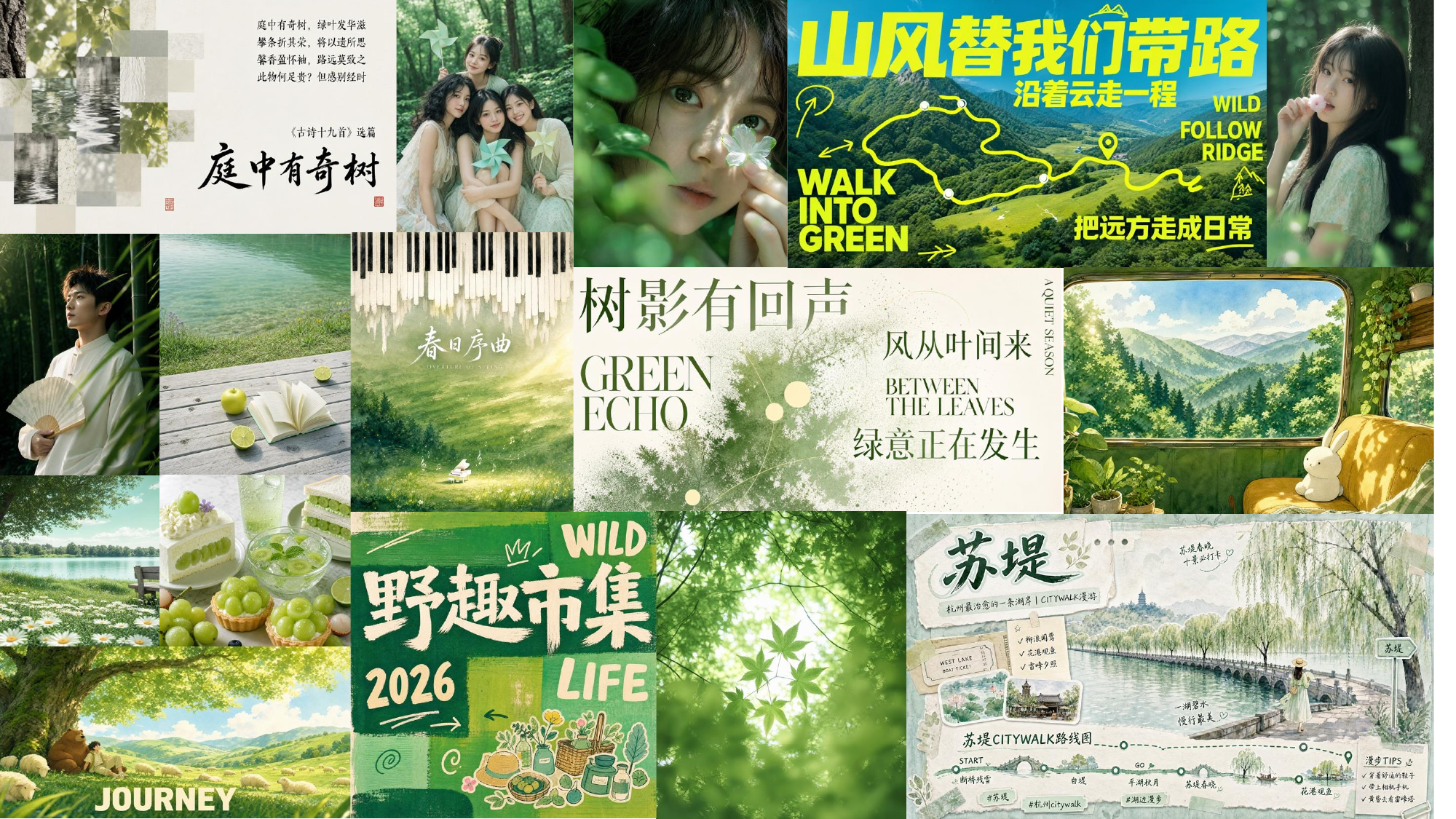}
\end{center}
\vspace{3mm}
\begin{center}
\begin{minipage}{0.95\textwidth}
  \small\color{WeChatInk}
  {\bfseries\color{WeChatGreen}Text-to-image generation.} Samples produced by \ModelName{} from text
  prompts alone, spanning photographic portraits and landscapes, illustration
  and paper-cut styles, and graphic-design layouts at several aspect ratios.
  Many of the examples set dense Chinese and English typography directly into
  posters, signage, and an annotated itinerary map.
\end{minipage}
\end{center}

\vfill

\begin{center}
  \includegraphics[
    width=\textwidth,
    height=.42\textheight,
    keepaspectratio,
    trim=18.5 28.5 21 23,
    clip
  ]{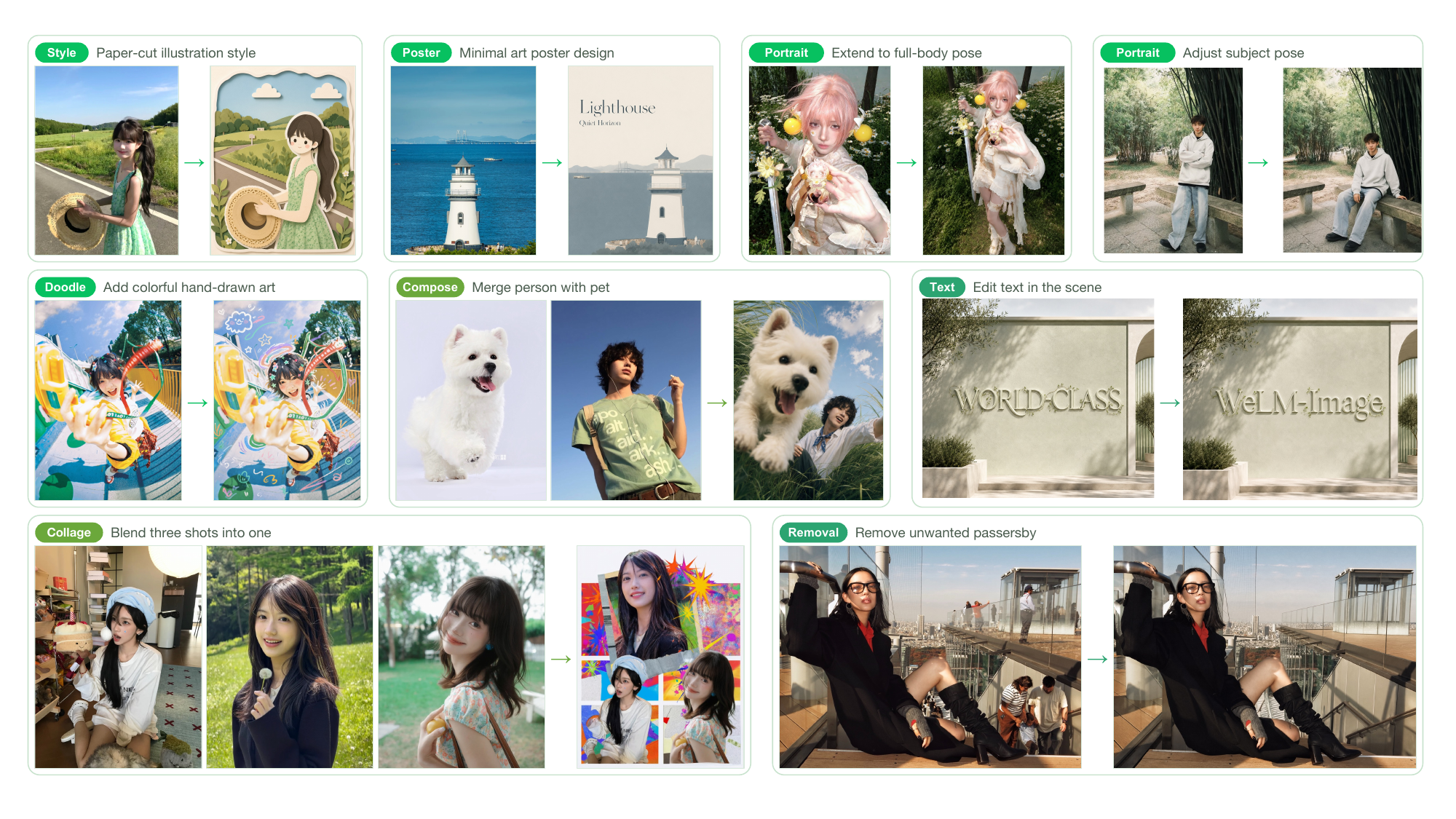}
\end{center}
\vspace{3mm}
\begin{center}
\begin{minipage}{0.95\textwidth}
  \small\color{WeChatInk}
  {\bfseries\color{WeChatGreen}Instruction-guided image editing.} Each block shows the input on the
  left and the result of a single natural-language instruction on the right,
  with the edit category tagged in green. The cases cover global restyling,
  pose and viewpoint change, local insertion and removal, and in-scene text
  replacement. The Compose and Collage panels additionally draw on two and
  three reference images, respectively.
\end{minipage}
\end{center}

\vfill

\clearpage
\tableofcontents
\clearpage

\pagenumbering{arabic}
\section{Introduction}
\label{sec:introduction}

Image generation and editing are important steps toward visual intelligence,
requiring models to translate language and visual context into coherent,
controllable images. Advances in diffusion and flow-based
transformers~\cite{peebles2023dit,lipman2023flowmatching} have substantially
improved image quality and prompt alignment. As the scope of these models
expands, their success increasingly depends on fine-grained control:
interpreting long Chinese and English prompts with simultaneous constraints
on objects, relations, counts, layout, and style~\cite{wei2025tiif};
rendering long Chinese and English text with accurate content and
layout~\cite{wu2025qwenimage};
executing complex edits while preserving subject identity~\cite{liu2025step1x};
and combining multiple references without confusing their attributes or
roles~\cite{xiao2025omnigen}. These
requirements motivate a common representation in which instructions,
reference images, and target latents interact throughout synthesis, allowing
multimodal understanding to support fine-grained visual control.

Recent image foundation models illustrate different ways to incorporate
such understanding into generation. Qwen-Image~\cite{wu2025qwenimage} uses
a multimodal language model to condition image synthesis and combines semantic and
reconstructive image features for consistent editing.
HunyuanImage~3.0~\cite{hunyuanimage2025} brings language modeling and image
generation into a shared multimodal backbone, while
SenseNova-U1~\cite{diao2026sensenova} further explores joint learning directly
from pixels and words. Autoregressive language
modeling and continuous image generation impose different optimization and
capacity requirements. A scalable unified model must accommodate these
differences while maintaining effective interaction between instructions,
reference images, and generated content.

For WeChat's consumer-facing services, the design must also account for
large request volumes and affordable serving costs.
We therefore pursue resource efficiency as a primary design objective,
treating single-GPU deployment and low per-request cost as constraints
on model scale and inference design from the outset. Our goal is to
maximize overall capability across generation, editing, multi-reference
composition, and text rendering within a practical parameter and serving
budget. This requires joint consideration of model capacity, the growing
context of multiple references, and the number of inference steps.

To address these modeling and deployment requirements, we present
\ModelName{}, a unified image foundation model that integrates multimodal
understanding into text-to-image generation, instruction-guided editing, and
multi-reference synthesis. All three tasks are expressed as interleaved text
and image sequences within one backbone. We adopt a dual-path
Mixture-of-Transformers (MoT)~\cite{liang2025mixtureoftransformers,deng2025bagel} design
to give understanding and generation independent parameters, and equip each
path with its own Mixture-of-Experts (MoE)
modules~\cite{shazeer2017moe,fedus2022switch}, forming a MoT--MoE
backbone. In this design, MoT provides specialization across the two
objectives to reduce optimization interference; MoE expands capacity within each path through
sparse expert activation. Despite this parameter separation, shared
attention connects both paths at every layer, allowing target latents to
draw jointly on textual instructions, semantic image features, and detailed
visual references. With greater
active capacity assigned to generation, the model scales to 31.7B stored
parameters while activating approximately 5.4B generation parameters per
target latent token.

We complement the backbone with \ModelName{}-VAE, trained through
reconstruction, semantic alignment, and text-oriented decoder fine-tuning to
support visual fidelity and text rendering. Building on rotary position
embeddings~\cite{su2024roformer} and their multimodal
extensions~\cite{wang2024qwen2vl}, our text-compatible unified
three-dimensional rotary positional embedding (3D RoPE) preserves the pretrained
language positional transformation while adding spatial structure to image
latents. During sampling, conditioning tokens are
computed once and cached, so each denoising step recomputes only the noisy
target latents (Section~\ref{sec:architecture}).

To develop these capabilities through training, we organize data
construction around semantic coverage, text accuracy, and
edit controllability. Starting from a curated pool of approximately 0.6B
images, a knowledge-grounded visual vocabulary makes named concepts
retrievable, while large-scale visual clustering organizes coherent appearance
patterns. Together, these structures guide semantic rebalancing and
task-balanced editing construction. Bilingual captions enriched with OCR
information describe visible text and layout; editing pairs are verified
against the actual source--target changes, and multi-reference samples bind
each input image to the attributes it contributes. Training-time evaluation
feeds back into filtering, sampling, task assignment, and mixture ratios to
address observed weaknesses. Progressive pretraining couples image-to-text
and text-to-image learning in its early stages, then concentrates on
generation, editing, and multi-reference tasks as resolution increases.
High-resolution continuation and supervised fine-tuning extend the model to
native 1.5K generation and complex visual instructions.

After supervised fine-tuning, we optimize generation and editing in separate
reinforcement-learning branches to address their distinct quality criteria.
Generation uses rubric-based prompt alignment~\cite{feng2025rubricrl} alongside
aesthetic, text-rendering, and visual-naturalness rewards. Editing uses
instruction-specific checklists to assess the requested change, content
preservation, and visual quality, with instruction following gating the reward.
Noise-aware on-policy distillation~\cite{zhou2026danceopd,li2026rethinking}
then consolidates the specialized policies
into one model: prompt-enhanced teacher predictions guide global semantics
and layout at high noise levels, while standard RL teacher predictions
supervise local refinement at low noise levels. Finally, MeanFlow-based
few-step distillation~\cite{geng2025meanflow} yields \ModelNameTurbo{},
which supports guidance-free generation and editing with eight sampling steps.

% On the training side, limited access to high-end compute makes the platform
% another practical constraint on model development. We therefore build on
% domestic accelerators to support the full model lifecycle. All stages of our
% model training, from pretraining through reinforcement learning and
% distillation, run on a cluster of more than 2{,}500 Cambricon MLU590
% accelerators. Token-budget packing~\cite{krell2021packing}, fully sharded data
% parallelism (FSDP)~\cite{zhao2023fsdp}, and
% context parallelism~\cite{jacobs2024ulysses} jointly
% support the 30B-class model and packed sequences of up to 64K tokens
% (Section~\ref{sec:training-infra}).

Across these stages, token-budget packing~\cite{krell2021packing}, fully
sharded data parallelism (FSDP)~\cite{zhao2023fsdp}, and context
parallelism~\cite{jacobs2024ulysses} jointly support the 30B-class model
and packed sequences of up to 64K tokens
(Section~\ref{sec:training-infra}).

To assess both model capability and deployment efficiency, we evaluate
\ModelName{} across image generation, bilingual text rendering,
image editing, multi-reference composition, and inference cost. In the
reported overall comparisons, the 1536-resolution model leads the evaluated open-source
models on PRISM-Bench~\cite{fang2026flux} and
Qwen-Image-Bench~\cite{li2026qwen}
(Table~\ref{tab:overall_results_summary}), as well as
UnicBench~\cite{ye2026unicedit} and
WeGenEditBench (Table~\ref{tab:alleditbench}). These results demonstrate
strong generation and editing capabilities within a single model.
Complementing these quality results, \ModelNameTurbo{} generates a
$1024\times1024$ image in 4.98\,s on one NVIDIA H20 GPU, an $11.4\times$
speedup over the 50-step base model
with classifier-free guidance~\cite{ho2022classifierfree}. On the same GPU,
Turbo completes editing
with one to four input images in 6.78--13.17\,s, producing a
$1024\times1024$ output in each case (Table~\ref{tab:latency}). Through this
report, we document the model design and the complete data, training, and
deployment methodology behind \ModelName{}, providing a practical basis
for further advances in unified visual intelligence.

% Building on this generation and editing foundation, we aim to extend
% \ModelName{} toward integrated image understanding and creation, including
% OCR, visual grounding, object detection, and segmentation. This direction
% serves our longer-term objective of making high-quality visual capabilities
% affordable and continuously available to a large consumer population.

\section{Model Architecture}
\label{sec:architecture}

\begin{figure}[!t]
  \centering
  \includegraphics[width=\linewidth]{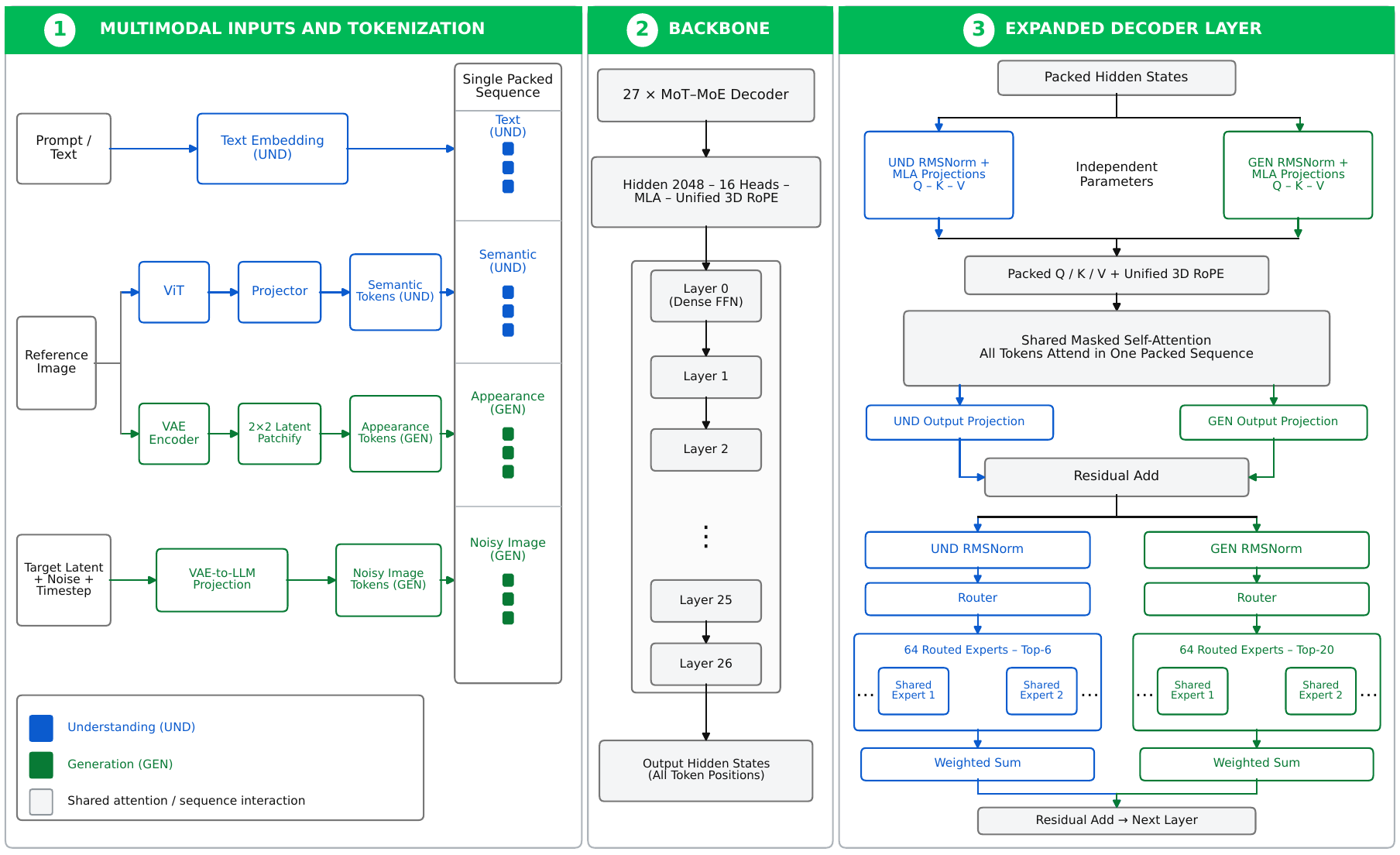}
  \caption{Overview of the \ModelName{} architecture. Tokens on the
  understanding (UND) and generation (GEN) paths retain path-specific
  projections and experts while sharing masked self-attention over one packed
  multimodal sequence.}
  \label{fig:architecture}
\end{figure}

\subsection{Architecture Overview}

\ModelName{} is a unified model for image creation, instruction-guided editing,
and multi-reference generation. Its architecture combines an interleaved
multimodal sequence representation with a dual-path Mixture-of-Transformers
(MoT) backbone~\cite{liang2025mixtureoftransformers,deng2025bagel} whose
understanding (UND) and generation (GEN) paths each incorporate
Mixture-of-Experts (MoE) feed-forward
layers~\cite{shazeer2017moe,fedus2022switch}. We refer to this architecture as
MoT--MoE. Understanding tokens (text and image ViT) take the \UNDPath{}, and
generation tokens (clean and noisy image VAE) take the \GENPath{}, while learned
routers select experts within the chosen path. Shared attention couples the two
paths, which otherwise keep independent projections, normalization, and
experts.
As illustrated in Figure~\ref{fig:architecture}, heterogeneous multimodal inputs
are normalized into a common interleaved representation. Text and visual
elements are independently encoded, projected into model-width token sequences,
and serialized in their original semantic order as typed segments within a
single sequence. Task structure is therefore expressed through token
arrangement and attention masks rather than task-specific input pipelines,
allowing the same backbone to process all examples through one uniform sequence
interface.
The unified backbone connects to modality-specific output heads: a
language-model head for discrete text prediction and a flow-matching
head~\cite{lipman2023flowmatching} for continuous image generation. This allows
understanding and generation to share one representation space while retaining
output mechanisms suited to each modality.

\subsection{Unified Interleaved Task Representation}
\label{sec:interleaved-representation}
\ModelName{} is trained jointly on text question answering, visual
understanding, text-to-image synthesis, instruction-guided editing, and
multi-reference generation. Because these tasks differ in schema, sequence
length, and image count, task-specific input pipelines are difficult to maintain, and
image-resolution bucketing does not yield balanced compute. We therefore
adopt a unified interleaved task representation in which the model, the
packer, and the training losses operate on one ordered multimodal sequence
rather than on task-specific formats. This shared interface removes the need
for task-specific input pipelines, while packing by total token count balances
computation using the complete multimodal sequence rather than image resolution
alone.

Formally, each sample is serialized as typed text and visual segments that
preserve source order,
\begin{equation}
  \mathcal{S}
  = [S_1;S_2;\ldots;S_K],
  \qquad
  S_k\in\{\text{text segment},\text{image segment}\}.
\end{equation}
Here, \(K\) is the number of segments, \(S_k\) is the \(k\)-th segment, and
\(\mathcal{S}\) is the resulting source-ordered sequence; semicolons denote
segment concatenation.
In implementation, text is tokenized with the pretrained language-model
tokenizer, while each visual segment is placed as continuous ViT or VAE
embeddings. After embedding, all segments share the model hidden width and
enter the same backbone.

\begin{figure}[t]
  \centering
  \includegraphics[width=\linewidth]{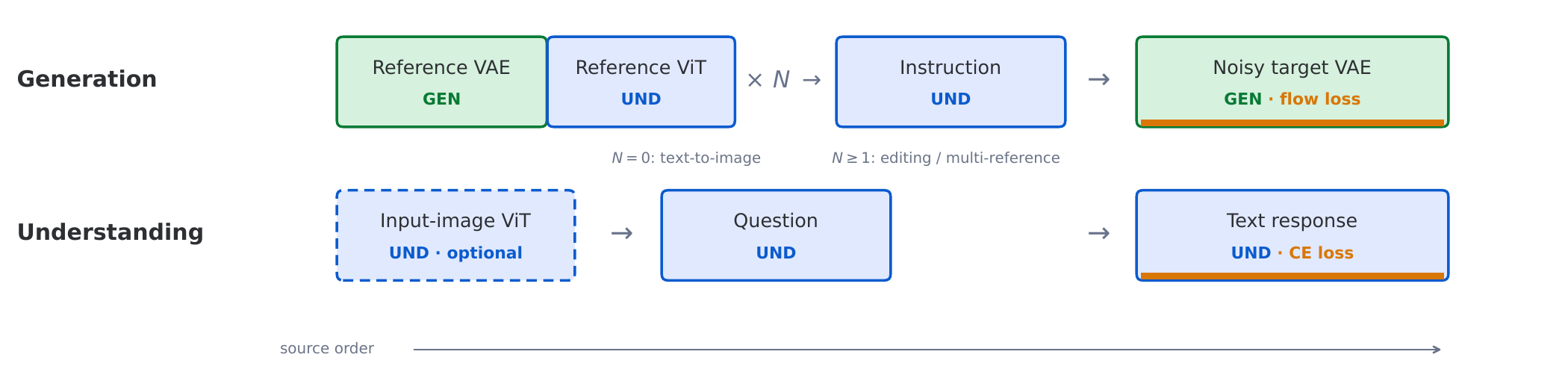}
  \caption{Representative source-ordered sequences under the unified
  interleaved task representation. Blue and green denote the \UNDPath{} and
  \GENPath{},
  respectively. Orange underlining marks supervised segments. The dashed ViT
  segment is optional.}
  \label{fig:interleaved-task-representation}
\end{figure}

Figure~\ref{fig:interleaved-task-representation} illustrates how this shared
interface instantiates two task families. In the generation template, each of
the \(N\) reference images contributes an adjacent clean-VAE segment and ViT
segment, which take the \GENPath{} and \UNDPath{}, respectively. The instruction
subsequently takes the \UNDPath{}, whereas the noisy target VAE takes the
\GENPath{} and receives flow-matching supervision. The number
of reference images then determines the task instance. Setting \(N=0\) yields
text-to-image generation, whereas \(N\geq1\) extends the same template to
editing and multi-reference generation. In the understanding template, an
optional input-image ViT segment and the question precede the response. All
included segments take the \UNDPath{}, while next-token
cross-entropy is applied only to the response. Consequently, omitting the
optional ViT segment reduces the same template to pure-text question answering.

At training time, these per-sample templates are concatenated into one packed
sequence while preserving their segment structure and resetting position
indices independently for each sample. Because concatenation would otherwise
permit information leakage between samples, the shared-attention mask imposes
cross-sample isolation and task-dependent visibility, as described in
Section~\ref{sec:mot-moe-backbone}.

\subsection{Visual Encoding}
\label{sec:visual-encoding}

To instantiate these interleaved templates, each image segment is represented
through two complementary visual encoders.
A Vision Transformer (ViT) encoder~\cite{dosovitskiy2021image,kimivl2025} captures high-level content such
as objects, spatial relations, layout, and text, while \ModelName{}-VAE maps
the image into compact continuous latents that retain the visual details
needed for reconstruction and generation. Both representations are projected
into backbone tokens and placed in the same interleaved sequence. Reference
images therefore provide both content-level guidance and reconstruction-level
visual context, whereas target images are modeled as noisy VAE latents and
recovered through flow matching.

\paragraph{VAE training.}
We train the VAE with a principal-component compression strategy that
encourages its compact latent space to preserve the dominant structure of the
visual representation instead of allocating equal capacity to all pixel-level
variation. This design is intended to retain semantically important content
and fine visual detail under compression, providing latents that support both
faithful reconstruction and generative modeling. Specifically, training
proceeds in three stages: reconstruction pretraining, semantic alignment, and
text-oriented decoder fine-tuning. Given an image $x$, the encoder
$\mathcal{E}$ produces $z=\mathcal{E}(x)$ and the decoder $\mathcal{D}$
reconstructs $\hat{x}=\mathcal{D}(z)$. The reconstruction and combined
objectives are
\begin{equation}
  \mathcal{L}_{\mathrm{rec}}
  = \lVert x-\hat{x}\rVert_1
  + \mathcal{L}_{\mathrm{LPIPS}},
  \qquad
  \mathcal{L}_{\mathrm{VAE}}
  = 1.5\,\mathcal{L}_{\mathrm{rec}}
  + 10\,\mathcal{L}_{\mathrm{rep}},
\end{equation}
where $\mathcal{L}_{\mathrm{LPIPS}}$ is the perceptual reconstruction
loss~\cite{zhang2018perceptual} and $\mathcal{L}_{\mathrm{rep}}$ is the
representation-alignment objective defined below. Stage I jointly optimizes
$\mathcal{E}$ and $\mathcal{D}$ with $\mathcal{L}_{\mathrm{rec}}$. Stage II
adds $\mathcal{L}_{\mathrm{rep}}$ and jointly trains the autoencoder and a
representation predictor. Stage III freezes $\mathcal{E}$ and fine-tunes
$\mathcal{D}$ on text-rich images with $\mathcal{L}_{\mathrm{rec}}$.

Stage II implements representation alignment by jointly training a lightweight
diffusion transformer $g_\phi$~\cite{peebles2023dit} to recover high-level
semantic features from corrupted latents and thereby inject them into the
latent space. We draw a scalar $\epsilon_t\sim\mathcal{N}(0,1)$ and set
$t=\sigma(\epsilon_t)$, where $\sigma$ is the logistic sigmoid. Independently,
we sample latent-shaped noise $\epsilon\sim\mathcal{N}(0,\mathbf{I})$, where
$\mathbf{I}$ is the identity covariance, construct
$z_t=(1-t)z+t\epsilon$, and predict $r_z=g_\phi(z_t,t)$. Here, $\phi$ denotes
the predictor parameters, $z_t$ the corrupted latent, and $r_z$ the predicted
semantic feature. We align $r_z$ with features $r$ from a frozen DINOv2
encoder~\cite{oquab2024dinov2} by matching their normalized self-similarity
matrices:
\begin{equation}
  \mathcal{L}_{\mathrm{rep}}
  =
  \left\lVert
    \bar{r}_z\bar{r}_z^{\top}
    -
    \bar{r}\bar{r}^{\top}
  \right\rVert_F^2,
  \qquad
  \bar{r}=\frac{r}{\lVert r\rVert_2},
  \quad
  \bar{r}_z=\frac{r_z}{\lVert r_z\rVert_2}.
\end{equation}
For each image, $r,r_z\in\mathbb{R}^{M\times d}$ are feature matrices over
$M$ aligned visual positions and $d$ feature channels. The bars denote
row-wise $\ell_2$ normalization, ${}^{\top}$ denotes transpose, and
$\lVert\cdot\rVert_F$ is the Frobenius norm.
This objective preserves the pairwise semantic geometry of DINOv2 features~\cite{oquab2024dinov2}
while remaining invariant to global scaling and orthogonal transformations of
the feature space.

\paragraph{Reconstruction and generation results.}
To validate this staged training, we evaluate reconstruction with PSNR and SSIM
on COCO and a 5k-image ImageNet subset at 512 resolution, and on an in-house
set of text-rich posters at 1024,
as summarized in Table~\ref{tab:vae-reconstruction}. \ModelName{}-VAE attains
the highest PSNR on all three reconstruction sets and, under the same
$8\times$ spatial compression and 16 latent channels used by
Qwen-VAE~\cite{wu2025qwenimage}, FLUX-VAE~\cite{flux2024}, and
NextStep-VAE~\cite{nextstep2025}, remains competitive in SSIM.
Reconstruction alone does not decide whether a tokenizer is a good generator,
so we freeze each VAE and train a class-conditional
LightningDiT-XL/2~\cite{yao2025lightningdit} on ImageNet-1K at
$256{\times}256$. The diffusion head uses
$v$-prediction~\cite{salimans2022progressive} with a batch size of 256 and a
learning rate of $10^{-4}$, without timestep shift. After 80 epochs we sample
50k images without classifier-free guidance~\cite{ho2022classifierfree} and
report FID~\cite{heusel2017gans} and IS~\cite{salimans2016improved}.
Hunyuan-3.0-VAE~\cite{hunyuanimage2025} is omitted from this comparison
because the autoencoder is larger than LightningDiT-XL/2 itself, so encoding
and decoding would dominate the training loop and make the 80-epoch protocol
impractical. Under this protocol \ModelName{}-VAE attains the highest IS and
the lowest FID among 16-channel tokenizers. A controlled ablation further
shows that \ModelName{}-VAE accelerates pretraining convergence relative to
the alternative VAE, providing end-to-end evidence for its use in the main
model.

\begin{table}[t]
  \centering
  \small
  \setlength{\tabcolsep}{2.6pt}
  \begin{tabular}{@{}l|ccc|cc|cc|cc|cc@{}}
    \toprule
    \multicolumn{1}{c|}{\multirow{3}{*}[-0.6ex]{Model}}
      & \multicolumn{3}{c|}{\multirow{2}{*}[-0.2ex]{Architecture}}
      & \multicolumn{6}{c|}{Reconstruction}
      & \multicolumn{2}{c}{Generation} \\
    \cmidrule(lr){5-10} \cmidrule(lr){11-12}
      & & &
      & \multicolumn{2}{c|}{COCO 512}
      & \multicolumn{2}{c|}{ImageNet 512}
      & \multicolumn{2}{c|}{\makecell{In-house\\posters 1024}}
      & \multicolumn{2}{c}{ImageNet 256} \\
    \cmidrule(lr){2-4} \cmidrule(lr){5-6} \cmidrule(lr){7-8} \cmidrule(lr){9-10} \cmidrule(lr){11-12}
      & Params & Spatial & Ch.
      & PSNR$\uparrow$ & SSIM$\uparrow$
      & PSNR$\uparrow$ & SSIM$\uparrow$
      & PSNR$\uparrow$ & SSIM$\uparrow$
      & FID$\downarrow$ & IS$\uparrow$ \\
    \midrule
    Qwen-VAE~\cite{wu2025qwenimage}
      & $126.89$M & $8\times$ & $16$
      & 33.60 & 0.9249
      & 32.96 & 0.9165
      & 37.98 & 0.9658
      & 15.29 & 84.62 \\
    FLUX-VAE~\cite{flux2024}
      & $83.82$M & $8\times$ & $16$
      & 34.20 & 0.9486
      & 33.91 & \underline{0.9488}
      & \underline{37.99} & \underline{0.9771}
      & 27.63 & 55.10 \\
    FLUX.2-VAE~\cite{flux22025}
      & $84$M & $8\times$ & $32$
      & \underline{34.61} & \textbf{0.9540}
      & \underline{34.34} & \textbf{0.9562}
      & 37.13 & \textbf{0.9780}
      & \textbf{9.87} & \underline{90.85} \\
    NextStep-VAE~\cite{nextstep2025}
      & $83.82$M & $8\times$ & $16$
      & 33.12 & 0.9263
      & 32.65 & 0.9216
      & 36.47 & 0.9644
      & 22.94 & 62.80 \\
    Hunyuan-3.0-VAE~\cite{hunyuanimage2025}
      & $1260.67$M & $16\times$ & $32$
      & 32.32 & 0.8998
      & 31.74 & 0.8893
      & 36.54 & 0.9536
      & -- & -- \\
    \midrule
    \ModelName{}-VAE
      & $83.75$M & $8\times$ & $16$
      & \textbf{34.89} & \underline{0.9498}
      & \textbf{34.47} & 0.9482
      & \textbf{38.49} & 0.9769
      & \underline{12.21} & \textbf{97.96} \\
    \bottomrule
  \end{tabular}
  \caption{VAE reconstruction and generation. Best results are in bold and
  second-best are underlined. A dash indicates that generation was not
  evaluated.}
  \label{tab:vae-reconstruction}
\end{table}

\begin{table}[t]
  \centering
  \small
  \begingroup
  \renewcommand{\arraystretch}{1.18}
  \setlength{\tabcolsep}{4pt}
  \begin{tabular}{@{}llcc@{}}
    \toprule
    \textbf{Group} & \textbf{Configuration}
      & \textbf{\UNDPath{}} & \textbf{\GENPath{}} \\
    \midrule
    \multirow{5}{*}{\textbf{Backbone}}
      & Decoder layers & \multicolumn{2}{c}{27} \\
    \cmidrule(lr){2-4}
      & Hidden size & \multicolumn{2}{c}{2,048} \\
    \cmidrule(lr){2-4}
      & FFN layout
      & \multicolumn{2}{c}{Layer 0 dense; Layers 1--26 MoE} \\
    \cmidrule(lr){2-4}
      & Intermediate size
      & \multicolumn{2}{c}{11,264 (dense); 1,408 (expert)} \\
    \cmidrule(lr){2-4}
      & Experts per MoE layer
      & \multicolumn{2}{c}{64 routed + 2 shared per path} \\
    \midrule
    \multirow{2}{*}{\textbf{Routing}}
      & Routed experts selected per token & 6 & 20 \\
    \cmidrule(lr){2-4}
      & Activated parameters per token & 2.2B & 5.4B \\
    \midrule
    \textbf{Scale}
      & Full-model stored parameters & \multicolumn{2}{c}{31.7B} \\
    \bottomrule
  \end{tabular}
  \endgroup
  \caption{Configuration and parameter scale of the \ModelName{} MoT--MoE
  backbone. The \UNDPath{} and \GENPath{} use independent weights. The
  full-model total includes the ViT encoder and excludes the VAE.}
  \label{tab:mot-moe-config}
\end{table}

\subsection{MoT--MoE Backbone}
\label{sec:mot-moe-backbone}

MoT and MoE address two distinct requirements. MoT separates understanding and
generation into independent parameter paths while retaining joint attention
for cross-modal interaction. This separation reduces optimization interference
between the two tasks, which differ substantially in their representations and
training objectives~\cite{liang2025mixtureoftransformers,deng2025bagel}.
However, scaling a dense MoT backbone requires widening its feed-forward
layers, increasing the parameters and computation activated by every token.
MoE offers a more efficient way to increase model capacity by adding experts
while activating only a sparse subset for each
token~\cite{shazeer2017moe,fedus2022switch}. A single expert pool shared by
understanding and generation would nevertheless weaken the task separation
introduced by MoT and could reintroduce interference that hinders convergence.
We therefore equip each MoT path with an independent MoE,
preserving understanding--generation decoupling while enabling sparse
parameter scaling. The resulting backbone retains standard ViT and VAE
interfaces and does not rely on additional timestep-conditioned normalization
modules such as AdaLN or AdaLN-Zero~\cite{peebles2023dit}.

Operationally, the backbone contains 27 decoder layers with a hidden size of
2,048.
The full multimodal model, including the ViT encoder, stores 31.7B parameters;
the independent 83.7M-parameter VAE is excluded.
Counts use $\mathrm{B}=10^9$ and include frozen parameters in stored totals
(Table~\ref{tab:mot-moe-config}). The backbone instantiates this dual-path MoT
design in every decoder layer: each layer keeps separate parameters for the
\UNDPath{} and \GENPath{} and couples them through joint attention. The first
layer (Layer 0) uses a path-specific dense
FFN with intermediate size 11,264, whereas Layers 1--26 replace each dense FFN
with a path-specific MoE sublayer. MoT therefore supplies the two computation
paths of the backbone, while MoE provides sparse conditional capacity
inside the feed-forward sublayer of each path.

With top-6 expert routing on the \UNDPath{} and top-20 expert routing on the
\GENPath{}, each token on the former activates approximately 2.2B decoder
parameters, while each latent token on the latter activates approximately
5.4B parameters in the generation component. During sampling,
conditioning tokens are computed once and cached, whereas noisy target latents
are recomputed at every denoising step; the 5.4B activated-parameter figure
therefore refers to the generation component repeatedly evaluated for each target
latent.

\paragraph{MoT paths and joint attention.}
Each token is assigned to a fixed path according to its representation type:
text and ViT tokens use the \UNDPath{}, while clean reference and noisy target
VAE tokens use the \GENPath{}. At each decoder layer, hidden states are gathered
by path and mapped to queries, keys, and values through path-specific
pre-attention RMSNorm~\cite{zhang2019root} and Multi-head Latent Attention
(MLA) projections~\cite{deepseekv2}. These tensors are then scattered back to
their original sequence positions, and a single masked scaled-dot-product
attention operation is applied across both paths. Parameter specialization
therefore does not partition the attention context. A noisy target query on the
\GENPath{}, for example, can attend to
text and ViT keys/values from the \UNDPath{} as well as clean reference
keys/values from the \GENPath{}. Which cross-path connections are visible is
determined by the attention mask.

\begin{figure}[!t]
  \centering
  \includegraphics[width=0.78\linewidth]{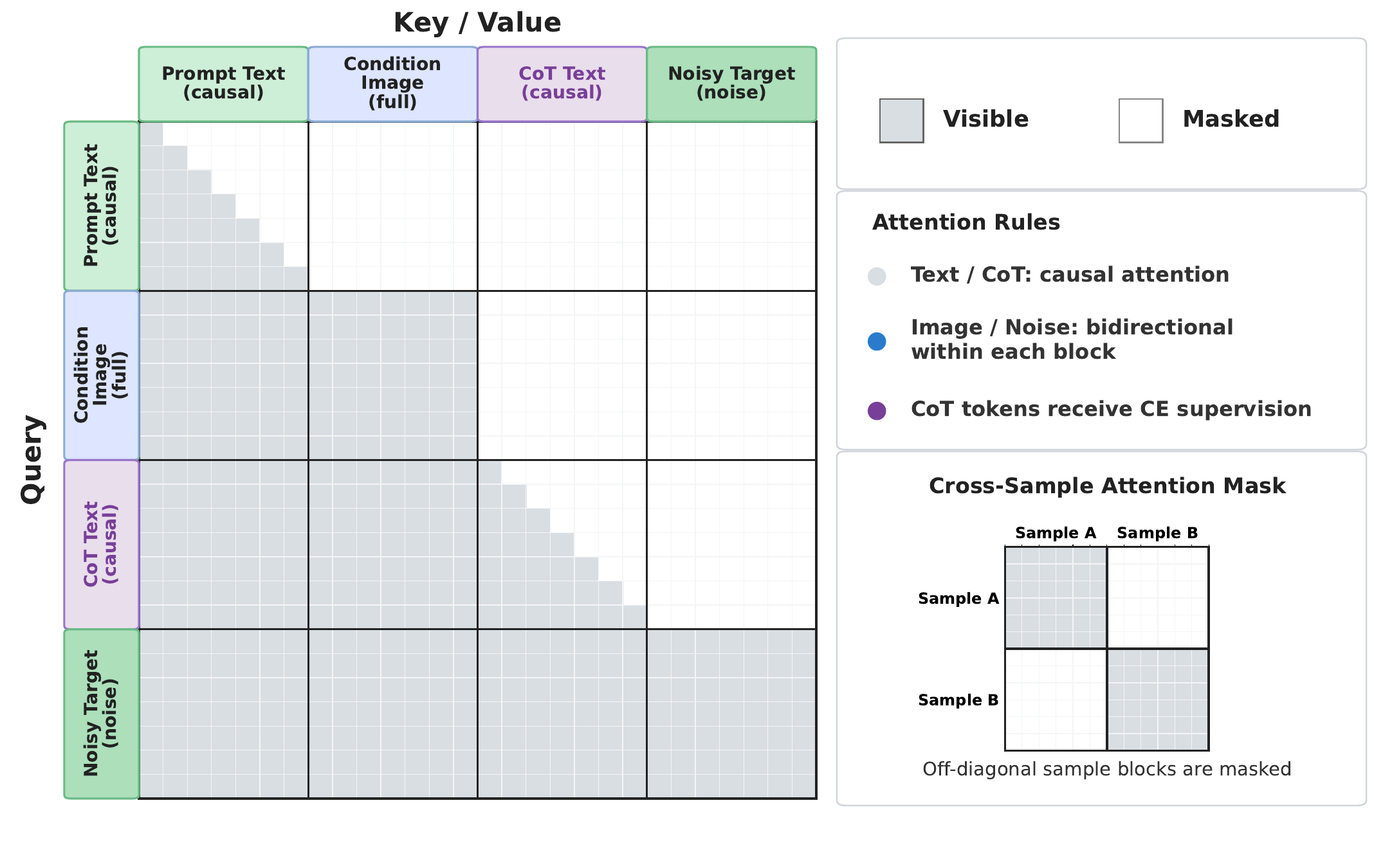}
  \caption{Block-structured attention mask for a generation sequence containing
  prompt text, a condition image, CoT text, and a noisy target. Rows denote
  queries and columns denote keys/values; gray cells are visible and white cells
  are masked. Text blocks use causal attention, whereas the condition image and
  noisy target use bidirectional attention within their respective blocks.
  Each block can attend to preceding conditions, while noisy-target keys/values
  remain hidden from earlier blocks. Purple marks CoT tokens receiving
  cross-entropy supervision, and the inset shows isolation between packed
  samples.}
  \label{fig:block-attention-mask}
\end{figure}

%TODO[huangzp]: to fix this paragraph
Figure~\ref{fig:block-attention-mask} illustrates the block-structured
attention mask for a representative prompt--condition--CoT--target sequence.
The mask first isolates packed samples, producing independent diagonal blocks
in which tokens can attend only within the same sample. Inside each block,
text follows causal order: a query can read permitted earlier segments and
only current-or-earlier positions within its own text segment. ViT and clean
reference VAE tokens instead use bidirectional attention within their
respective image segments while retaining access to permitted earlier context.
Noisy-target VAE tokens likewise attend bidirectionally within the target
latent grid and read all preceding non-target conditions, but their keys and
values remain invisible to every other segment. This asymmetry prevents target
leakage while preserving full spatial interaction during denoising.

% \clearpage

After joint attention, the outputs are regrouped by path and passed through
their respective output projections before the residual update. Each decoder
layer therefore combines path-specific parameters with a shared attention
context whose information flow is controlled by the mask.

%TODO[huangzp]: to check this later
\paragraph{MoE feed-forward sublayers.}
Following the shared-attention sublayer, MoE expands the feed-forward capacity
of each MoT path without activating every parameter for every token. This
allows the model to represent the heterogeneous concepts and appearance
patterns present in multimodal data at tractable compute. Layers 1--26 use this
sparse computation, while the first block retains path-specific dense FFNs.

A purely routed MoE is less suitable for computation required by most tokens.
Recurring conditions, such as flow-matching timestep information and common
language or visual concepts, benefit from a stable path, whereas load balancing
deliberately distributes traffic across routed experts. Without shared experts,
such common computation may be fragmented and redundantly learned throughout
the routed pool. We therefore combine 64 routed experts with two always-active
shared experts~\cite{deepseekv2} in each path: the shared experts provide a
common computational basis, while routing reserves most sparse capacity for
specialized and long-tail patterns. For the hidden state $h_i$ of token $i$ at
a sparse decoder layer and path $p\in\{\mathrm{und},\mathrm{gen}\}$, the
feed-forward output is
\begin{equation}
  \operatorname{FFN}_{p}(h_i)
  =
  \sum_{s=1}^{2} E^{\mathrm{shared}}_{p,s}(h_i)
  +
  \sum_{e\in\mathcal{T}_{p}(i)}
    \alpha_{p,e}(h_i)E^{\mathrm{routed}}_{p,e}(h_i).
\end{equation}
Here, $s$ and $e$ index shared and routed experts, respectively;
$E^{\mathrm{shared}}_{p,s}$ and $E^{\mathrm{routed}}_{p,e}$ are their
feed-forward mappings, $\mathcal{T}_{p}(i)$ is the selected subset of the 64
routed experts, and $\alpha_{p,e}(h_i)$ is the corresponding mixture weight.
Each path owns its sigmoid-scored router and expert
pool~\cite{deepseekv3}. The shared experts are active for every token on that
path. Here, ``shared'' means shared within a path, not shared between the
\UNDPath{} and \GENPath{}. For token $i$ on path $p$, routing uses
\begin{equation}
  u_{i,e}=\sigma\!\left(g_{p,e}(h_i)\right),\qquad
  \mathcal{T}_{p}(i)
  =\operatorname{TopK}_{e}\!\left(u_{i,e}+b_{p,e},K_p\right),
  \quad K_{\mathrm{und}}=6,\;K_{\mathrm{gen}}=20,
\end{equation}
where $h_i$ is the hidden state of token $i$, $g_{p,e}(h_i)$ is the router
logit for expert $e$, $\sigma$ is the sigmoid function, $u_{i,e}$ is the
resulting routing score, $b_{p,e}$ is a non-gradient correction bias, $K_p$ is
the number of selected experts, and $\mathcal{T}_p(i)$ is their index set. The
mixture coefficients $\alpha_{p,e}$ are obtained by normalizing the original
sigmoid scores $u_{i,e}$ over the selected experts and applying a routed
scaling factor of 2.446. The correction bias changes expert selection but not
the mixture weights. The asymmetric top-$k$ assigns more active feed-forward
capacity to appearance synthesis, while the disjoint routers and expert
parameters prevent the two token populations from competing for one expert
pool.

Beyond asymmetric capacity allocation, stable expert utilization requires an
explicit balancing mechanism. The MoE on each path therefore maintains its own
load statistics and a non-gradient correction bias. If $n_{p,e}$ is the
assignment count accumulated for expert $e$ on path $p$ during the current
balancing interval and $\bar n_p$ is the mean over the 64 routed experts, the
bias is updated by
\begin{equation}
  b_{p,e}
  \leftarrow
  b_{p,e}+\eta\,\operatorname{sign}(\bar n_p-n_{p,e}),
  \qquad
  b_{p,e}
  \leftarrow
  b_{p,e}-\operatorname{mean}_{m}b_{p,m}.
\end{equation}
Here, $\eta>0$ is the bias-update step size, and
$\operatorname{mean}_{m}$ averages over the 64 routed experts indexed by $m$.
Overloaded experts become less likely to be selected and underloaded experts
more likely to be selected. The subsequent zero-mean projection prevents the
bias drift observed when this constraint is omitted. A sequence-level
auxiliary loss~\cite{deepseekv3} with coefficient $10^{-4}$ adds a weak
differentiable regularizer, but the bias update remains the primary assignment
correction so that load balancing does not dominate the language or
flow-matching objectives.

\subsection{Text-Compatible Unified 3D RoPE}
\label{sec:rope3d}

\begin{figure}[t]
  \centering
  \includegraphics[width=0.72\linewidth]{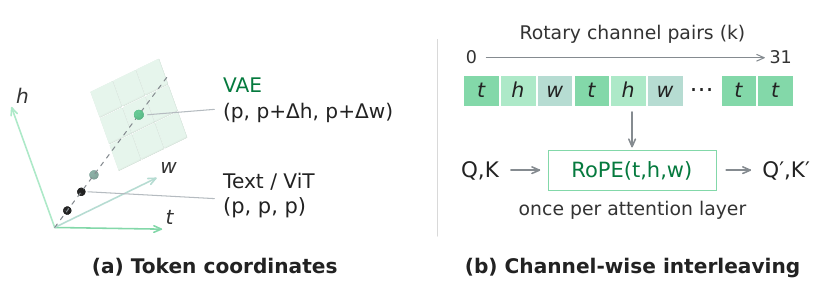}
  \caption{Text-compatible unified 3D RoPE. (a) In token-coordinate space,
  text and ViT tokens remain on the diagonal $(p,p,p)$, whereas VAE tokens
  retain $t=p$ and occupy a centered local $h$--$w$ grid
  $(p,p+\Delta h,p+\Delta w)$. (b) The pretrained rotary channel pairs are
  assigned to $t$, $h$, and $w$ in interleaved order, and the resulting
  three-axis RoPE is applied once to the rotary subspaces of queries and keys
  in each attention layer.}
  \label{fig:rope3d}
\end{figure}

Text and visual tokens share one attention context but require different
positional structures. The backbone inherits a pretrained one-dimensional
RoPE~\cite{su2024roformer}. Visual generation, however, requires intra-image
spatial structure that a purely sequential index cannot express: patches in the
same VAE latent grid often share one packed sequence position, so 1D RoPE does
not distinguish height from width. Following multidimensional RoPE designs for
multimodal inputs~\cite{wang2024qwen2vl}, we lift rotary embeddings to a single
three-axis coordinate system $(t,h,w)$, where $t$ is the sequence-axis
coordinate and $h$ and $w$ are the spatial row and column coordinates, and use
one rotary operator in each attention layer rather than stacking a second
spatial encoding on top of the language RoPE.
Figure~\ref{fig:rope3d} presents the resulting construction from complementary
token-coordinate and channel-allocation views.

\paragraph{Interleaved axis allocation.}
Panel~(b) of Figure~\ref{fig:rope3d} shows the channel-wise mapping. The
64-dimensional rotary subspace of each query and key forms 32 channel pairs.
We reuse their original inverse frequencies and assign 12 pairs to $t$ and 10
pairs each to $h$ and $w$. These assignments are interleaved across the
frequency order rather than grouped into contiguous blocks.
This interleaving keeps the range of rotary periods comparable across the three
axes. By contrast, a grouped split would assign each axis a contiguous
frequency band and therefore impose systematically different period ranges.
This mismatch is particularly harmful when the temporal axis receives the
higher-frequency pairs: its shorter periods wrap within long packed prompts,
which in our experiments led to gradient-norm spikes and disrupted training.
Distributing every axis across the full frequency spectrum avoids this
imbalance while retaining axis-specific geometric coordinates.

\paragraph{Compatibility with pretrained text RoPE.}
Panel~(a) shows why text compatibility is exact rather than approximate. Let
$k\in\{1,\ldots,32\}$ index the rotary channel pairs and $\omega_k$ denote the
inverse frequency of pair $k$. The 3D construction reuses these frequencies
and assigns each pair to one axis $a(k)\in\{t,h,w\}$ using the interleaved
mapping defined above. For a text token, $p$ is the same position index used by
the pretrained language model, and its three coordinates are
$(t,h,w)=(p,p,p)$. The 3D phase and its reduction to the pretrained 1D phase
can then be written together as
\begin{equation}
  \begin{aligned}
    \phi_k^{\mathrm{3D}}(t,h,w)
    &=
    \begin{cases}
      t\,\omega_k, & a(k)=t,\\
      h\,\omega_k, & a(k)=h,\\
      w\,\omega_k, & a(k)=w,
    \end{cases}\\
    \phi_k^{\mathrm{3D}}(p,p,p)
    &=p\,\omega_k=\phi_k^{\mathrm{1D}}(p),
    \qquad \forall k.
  \end{aligned}
  \label{eq:rope-text-compatibility}
\end{equation}
Because RoPE applies the same sine and cosine rotations whenever the phases
match, text tokens receive exactly the same positional transformation as in
the pretrained language model. The interleaved axis assignment changes the
rotation only for tokens with spatially offset coordinates, namely VAE latent
tokens. It therefore adds spatial structure to visual latents while preserving
the pretrained positional encoding for text.

\paragraph{Visual-token coordinates.}
ViT and VAE tokens use different coordinate assignments within this scheme.
All tokens in a ViT image segment retain the shared segment coordinate
$(p,p,p)$ because their patch-level geometry has already been encoded by the
vision encoder.
The VAE grid likewise shares one packed position, but it has no such
encoder-side spatial RoPE and therefore needs explicit height--width structure
in the backbone. Consequently, only VAE appearance tokens keep the temporal
index fixed while adding a centered spatial offset,
\begin{equation}
  (t,h,w)=(p,\;p+\Delta h,\;p+\Delta w).
\end{equation}
For each latent token, $\Delta h$ and $\Delta w$ are its row and column offsets
from the center of the VAE latent grid; all tokens in the same image segment
share the anchor $p$.
The local grid is centered so that images of different sizes share a common
spatial origin and do not inherit a resolution-dependent top-left bias. Adding
$p$ as an anchor keeps the phase at text--image boundaries continuous, while
intra-image relative geometry is carried by $(\Delta h,\Delta w)$. Attention
therefore sees a single rotary system in which text remains on the original 1D
manifold, ViT tokens occupy a shared segment index, and VAE latents occupy a
local 2D neighborhood around that index.

\subsection{Output Heads and Joint Objectives}
\label{sec:objective}

\paragraph{Modality-specific supervision.}
After the unified sequence has passed through the backbone, the resulting
hidden states are decoded by separate modality-specific heads.
Text states on the \UNDPath{} are read by the language-model head and trained
with autoregressive next-token prediction. Noisy target states on the
\GENPath{} are projected into the VAE latent space, where they parameterize a
flow-matching velocity. Condition tokens---instructions, ViT features, and
clean reference latents---enter attention as context and do not receive
generation-head supervision.

Accordingly, generation supervision is implemented through conditional flow
matching~\cite{lipman2023flowmatching}. For a clean latent $z$, Gaussian noise
$\epsilon\sim\mathcal{N}(0,\mathbf{I})$, and a sampled timestep
$t\in[0,1]$, we define the interpolant $z_t$ and target velocity $v^\star$ as
\begin{equation}
  z_t=(1-t)z+t\epsilon,\qquad v^\star=\epsilon-z.
\end{equation}
The generation head predicts $\hat v_\theta(z_t,c,t)$, where $c$ is the
conditioning context and $\theta$ denotes the model parameters, and is trained
by
\begin{equation}
  \mathcal{L}_{\mathrm{FM}}
  =\mathbb{E}\left[\left\|\hat v_\theta-v^\star\right\|_2^2\right].
\end{equation}
The expectation is taken over training latents, Gaussian noise, and sampled
timesteps, and $\hat v_\theta$ has the same shape as $v^\star$.

\paragraph{Joint objective and loss normalization.}
Supervised response and CoT positions use standard autoregressive cross entropy
$\mathcal{L}_{\mathrm{CE}}$. Together with the router regularizer, these two
task losses form the joint objective
\begin{equation}
  \mathcal{L}
  =
  \lambda_{\mathrm{CE}}\,\mathcal{L}_{\mathrm{CE}}
  +
  \lambda_{\mathrm{FM}}\,\mathcal{L}_{\mathrm{FM}}
  +
  \lambda_{\mathrm{aux}}\,\mathcal{L}_{\mathrm{router}}.
\end{equation}
Here, $\mathcal{L}_{\mathrm{router}}$ is the expert-load regularizer, and
$\lambda_{\mathrm{CE}}$, $\lambda_{\mathrm{FM}}$, and
$\lambda_{\mathrm{aux}}$ weight the text, flow-matching, and router terms,
respectively. In the current configuration, these weights are $0.25$, $1$, and
$10^{-4}$.

Packed multi-resolution training uses different reduction units for the two
primary objectives. The cross-entropy loss is averaged over supervised text
tokens, whereas the flow-matching loss is first averaged over all VAE tokens
within each target image and then aggregated at the image level. Consequently,
a $1024\times1024$ target and a $1536\times1536$ target contribute equally to
the flow-matching objective despite containing different numbers of latent
tokens. This image-level normalization prevents larger images from receiving
greater weight solely because of their resolution and ensures balanced
optimization across image areas. The timestep schedule and remaining
optimization details are given in Section~\ref{sec:training}.
\section{Data}
\label{sec:data}

\begin{figure}[t]
  \centering
  \includegraphics[width=\textwidth]{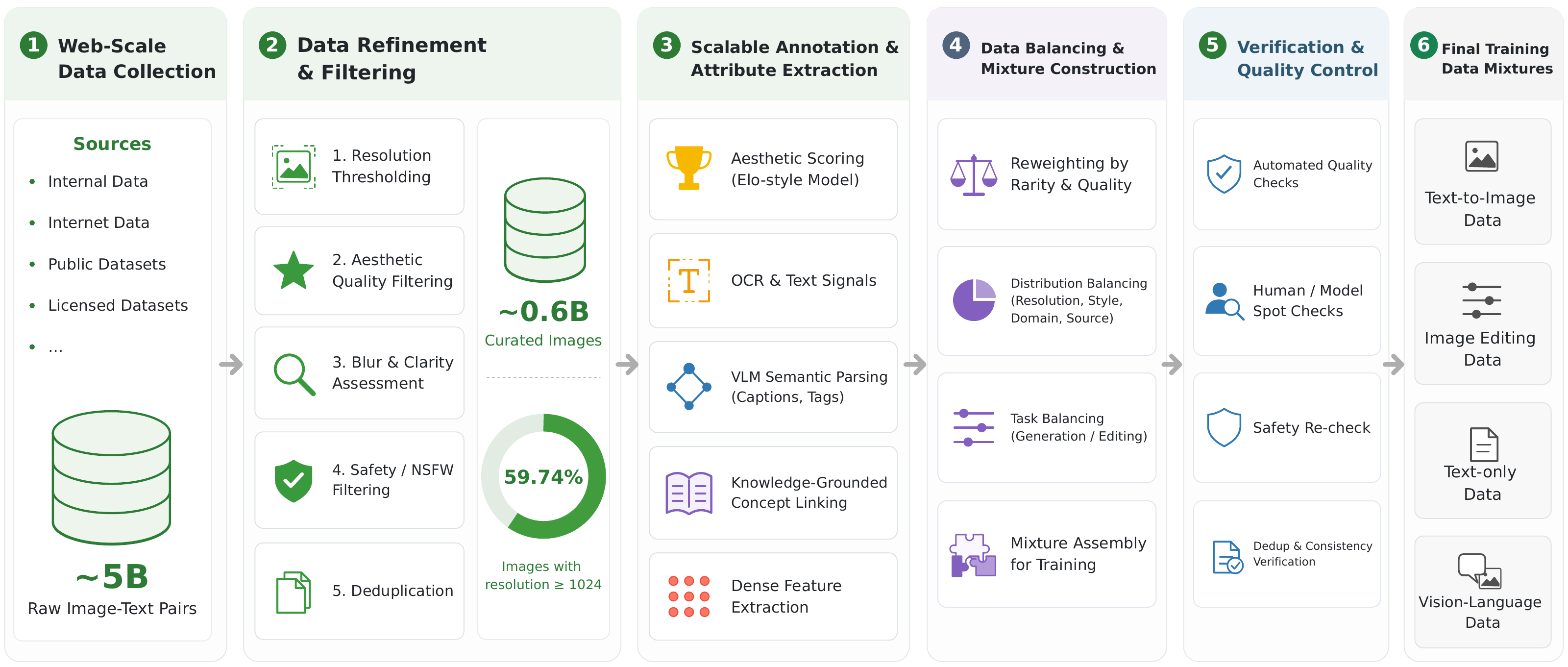}
  \caption{Overview of the data refinement, annotation, and training-mixture
  construction pipeline.}
  \label{fig:data_framework}
\end{figure}

Data plays a central role in \ModelName{}. The natural frequency of web-scale image-text pairs is not a usable recipe for controllable generation: head concepts dominate the pool, long-tail visual semantics remain under-exposed, and free-form editing construction collapses onto a few common operations. Prior work on billion-scale image-text datasets likewise shows that filtering and curation materially determine the value of a raw web pool~\cite{schuhmann2022laion5b,gadre2023datacomp}. As shown in Figure~\ref{fig:data_framework}, we therefore build a generation-oriented data engine that reorganizes this pool around what the model must learn to generate, edit, and compose, rather than around becoming a general vision--language assistant.

Specifically, the pipeline first refines noisy web data for quality, retaining a curated image pool with aesthetic, resolution, OCR, and safety signals. It then indexes the pool along two complementary structures: a knowledge-grounded vocabulary that makes named visual concepts addressable, and large-scale visual clustering that groups images by unnamed but coherent appearance. These two structures supply coverage and balance. On this organized pool we construct text-to-image data, instruction-guided editing and multi-reference generation data, and text-only together with vision-language instruction data, including VQA, that support the prompt interpretation on which generation depends. Finally, training-time evaluation feeds back into filtering, sampling, and mixture ratios, so that data composition remains an active optimization variable rather than a fixed recipe.

\subsection{Web-Scale Curation}
\label{sec:data-refine}

\subsubsection{Infrastructure}
\label{sec:data-infrastructure}

\textbf{Storage and Resource Scheduling.}
The raw data is stored as individual objects in cloud object storage (COS), using a bucketed layout that supports cross-region and cross-cluster
access without packaging images into large archives. Ray services~\cite{moritz2018ray} dynamically
schedule heterogeneous GPU and CPU workers, while Pulsar queues decouple
storage I/O from data ingestion, cleaning, annotation and
quality-control jobs. This compute- and message-driven design supports
parallel processing and asynchronous result feedback at web scale.

\begin{figure}[!t]
  \centering
  \includegraphics[width=\textwidth]{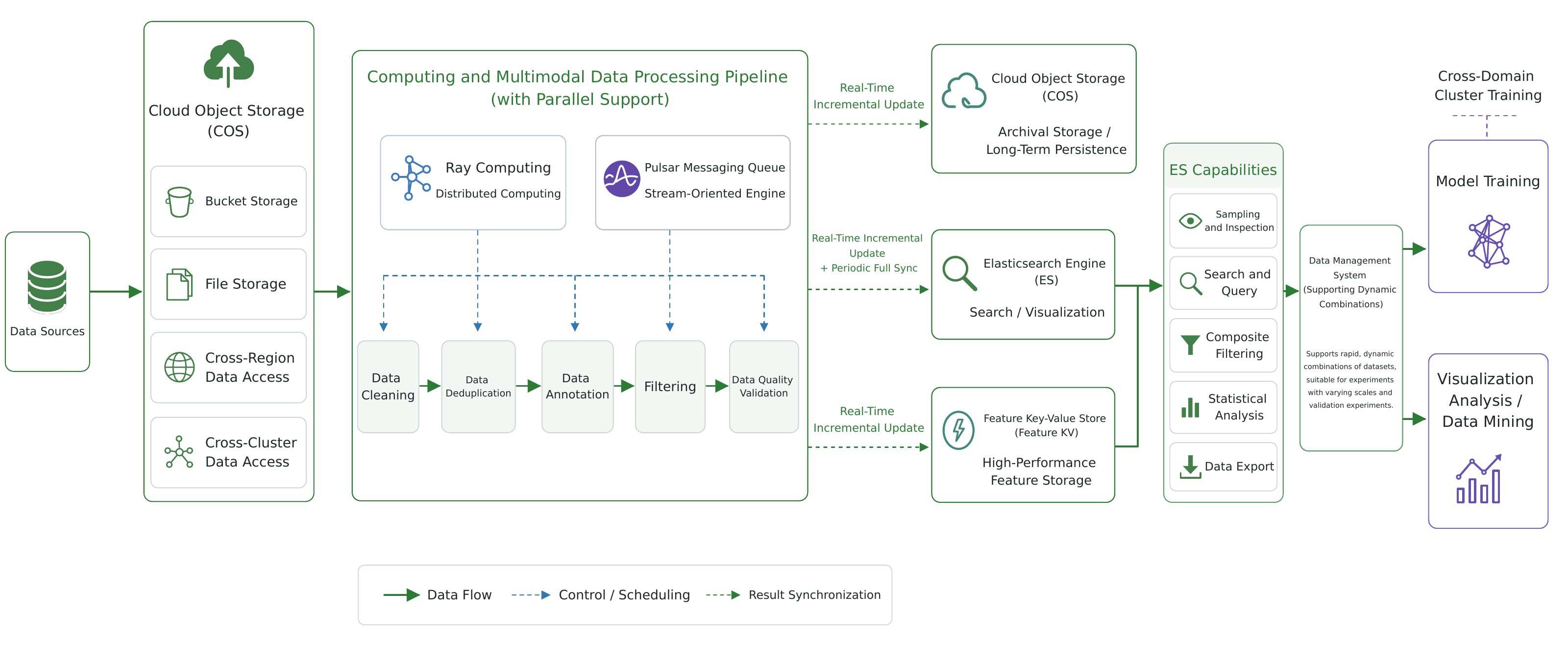}
  \caption{Scalable data infrastructure based on distributed COS storage,
  Ray services, Pulsar queues, ES retrieval, feature KV storage, and dynamic
  dataset assembly.}
  \label{fig:data-infrastructure}
\end{figure}

\textbf{Data Indexing.}
As shown in Figure~\ref{fig:data-infrastructure}, the processed results are synchronized to three complementary systems. Structured and semi-structured outputs are persistently stored in COS. Textual, vector, attribute, and metadata fields are indexed in Elasticsearch (ES) for sample-level inspection, retrieval, online queries, combined filtering, statistical analysis, and export. Dense annotation features are stored in a high-performance feature key--value store. A data-management system built on these services supports dynamic dataset assembly for different training and validation requirements. Targeted subsets with specific quality levels, sources, styles, or vertical domains can typically be selected and exported within half a day. The full pipeline sustains more than 15,000 QPS, with selected modules exceeding 100,000 QPS, and has remained stable over multi-month production runs. This shared infrastructure provides the foundation for consistently normalizing and processing heterogeneous data sources.

\subsubsection{Data Schema}

On top of this infrastructure, each raw image-text sample is normalized into a unified schema that preserves image identity, source metadata, resolution, aspect ratio, raw text, and storage information. The shared representation gives filtering, annotation, retrieval, and task-construction stages access to the same sample identity and metadata fields, so the same quality-control and attribute-extraction operators can be applied consistently across otherwise heterogeneous sources.

\subsubsection{Quality Filtering}

\begin{figure}[t]
  \centering
  \begin{subfigure}{0.48\textwidth}
    \centering
    \includegraphics[width=\linewidth]{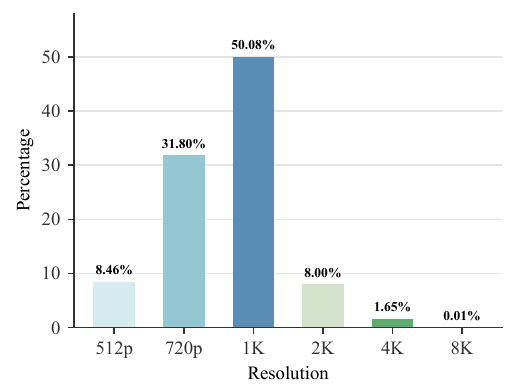}
    \caption{Resolution distribution.}
  \end{subfigure}\hfill
  \begin{subfigure}{0.48\textwidth}
    \centering
    \includegraphics[width=\linewidth]{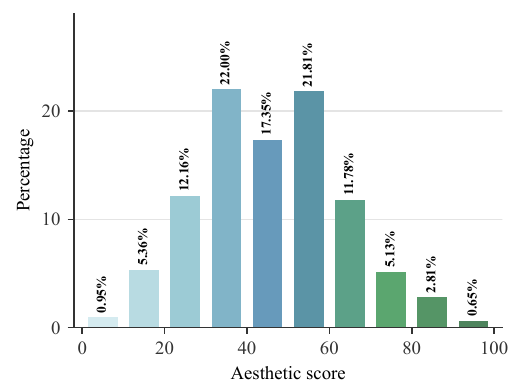}
    \caption{Aesthetic score distribution.}
  \end{subfigure}
  \caption{Resolution and internal aesthetic score distributions of the filtered 0.6B-image pool.}
  \label{fig:data-distributions}
\end{figure}

\textbf{Pool Refinement.}
Using this unified representation, we refine a raw pool of approximately 5 billion images collected from
internal data, Internet data, public datasets, and licensed datasets. The
pool is progressively refined through successive quality-control stages. After this process,
approximately 0.6 billion images are retained as the curated image pool. Both
the resolution and aesthetic statistics in Figure~\ref{fig:data-distributions} are
computed over this filtered 0.6B-image pool. Notably, 59.74\% of the retained
images fall into the 1024-or-higher resolution groups, providing a substantial
high-resolution foundation for subsequent training-data construction.

\textbf{Quality and Safety.}
The raw pool is processed by scalable operators for resolution thresholding, blur and clarity assessment, deduplication, and safety filtering, following quality-control dimensions established in large-scale image--text curation~\cite{schuhmann2022laion5b,gadre2023datacomp}. Low-resolution, blurry, duplicated, unsafe, and otherwise low-quality samples are removed, and pornographic or other sexually explicit NSFW content is excluded before downstream annotation and training.

\textbf{Aesthetic Scoring.}
For aesthetic quality assessment, we train an internal model inspired by Elo-style pairwise preference modeling, a formulation also used to learn preferences among text-to-image outputs~\cite{kirstain2023pickapic}. Pairwise human or model judgments are aggregated into relative rankings and distilled into a scoring model that captures our internal criteria for visually appealing and trainable images.
Figure~\ref{fig:data-distributions} jointly summarizes the resolution and
internal aesthetic distributions of the filtered pool. Most samples are
concentrated in the middle aesthetic score ranges, while the continuous score
provides a flexible signal for constructing mixtures with different quality
and coverage requirements.

% \begin{figure}[H]
%   \centering
%   \subcaptionbox{Resolution distribution.\label{fig:resolution-distribution}}[0.47\linewidth]{%
%     \includegraphics[width=\linewidth]{assets/resolution_distribution.pdf}}
%   \hfill
%   \subcaptionbox{Aesthetic score distribution.\label{fig:aes-distribution}}[0.47\linewidth]{%
%     \includegraphics[width=\linewidth]{assets/aes_score_distribution.pdf}}
%   \caption{Resolution and internal aesthetic score distributions of the filtered
%   0.6B-image pool. The aesthetic statistics are measured after applying the data
%   filtering operators.}
%   \label{fig:data-distributions}
% \end{figure}

\begin{figure}[H]
  \centering
  \begingroup
  \setlength{\tabcolsep}{1.2pt}
  \newcommand{\aesexample}[2]{%
    \includegraphics[
      width=0.218\linewidth,
      height=0.1453\linewidth,
      trim={#1},
      clip
    ]{#2}}
  \begin{tabular}{@{}cccc@{}}
    \textbf{0--25} & \textbf{25--50} &
    \textbf{50--75} & \textbf{75--100} \\[2pt]
    \aesexample{0 174 0 174}{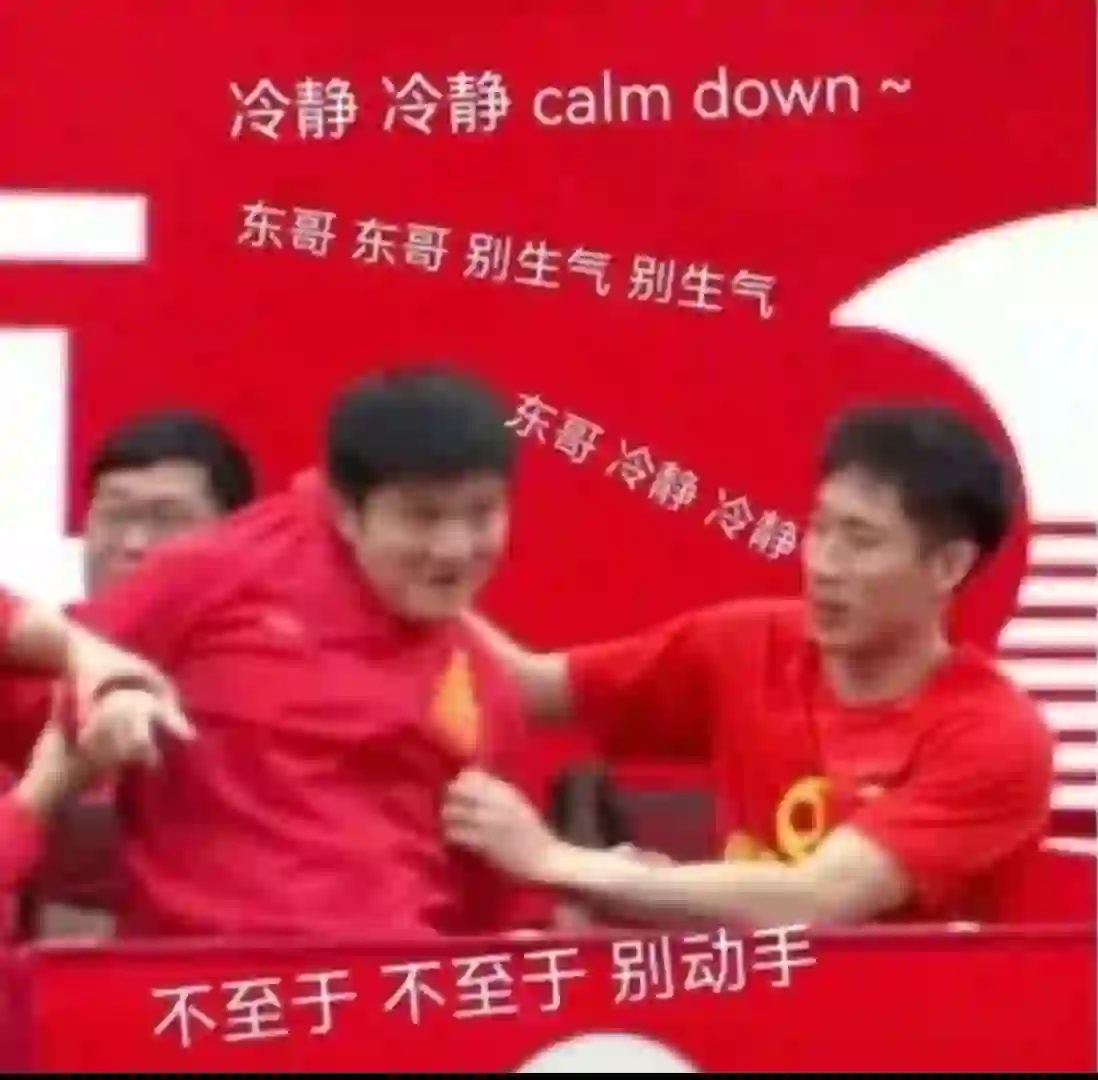} &
    \aesexample{0 0 0 0}{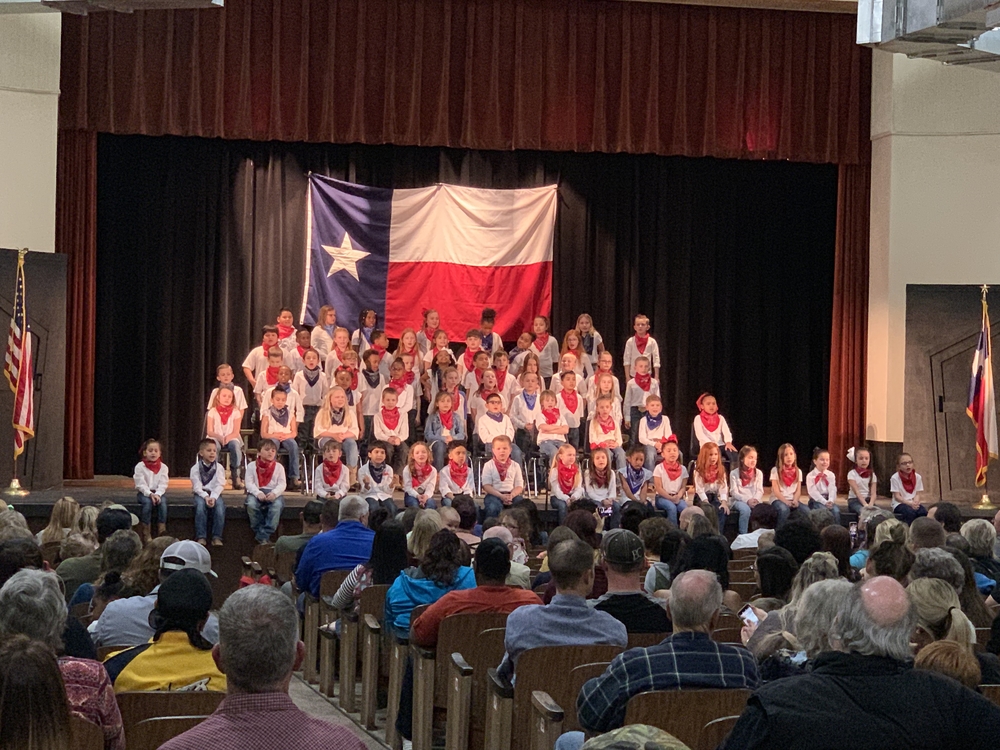} &
    \aesexample{0 0 0 0}{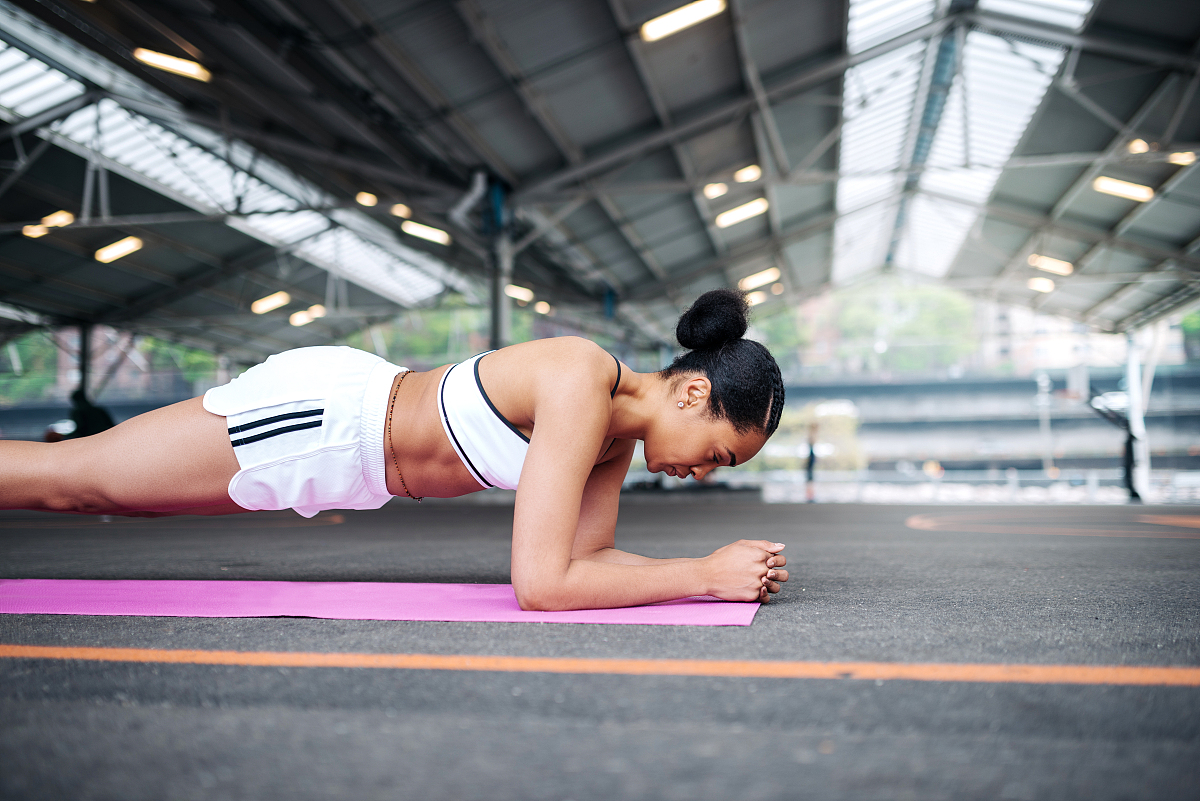} &
    \aesexample{0 106 0 107}{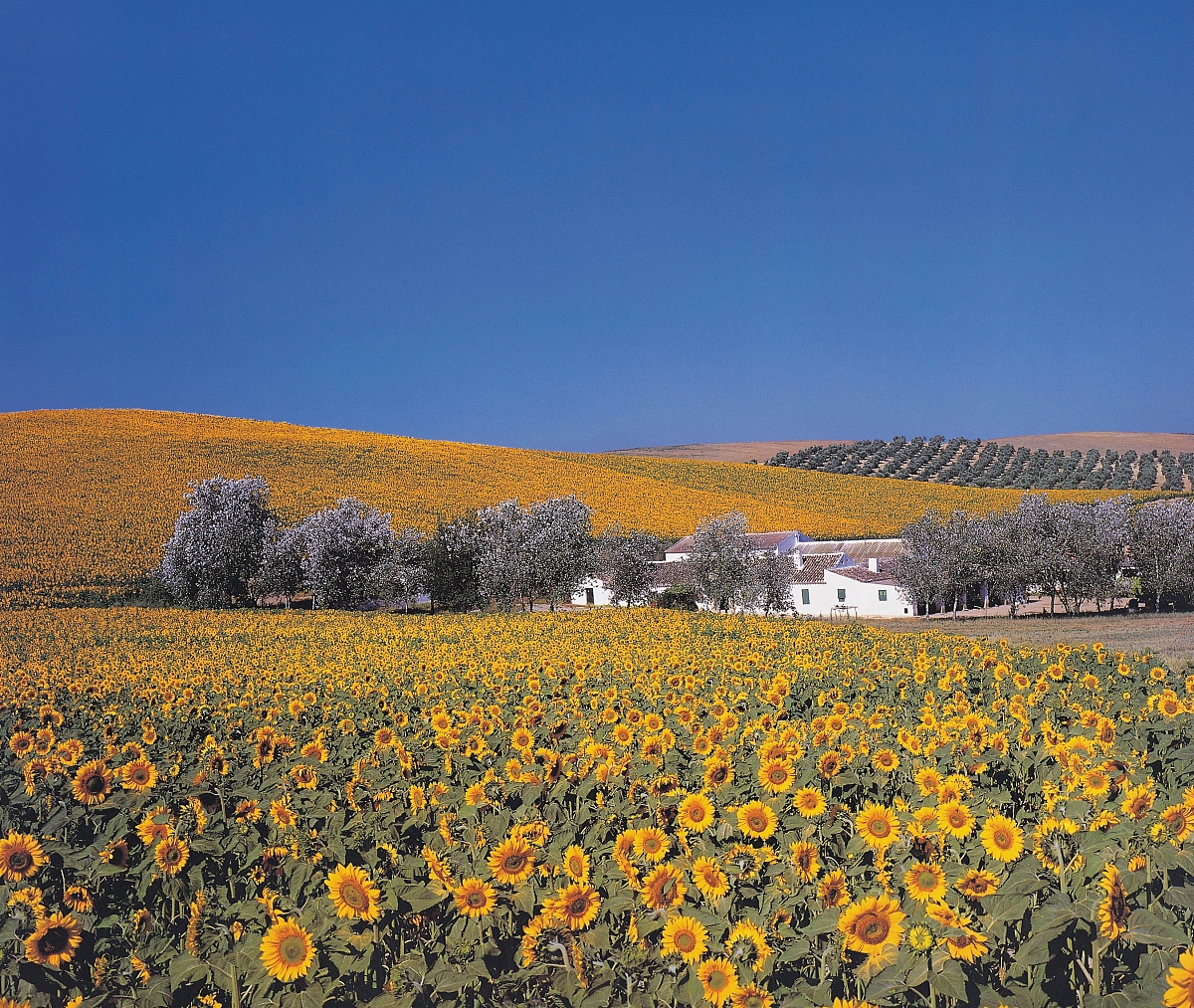} \\[2pt]
    \aesexample{0 33 0 34}{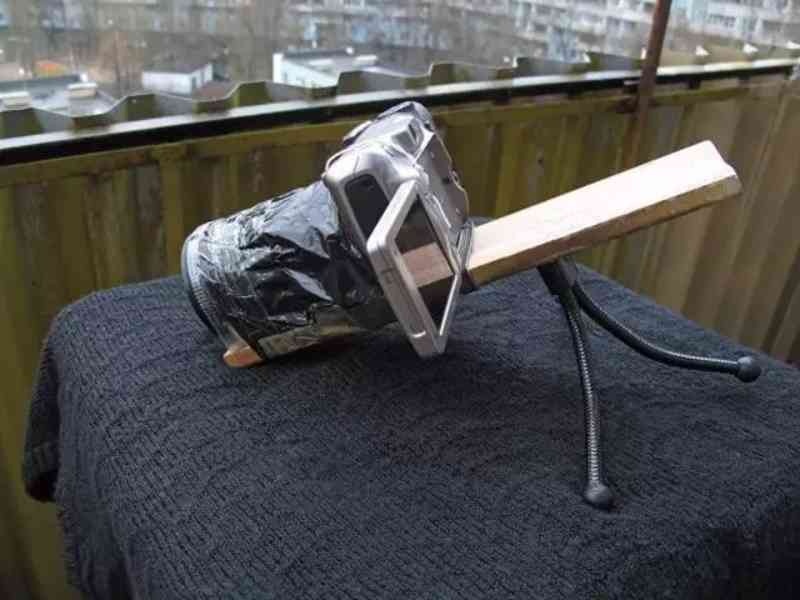} &
    \aesexample{0 0 0 0}{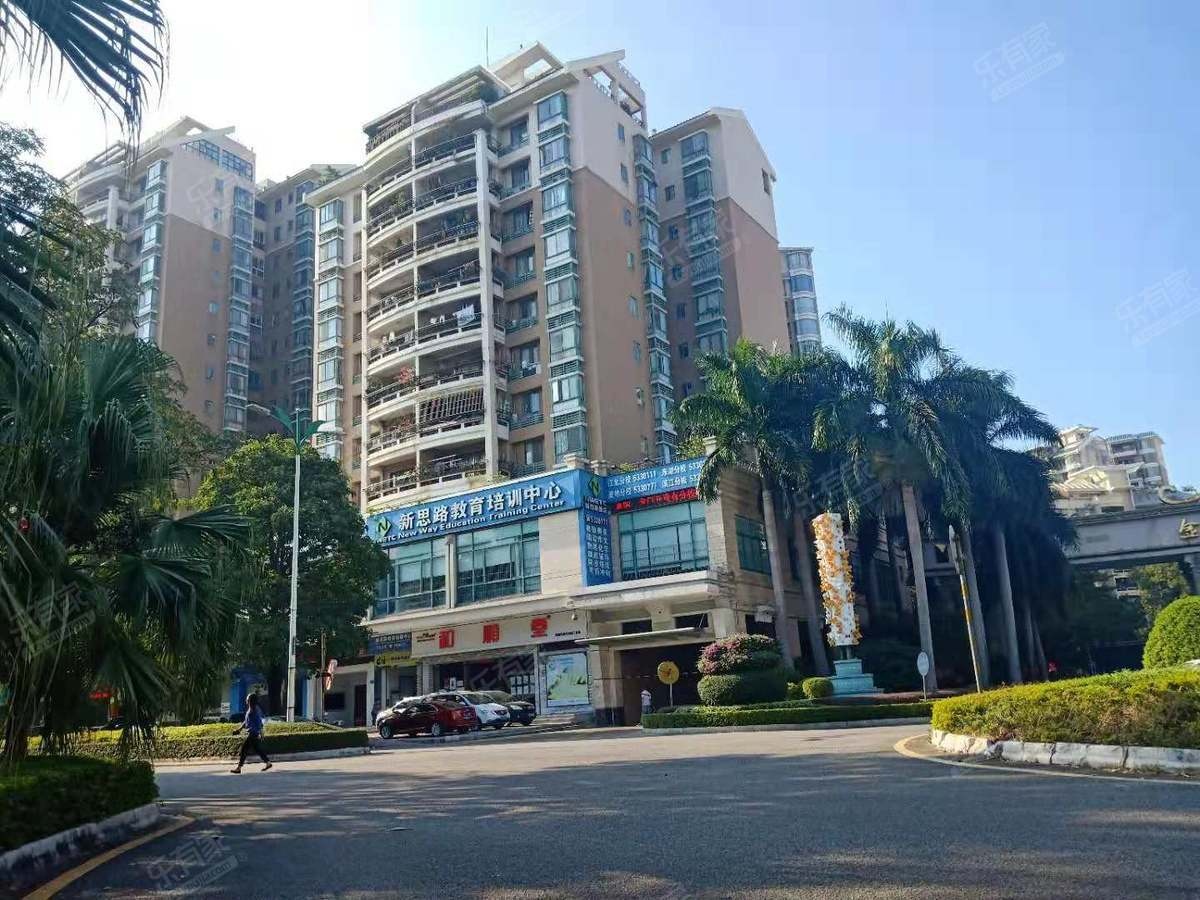} &
    \aesexample{0 35 0 36}{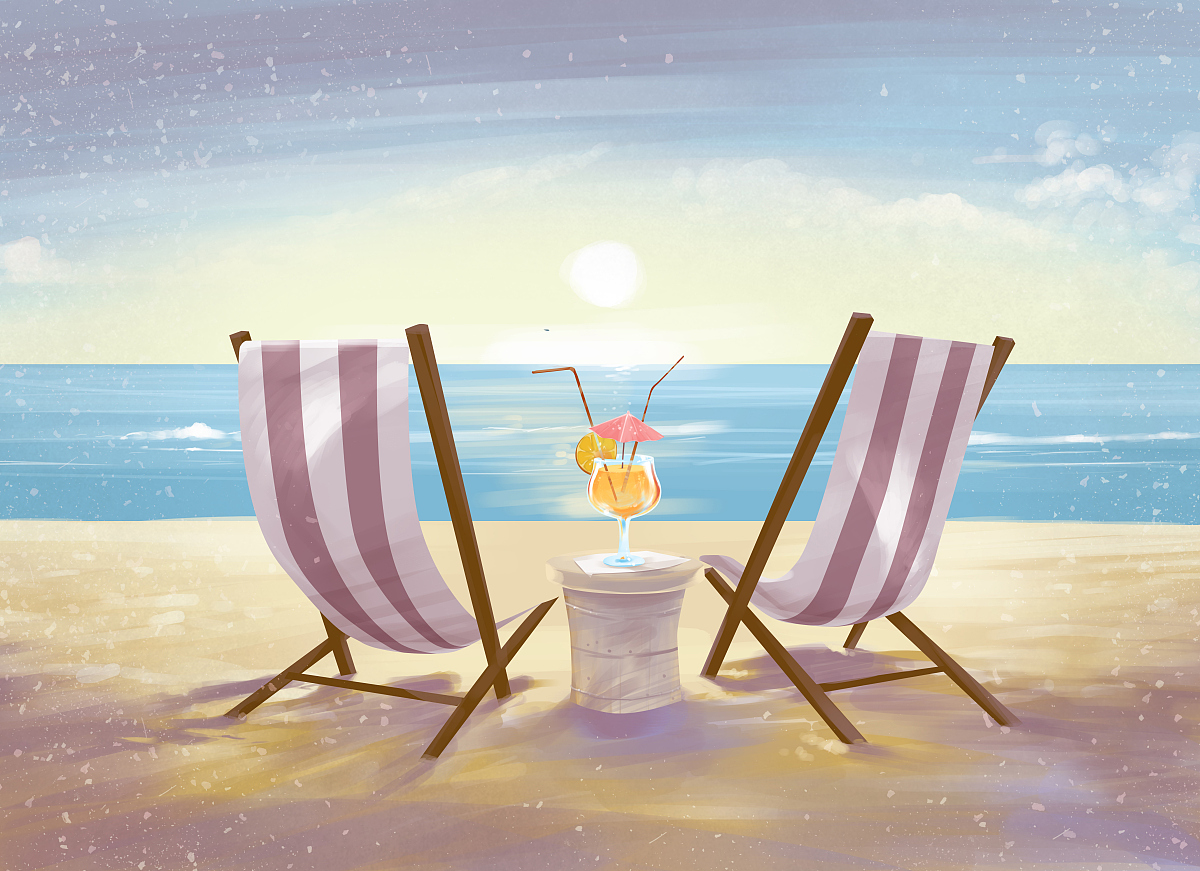} &
    \aesexample{0 27 0 27}{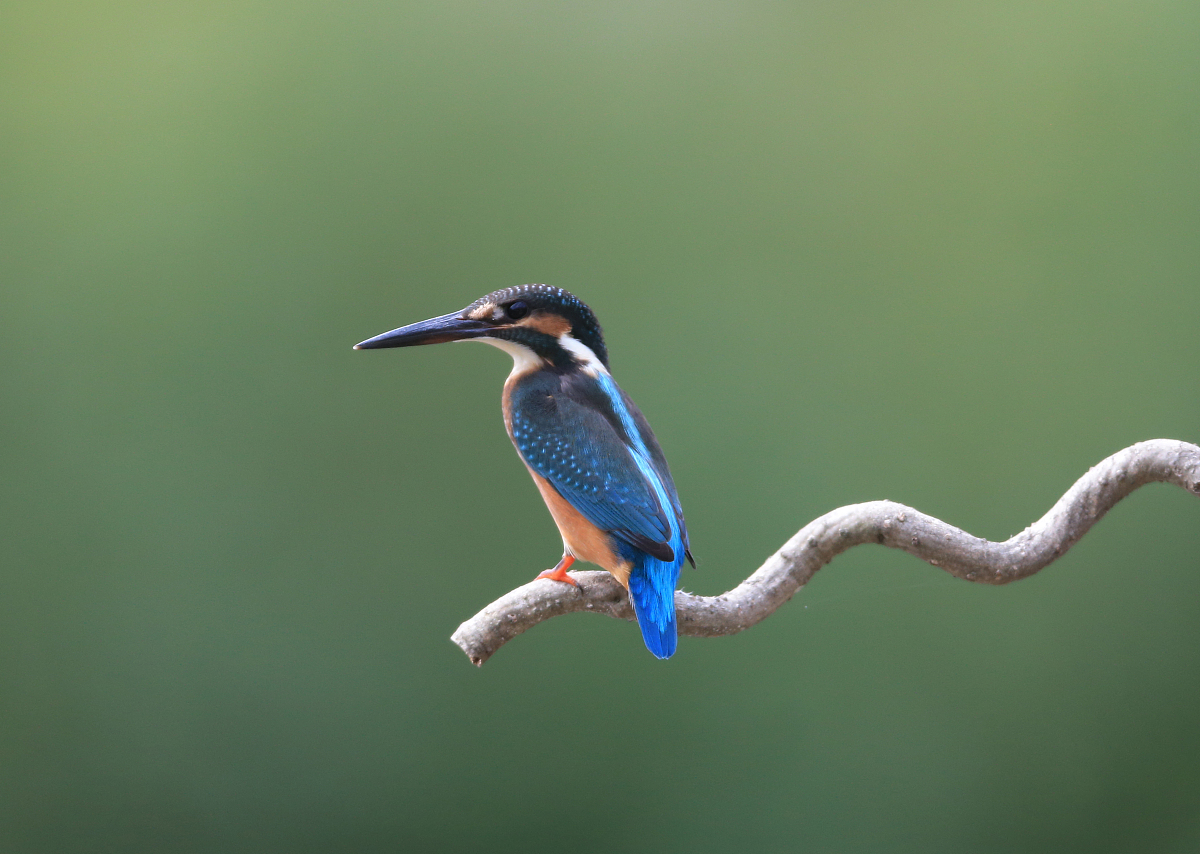} \\[2pt]
    \aesexample{84 0 84 0}{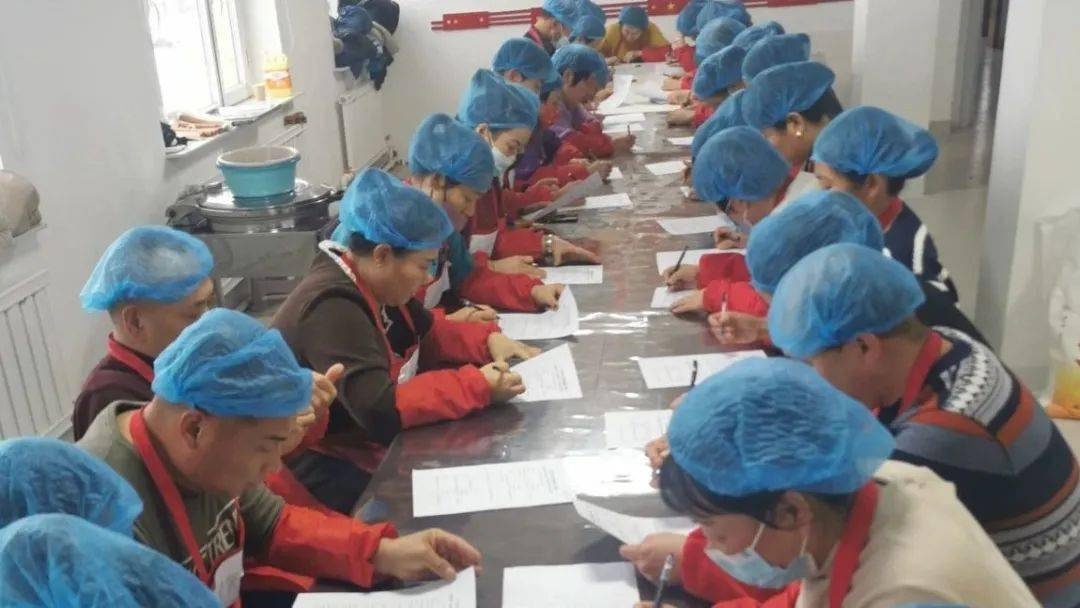} &
    \aesexample{0 50 0 50}{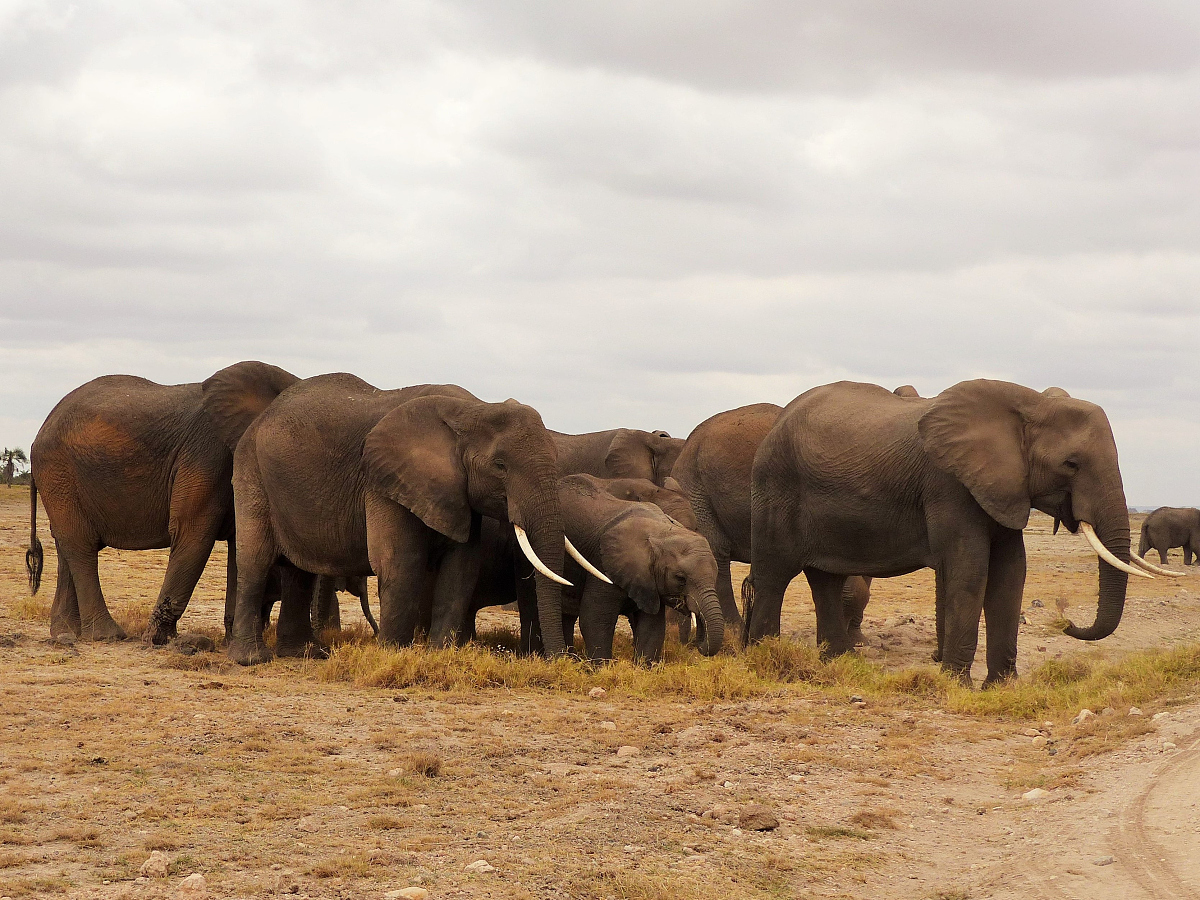} &
    \aesexample{0 28 0 29}{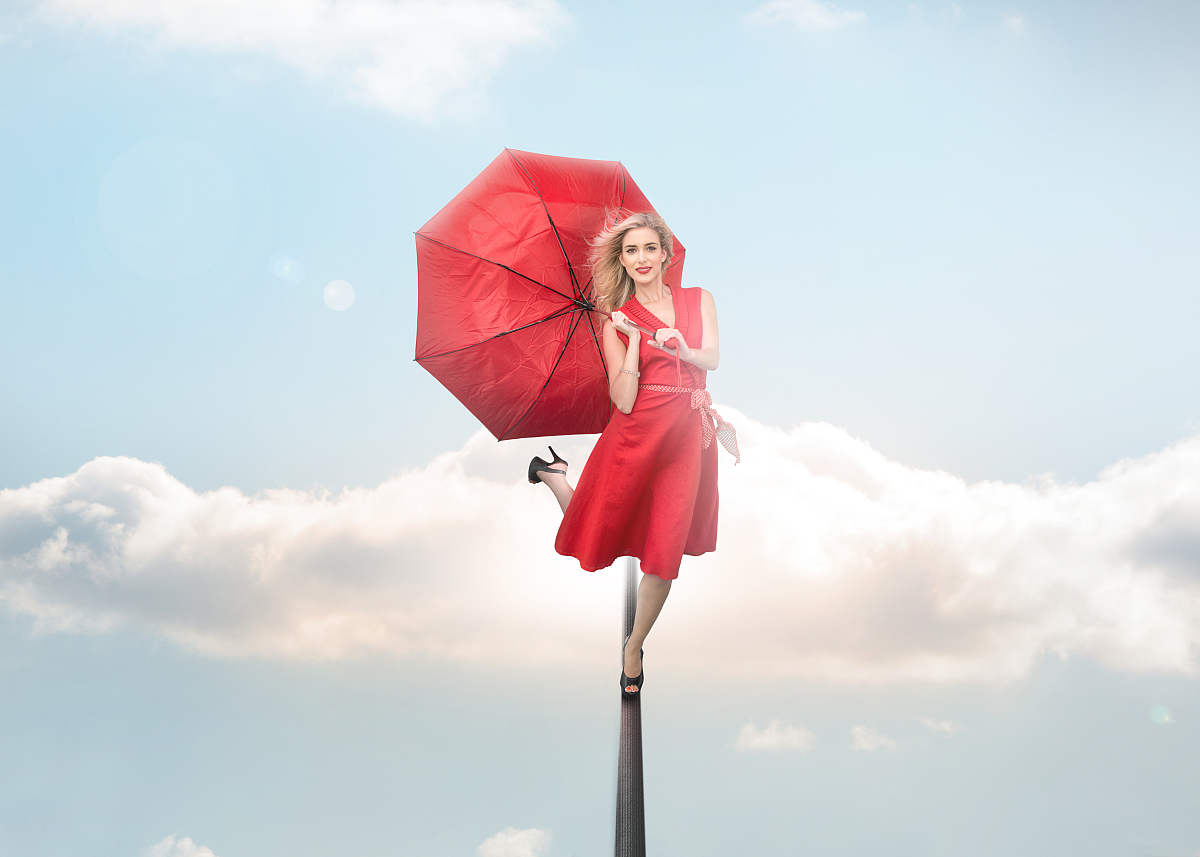} &
    \aesexample{0 0 0 1}{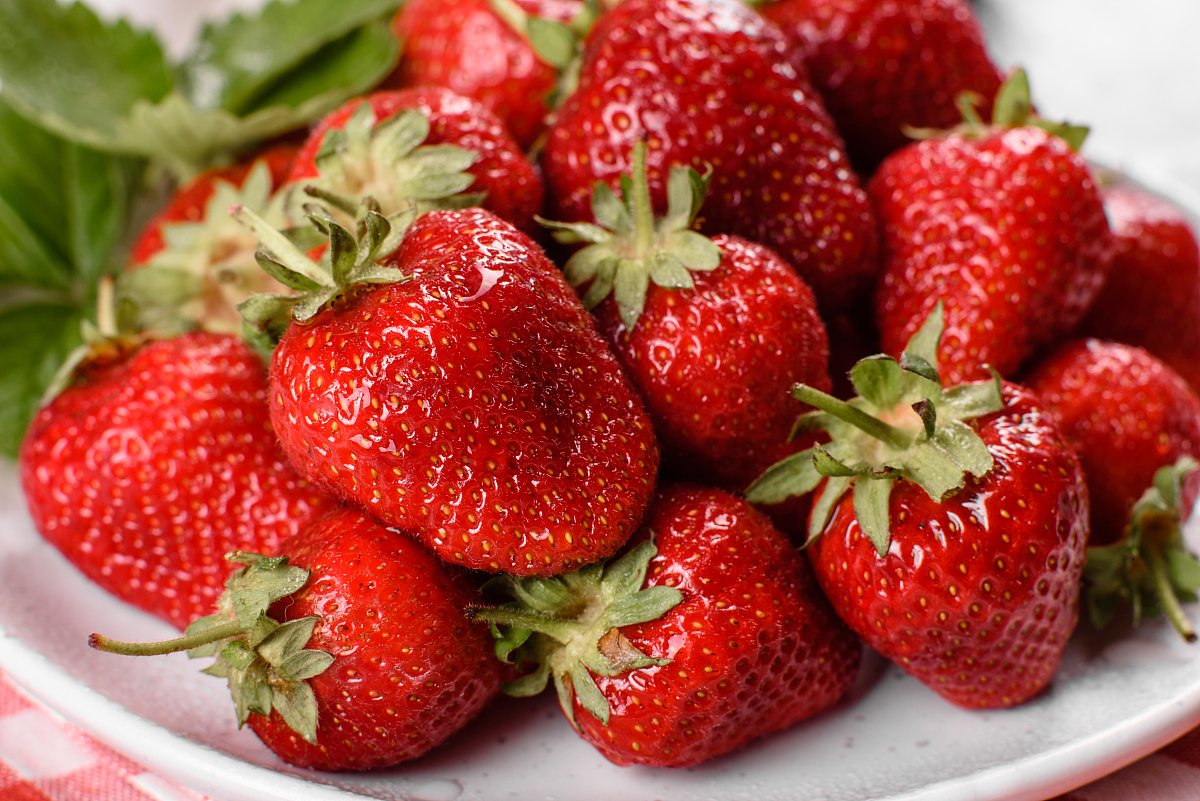} \\[2pt]
    \aesexample{0 200 0 200}{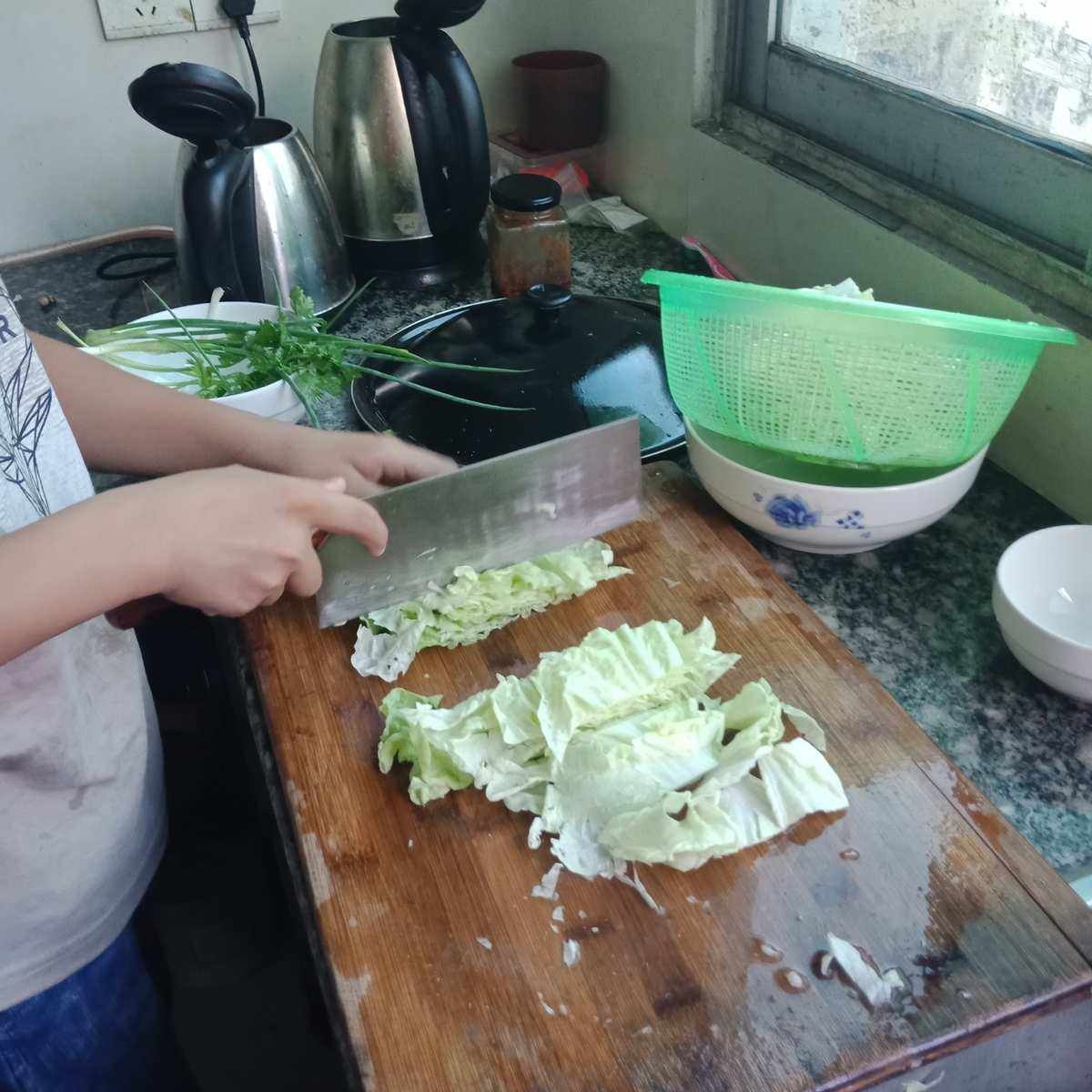} &
    \aesexample{0 400 0 400}{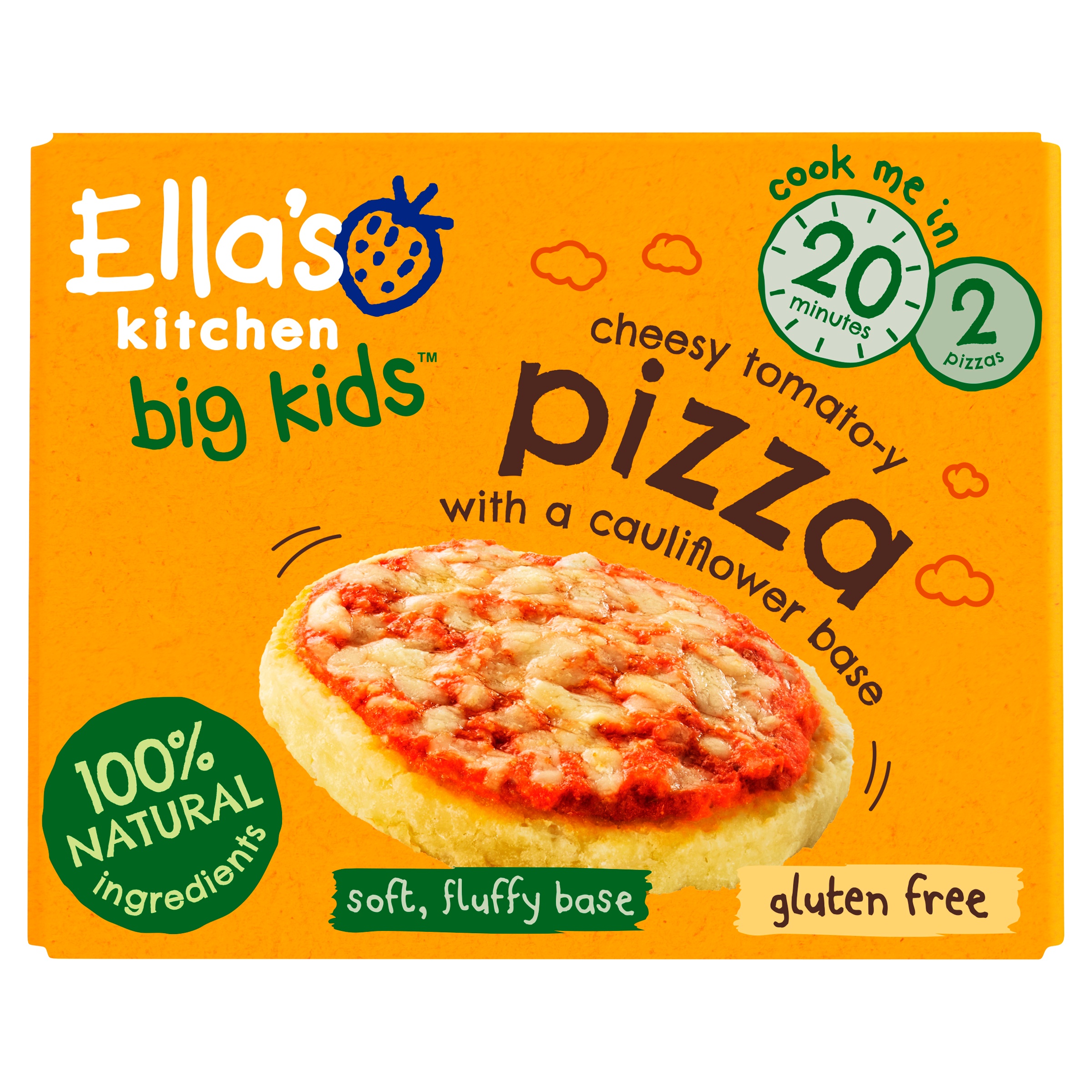} &
    \aesexample{0 0 0 0}{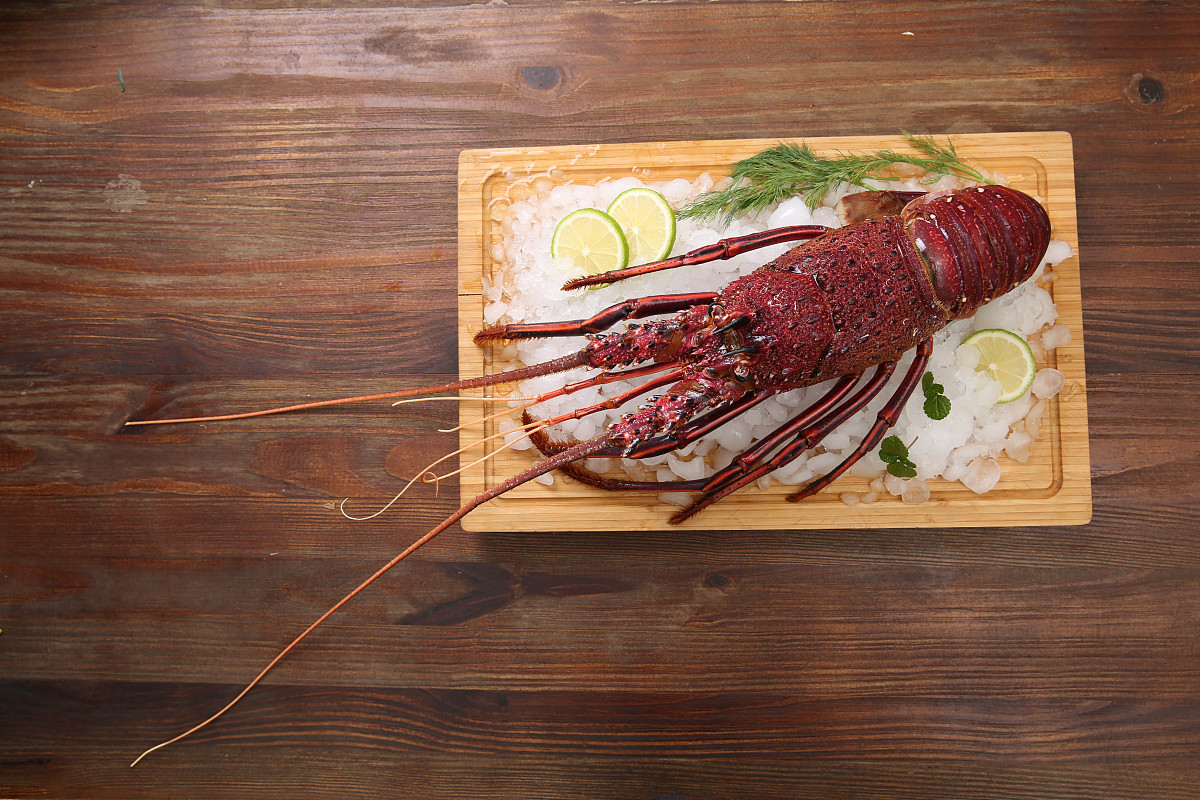} &
    \aesexample{0 0 0 0}{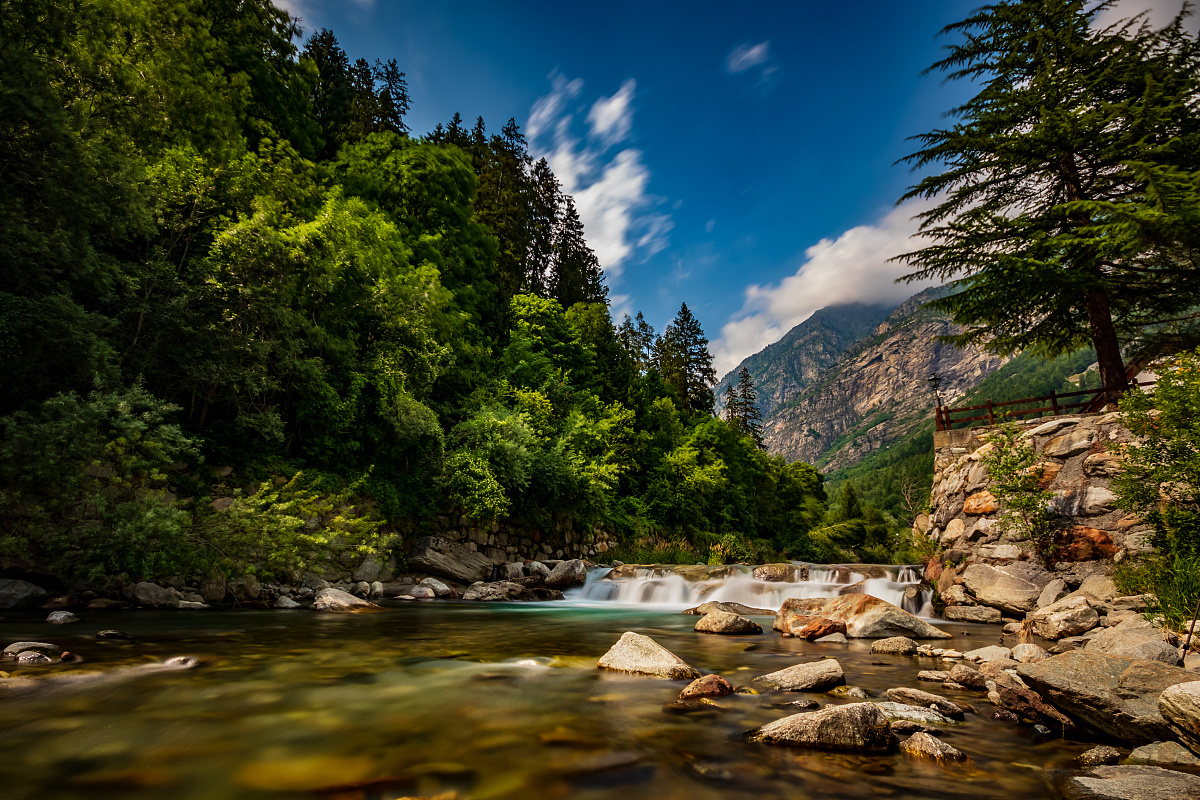} \\
  \end{tabular}
  \endgroup
  \caption{Representative samples from four AES score ranges. Images are
  center-cropped to a common aspect ratio for comparison. Each column
  contains four examples from the corresponding interval.}
  \label{fig:aes-examples}
\end{figure}

To make the aesthetic score qualitatively interpretable, we organize representative
images into four broad score ranges. Figure~\ref{fig:aes-examples} reserves
four examples for each interval, allowing the visual characteristics captured
by the aesthetic model to be inspected directly.

The filtering operators also act as attribute extractors rather than only as binary gates. Their fine-grained quality, resolution, safety, and content signals are retained for downstream sampling, routing, mixture construction, and distribution diagnosis.

\subsubsection{Visual Annotation}
\label{sec:data-annotation}

\textbf{VLM Annotation.}
With visual quality controlled, a large VLM produces structured annotations that describe the content of each retained image. These annotations provide a reusable semantic layer for retrieval, clustering analysis, knowledge-graph alignment, and later task-specific data construction.

\textbf{OCR and Watermarks.}
OCR and watermark operators identify text-heavy content, including screenshots, posters, advertisements, documents, UI layouts, and other special cases. The extracted signals preserve textual and layout information that can later be incorporated into task-specific captions instead of being lost during quality filtering.

\textbf{Semantic Tags.}
Tags capture entities, attributes, styles, scenes, and other compact visual semantics. Together with the structured VLM, OCR, and watermark outputs, they convert the curated image pool into quality-aware and semantically annotated assets from which downstream datasets can be retrieved, balanced, and constructed.

\subsection{Coverage and Balance}
\label{sec:data-organization}

After refinement and reusable annotation, we organize the curated pool along two complementary axes. The knowledge-grounded vocabulary makes named visual concepts explicitly addressable, while visual clustering captures coherent appearance patterns that may not yet have a stable semantic name. The two structures are maintained as a loop rather than as independent indexes: unnamed visual groups expose coverage gaps that can expand the vocabulary, and named concepts supply the semantic targets against which the clustered pool is retrieved and rebalanced.

\subsubsection{Visual Vocabulary}
\label{sec:data-kg}

\textbf{Concept Selection.}
To improve semantic coverage beyond the natural frequency of web data, we construct a knowledge-grounded visual vocabulary from both open and internal sources. We start from open lexical resources such as Cilin~\cite{mei1983tongyici} and filter them together with an internal WeChat Knowledge Graph (KG), retaining concepts that are visually meaningful for image generation. The resulting vocabulary contains both entities and attributes, covering objects, people, scenes, styles, materials, actions, relations, and other visually grounded concepts.

\textbf{Concept Cleaning.}
During vocabulary construction, we preserve the entity / attribute distinction while applying extensive cleaning to improve concept purity. LLM-based checks remove abstract, ambiguous, non-visual, or noisy items. We further use a Qwen3-embedding~\cite{li2026qwen3vlembedding} plus reranking pipeline to retrieve and verify near-synonyms, aliases, and visually similar expressions, allowing fragmented surface forms to be merged into cleaner concept nodes. This process turns a flat word list into a structured topology of concepts, synonym relations, and parent-child relations rather than a collection of isolated terms.

\textbf{Concept Hierarchy.}
To reduce the singularity of individual words, we build a complete hierarchical structure for the vocabulary. Each concept is placed in a broader semantic context, with upper-level categories supporting distribution analysis and lower-level nodes preserving fine-grained visual distinctions.

\textbf{Query Expansion.}
To quickly supplement data for any concept, we expand retrieval queries with aliases, descriptive phrases, visual keywords, and search-oriented expressions. The resulting vocabulary makes long-tail concepts addressable, retrievable, sampleable, and evaluable. As shown in Figure~\ref{fig:kg_schema}, each final node connects ontology, retrieval, and synonym knowledge in one structured representation.

\begin{figure}[t]
    \centering
    \begin{jsonbox}
    {
      "word": "canonical visual concept name used as the primary node key",
      "category_type": "entity or attribute",
      "ontology": {
        "level1": "coarse semantic category",
        "level2": "middle-level category",
        "level3": "fine-grained category",
        "description": "short visual definition"
      },
      "search_queries": [
        "expanded image retrieval query"
      ],
      "knowledge_graph": {
        "synonyms": [
          "verified near-synonym"
        ],
        "aliases": [
          "alternative name or common surface form"
        ]
      },
      "confusable_concepts": [
        "visually or semantically similar concept"
      ]
    }
    \end{jsonbox}
    \captionsetup{labelfont={bf,color=WeChatInk}}
    \caption{Structured representation of a final visual concept node.}
    \label{fig:kg_schema}
\end{figure}

\subsubsection{Visual Clustering}
\label{sec:data-clustering}

\textbf{Image Embeddings.}
In parallel with the knowledge vocabulary, we build a large-scale visual embedding space over the image pool. Images are encoded with Qwen3-VL-Embedding~\cite{li2026qwen3vlembedding}, and masks derived from filtering signals such as aesthetic scores select high-quality subsets for downstream clustering and analysis.

\textbf{Hierarchical Clustering.}
We apply a balanced hierarchical variant of spherical k-means~\cite{dhillon2001spherical} to cluster hundreds of millions of images into a large number of visually coherent groups. The resulting hierarchy provides a structured coverage map over the image distribution.

\textbf{Joint Coverage.}
Clustering is not used merely for indexing or compression. Together with the knowledge vocabulary, it organizes the same pool along complementary axes. The vocabulary makes named concepts addressable, retrievable, and quota-sampleable, so text-to-image construction can ground images in explicit semantic targets rather than in raw web frequency, as detailed in Section~\ref{sec:data-sampling}. Clusters provide the visual counterpart: they map unnamed but coherent appearance, make coverage gaps inspectable, and serve as routing units for downstream generation and editing. Representative images from each cluster can be classified by a VLM to estimate task suitability, which reduces the instability of per-image free-form construction and lets sampling and assignment operate on visually coherent groups rather than isolated samples.

Patterns that the vocabulary does not yet name become candidates for concept expansion, query refinement, and targeted supplementation, feeding newly discovered visual regularities back into the semantic scaffold. Coverage therefore comes from making both named semantics and unnamed appearance addressable, while balance comes from reweighting long-tail concepts through the vocabulary and routing visually coherent groups through clusters instead of sampling by natural frequency.

\subsection{Task Data}
\label{sec:data-task-construction}

The semantic vocabulary and visual clusters turn the curated pool into an organized source from which task-specific supervision can be constructed. We use this common foundation to build text-to-image, instruction-guided editing, multi-reference generation, and understanding-oriented datasets with task-appropriate sampling and verification.

\subsubsection{Text-to-Image}
\label{sec:data-sampling}

\textbf{Semantic Sampling and Rebalancing.}
Building on the complementary semantic and visual organization introduced in Section~\ref{sec:data-organization}, we use the KG vocabulary from Section~\ref{sec:data-kg} to define explicit semantic targets for text-to-image pretraining. Each KG node is encoded into a retrieval representation, and concept-to-image retrieval is used to recall candidate images from the large-scale image pool. To improve semantic precision, we further apply image-to-concept reverse verification: candidate images are matched back against the target concept vocabulary, and images with bidirectional semantic agreement are prioritized.

To convert this semantic coverage into a balanced training set, we assign per-concept quotas and select globally unique images for each KG word instead of sampling purely according to raw web frequency. This reduces repeated exposure to frequent concepts and gives long-tail entities and attributes greater training visibility. The resulting KG matching statistics, together with the cluster-side coverage map from Section~\ref{sec:data-clustering}, reveal which named concepts or coherent visual regions remain missing or under-covered. For these gaps, we use the expanded queries associated with each vocabulary node to retrieve and supplement additional data, prioritizing preferred sources when available and using fallback sources to maintain scale. This stage outputs sampled training rows, per-word statistics, unsatisfied concepts, and summary files, making it possible to distinguish well-covered concepts from those still limited by the underlying pool. Text-to-image data is therefore organized around semantic coverage and balance rather than raw frequency.

\textbf{Caption Construction.}
The sampled images are paired with captions designed along multiple axes, including language, length, semantic granularity, and style awareness. Randomized bilingual templates diversify their surface form across neutral descriptions, concise user-like expressions, and generation-style prompts while preserving the same underlying visual semantics. This design follows evidence that recaptioning noisy web pairs with richer synthetic descriptions can substantially improve prompt following~\cite{betker2023bettercaptions}.

For text-heavy images, posters, documents, UI screenshots, and other special layouts, a later rewrite stage incorporates the reusable OCR and watermark signals from Section~\ref{sec:data-annotation}. Captions are produced at different levels of detail so that visible text, layout, and ordinary visual content can be described faithfully without forcing every sample into a maximal description.

\subsubsection{Image Editing}
\label{sec:data-edit}

\textbf{Task-Balanced Assignment.}
Image editing data is constructed around a structured editing ontology. We define editing tasks using a hierarchy of coarse- and fine-grained categories, including object addition, object removal, background replacement, local attribute modification, portrait editing, product editing, restoration, and reasoning-oriented edits.

Recent general-purpose editing datasets and systems similarly organize supervision by operation type~\cite{liu2025step1x,ye2025imgedit,ye2026unicedit}. A key challenge is that free-form editing instruction generation naturally produces a long-tailed task distribution. Common operations such as removal, addition, background replacement, and global stylization tend to dominate, while more specific editing categories receive far fewer samples. To address this, we control task balance earlier, at the cluster assignment and sample selection stage, rather than relying on the prompt generator to choose tasks opportunistically.

\begin{figure}[t]
  \centering
  \includegraphics[width=\textwidth]{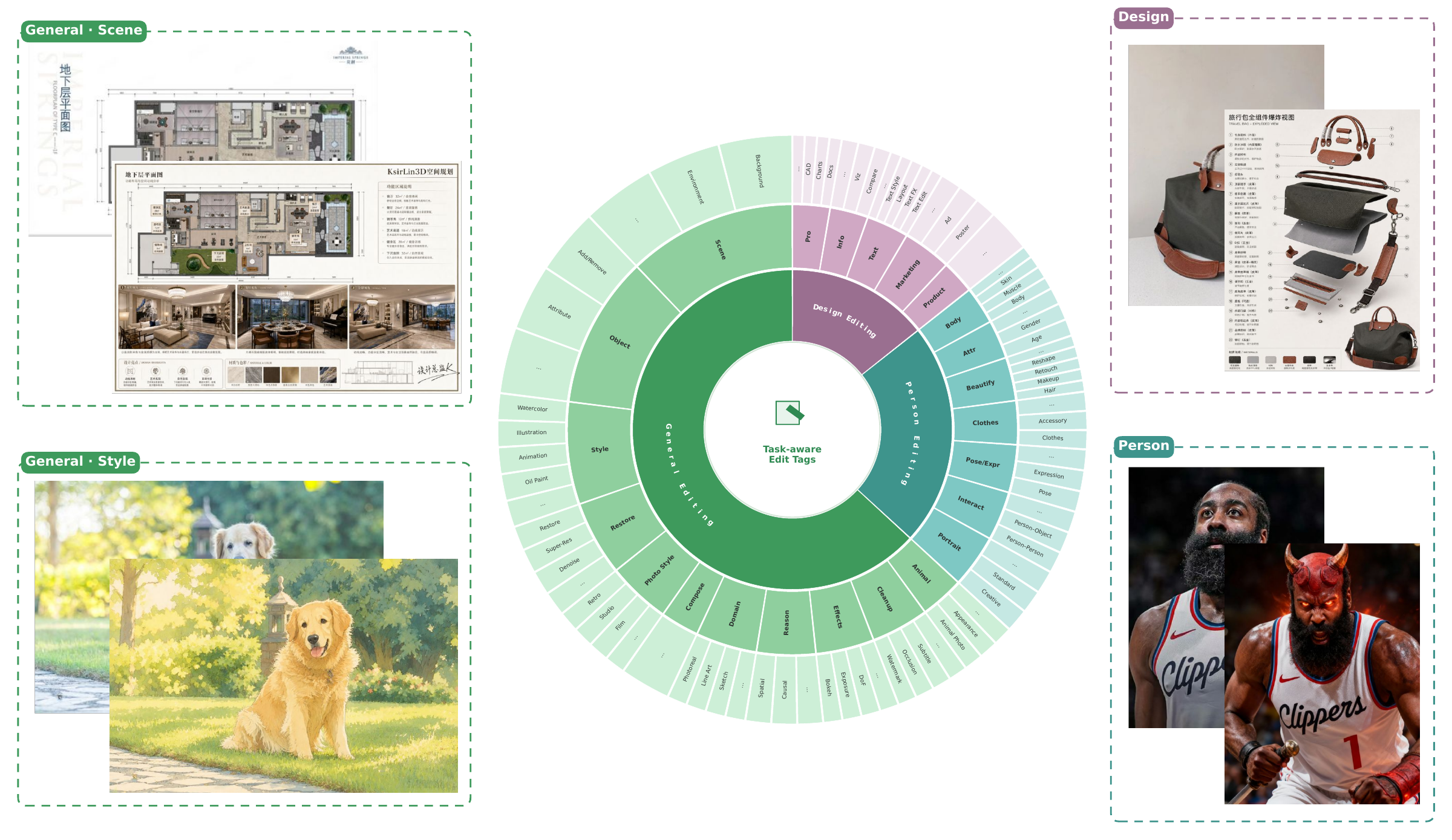}
  \caption{Illustration of task-aware edit tags used to guide editing-instruction
  construction.}
  \label{fig:edit_tag}
\end{figure}

We first classify visual clusters with a VLM, which predicts strong and possible editing-task affinities from representative images. These affinities form a candidate pool for a balanced cluster-to-task assignment algorithm that jointly considers task balance and visual suitability. Images sampled from the assigned clusters carry a fixed \textit{edit tag}, as shown in Figure~\ref{fig:edit_tag}, which acts as a hard prior so that instruction generation follows the assigned task rather than selecting an operation opportunistically.

\textbf{Pair Construction.}
After clusters are assigned to editing tasks, we construct editing pairs from two complementary sources. First, within a visually coherent cluster, we mine real-image pairs that share common transformation patterns, such as portrait changes, object changes, or viewpoint changes. Inspired by the graphical representation used in Z-Image~\cite{cai2025z}, we treat an image and its related variants as a small graph. Pairwise combinations can expand the number of training pairs, create mixed-editing supervision, and produce useful inverse pairs. This turns cluster-level visual regularities into reusable editing supervision instead of relying only on independent free-form generation.

Second, we decompose editing tasks into vertical categories and use different expert models for object-level, background-level, portrait-level, product-level, restoration, and reasoning-oriented edits. For many tasks, we construct pairs in a reverse direction: the target image is kept as a real image from the data pool, and the corresponding source image and instruction are derived around it. This reduces synthetic target artifacts and keeps the final target distribution grounded in real images.

\textbf{Verification and Prompt Diversification.}
All constructed pairs pass through a VLM-based quality gate. The checker verifies whether the source-target difference is valid, detects no-change cases, and fixes instructions that miss or misdescribe edit points.

Finally, we use the verified differences and refined instruction as the factual basis for prompt diversification. The same verified edit is rewritten into multiple semantically equivalent user prompts, covering concise commands, tag-like prompts, long narrative descriptions, spoken-style instructions, and noisy user-like phrasing. In this way, task-aware cluster assignment controls task balance, expert construction provides diverse pair sources, VLM checking guarantees correctness, and difference-grounded rewriting exposes the model to realistic user expressions while reducing semantic drift.

\subsubsection{Multi-Reference}
\label{sec:data-multi-image}

\textbf{Construction Paths.}
Single-image editing pairs cannot capture requests that require \emph{cross-image consistency}, such as preserving identity while transferring a garment, product, scene, or style from another image. Prior unified generators have likewise shown that mixed image--text conditioning can express subject-driven and visual-conditional generation tasks~\cite{xiao2025omnigen}. We therefore construct samples containing two to four source or reference images, an instruction grounded in their content, and one target image through two complementary paths. The reverse path retains a real target from the curated pool and derives suitable references and an instruction around it, reducing synthetic-target artifacts. The forward path instead assembles real source images and synthesizes the target with a multi-reference image editor.

\textbf{Role-Aware Assembly.}
Each source is assigned a role, such as identity, garment, object, scene, style, or layout reference. Combinations of these roles define 18 tasks, including co-shoot, virtual try-on, product placement, and reference-guided relighting, together with constraints on source diversity and compatibility. For the forward path, a VLM annotates cluster representatives to form an entity pool of approximately 750,000 images, and a quota-based assembler selects compatible but visually diverse source combinations from the embedding space. Hard constraints remove invalid combinations before a VLM writes an instruction that explicitly binds each Image~$k$ to the attributes to preserve or change, after which the editor synthesizes the target.

\textbf{Verification and Prompt Diversification.}
A VLM checker verifies reference roles, removes irrelevant inputs, renumbers image indices, and corrects inconsistent instructions. We retain facial and subject similarity, CLIP alignment~\cite{radford2021clip}, OCR, and aesthetic scores for diagnosis, but a controlled corruption study showed that thresholding these metrics rejects many valid samples, whereas a VLM judge detects nearly all introduced failures with a low false-rejection rate. Production filtering therefore enforces resolution and deduplication requirements, rejects source-target copies, and requires a positive VLM usability verdict; the remaining metrics are advisory. The verified differences and roles then support semantically equivalent prompt rewrites with varied lengths and styles while preserving the contribution of each reference. The resulting dataset contains several hundred thousand examples across the 18 tasks.

\subsubsection{Understanding Data}
\label{sec:data-umm}

\ModelName{} is designed around a unified model architecture. The current version still focuses primarily on image generation and image editing, but we believe the unified formulation has greater long-term potential. Adding understanding-oriented data can directly improve the model's ability to interpret prompts, reason over visual semantics, and produce better images.

\textbf{Text Data.}
We incorporate a proportion of text-only data to preserve the model's general language capabilities and its ability to interpret complex textual instructions.

\textbf{VLM Instructions.}
The current version primarily relies on open-source vision-language datasets, such as LLaVA-OneVision-1.5-Instruct~\cite{an2025llavaonevision15fullyopenframework}, FineVision~\cite{wiedmann2025finevisionopendataneed}, and Honey-15M~\cite{zhang2025beehighqualitycorpusfullstack}, together with proprietary captioning data. After dataset-level deduplication, cleaning, quality filtering, and training-recipe optimization, these sources are assembled into the vision-language instruction dataset.

\textbf{Task Taxonomy.}
We establish a unified labeling taxonomy covering General VQA, Caption \& Knowledge, Chart, OCR, Document, Grounding, Code, Math, Table, Science, Counting, GUI, Medical, Spatial, and Egocentric tasks. The filtering pipeline removes repetitive text, low-quality images, and images with extreme aspect ratios while mitigating language mixing and other sources of noise. Together, these procedures improve the quality, consistency, and diversity of understanding-oriented supervision.

\textbf{Reverse Alignment.}
In addition to curated text-only and vision-language data, a large amount of text-to-image data can naturally become vision-language alignment data after reverse construction. By treating generated or retrieved images as visual inputs and deriving corresponding descriptions, judgments, or instructions, the same data pool can support both generation training and image-text understanding, consistent with recent unified multimodal pretraining studies~\cite{deng2025bagel,chen2026booguimage01}. In the next version, we plan to further explore improvements to the model's visual understanding capability while keeping generation and editing as the main application targets.

\subsection{Training Mixtures and Feedback}
\label{sec:data-training}

After construction and task-specific verification within each data pipeline, the retained data is assembled into stage-dependent training mixtures and continually updated using feedback from model evaluation.

\subsubsection{Stage-Wise Mixtures}

The verified and rebalanced data is assembled according to the capability contributed by each source. High-quality text-to-image samples provide broad visual coverage, KG-guided samples improve long-tail semantic exposure, editing and multi-reference data strengthen controllable and compositional generation, and text-only and vision-language supervision improve prompt understanding and visual semantic parsing.

Mixture ratios are scheduled by training stage rather than held fixed throughout the pipeline. Early pretraining retains understanding-oriented data to establish the shared multimodal interface, whereas high-resolution continuation and SFT concentrate the budget on text-to-image, editing, and multi-reference generation, as detailed in Table~\ref{tab:training-recipe}.

\subsubsection{Feedback Loop}

Data construction is treated as a training-time feedback loop rather than a one-shot filtering pipeline. Evaluations run throughout training to reveal data issues, capability regressions, semantic coverage gaps, and task-distribution biases. The resulting diagnostics are returned to the data engine to adjust filtering thresholds, retrieval queries, sampling quotas, task assignments, and mixture ratios, connecting observed model weaknesses to the stage of data construction that can address them. As new weaknesses emerge, the verified mixture is updated for the next training iteration, so data composition remains an active optimization variable rather than a fixed recipe. Overall, filtering, knowledge- and cluster-based rebalancing, task-specific construction, and training-time feedback are successive stages of the same generation-oriented objective: turning web-scale data into a mixture from which controllable generation can actually be learned.

\section{Training}
\label{sec:training}

\begin{figure}[t]
  \centering
  \small
  \resizebox{0.68\linewidth}{!}{%
  \begin{tikzpicture}[
    stage/.style={
      draw=WeChatDeepGreen,
      fill=WeChatMint,
      rounded corners=2.2pt,
      minimum height=9.6mm,
      minimum width=2.05cm,
      align=center,
      inner xsep=3pt,
      font=\small\bfseries
    },
    mixstage/.style={
      stage,
      minimum height=1.42cm,
      minimum width=2.45cm,
      inner ysep=2.5pt
    },
    rlbox/.style={
      draw=WeChatDeepGreen,
      fill=white,
      rounded corners=2.2pt,
      minimum height=7.4mm,
      minimum width=1.85cm,
      align=center,
      font=\small\bfseries
    },
    arr/.style={-{Latex[length=2mm]}, thick, draw=WeChatGreen}
  ]
    \node[mixstage] (pt) at (0,0)
      {Pretraining\\[-1pt]{\normalfont\scriptsize $512{\to}1024{\to}1536$}\\[1pt]
      {\normalfont\scriptsize T2T, I2T, T2I, I2I}};
    \node[mixstage] (sft) at (3.55,0)
      {SFT\\[3pt]{\normalfont\scriptsize T2I, I2I}};
    \node[rlbox] (t2i) at (6.25,0.62) {T2I RL};
    \node[rlbox] (i2i) at (6.25,-0.62) {I2I RL};
    \node[stage] (opd) at (8.95,0) {OPD};
    \node[stage] (meanflow) at (11.55,0) {Meanflow};
    \draw[arr] (pt) -- (sft);
    \draw[arr] (sft.east) -- ++(0.22,0) |- (t2i.west);
    \draw[arr] (sft.east) -- ++(0.22,0) |- (i2i.west);
    \draw[arr] (t2i.east) -- ++(0.22,0) |- (opd.west);
    \draw[arr] (i2i.east) -- ++(0.22,0) |- (opd.west);
    \draw[arr] (opd) -- (meanflow);
  \end{tikzpicture}%
  }
  \caption{Training pipeline of \ModelName{}. Pretraining uses T2T, I2T, T2I,
  and I2I. SFT retains T2I and I2I. Task-separated RL then specializes the two
  generation tasks before OPD and Meanflow distill the policies for deployment.}
  \label{fig:training-pipeline}
\end{figure}
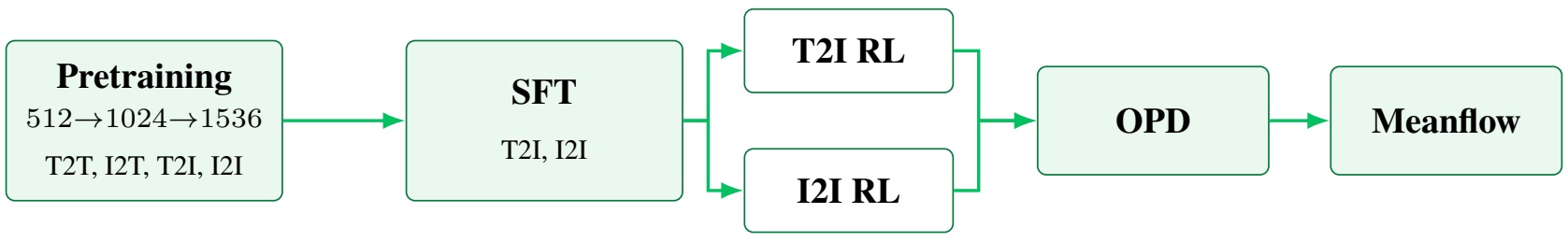

\ModelName{} is trained as one pipeline from progressive pretraining through
instruction SFT, task-separated reinforcement learning, and two distillation
stages, as shown in Figure~\ref{fig:training-pipeline}. Pretraining and SFT
install the dual-path backbone and raise resolution. T2I RL and I2I RL then
align generation and editing separately. On-policy distillation (OPD)
consolidates the resulting policies, and Meanflow-based distillation shortens the sampling trajectory for deployment. Table~\ref{tab:training-recipe}
records the pretraining and SFT recipe.

\subsection{Initialization}

Both paths of the dual-path backbone are initialized from the same pretrained
Kimi-VL~\cite{kimivl2025} language--vision checkpoint. Token embeddings and
the language-model head are loaded once and shared across the two paths. For
every transformer layer, the understanding-path
MLA projections, LayerNorms, and feed-forward weights---including the
dense first layer and the full MoE of routed experts, shared experts, and
the router---are taken from Kimi-VL, then replicated onto the generation
path so that both paths start from the same pretrained snapshot. The
final RMSNorm is copied in the same way. The ViT and its multimodal
projector are loaded from the same checkpoint. Only the
generation-specific VAE bridges and the timestep embedder start from
random initialization.

\subsection{Pretraining and SFT}

Table~\ref{tab:training-recipe} collects the four-stage recipe, including
optimization, resolution, parallelism, and the task mixture.

The schedule progressively increases synthesis resolution and context length
while shifting the data mixture toward generation. The 512 and 1024 stages
retain a modest amount of understanding data, including reverse image-to-text
examples formed from the same image--caption pairs used for generation: the
image becomes the input and its caption the prediction target. This reuse
provides complementary visual--semantic supervision without requiring a
separate data source. At 1536 resolution and during SFT, the mixture is
restricted to text-to-image, editing, and multi-reference generation.

\textbf{Stage 1: 512 pretraining.}
The first stage jointly trains understanding, text-to-image generation, and
editing at moderate resolution and context length. It establishes the shared
visual--semantic representation before higher-resolution continuation.

\textbf{Stage 2: 1024 continual pretraining.}
The second stage increases both synthesis resolution and context length,
adjusts the flow-matching timestep shift, and introduces multi-reference data.
Understanding supervision is retained to preserve cross-modal alignment during
this transition.

\textbf{Stage 3: 1536 continual pretraining.}
The third stage moves to the final pretraining resolution and a 64K-token
context. Understanding data is removed, and training focuses on text-to-image,
editing, and multi-reference examples. Context parallelism is enabled to
support the longer packed sequences.

\textbf{Stage 4: supervised fine-tuning.}
SFT retains the resolution and context settings of Stage~3 while lowering the
learning rate and using a curated instruction mixture of the same three
generation tasks. Caption lengths follow a long/medium/short ratio of
$4{:}3{:}3$ to expose the model to varied prompt styles. The resulting
checkpoint initializes the reinforcement-learning and distillation stages in
Figure~\ref{fig:training-pipeline}.

\begin{table}[t]
  \centering
  \fontsize{8pt}{10.5pt}\selectfont
  \renewcommand{\arraystretch}{1.22}
  \setlength{\tabcolsep}{2.5pt}
  \begin{tabularx}{0.86\linewidth}{@{}
      >{\raggedright\arraybackslash}p{0.21\linewidth}
      *{4}{>{\centering\arraybackslash}X}
      @{}}
    \toprule
    \textbf{Stage}
      & 512 & 1024 & 1536 & SFT \\
    \midrule
    \multicolumn{5}{@{}l}{\textbf{Optimization}} \\
    Optimizer
      & \multicolumn{4}{c}{AdamW~\cite{loshchilov2019adamw} ($\beta_1{=}0.9$, $\beta_2{=}0.95$,
        $\epsilon{=}10^{-15}$)} \\
    Learning rate
      & $10^{-4}$ & $10^{-4}$ & $10^{-4}$ & $2.5{\times}10^{-5}$ \\
    LR scheduler
      & Constant & Constant & Constant & Constant \\
    Warmup steps
      & $2{,}500$ & $500$ & $500$ & $500$ \\
    Weight decay
      & $0.01$ & $0.01$ & $0.01$ & $0.01$ \\
    Gradient clip
      & $1.0$ & $1.0$ & $1.0$ & $1.0$ \\
    EMA decay
      & $0.999$ & $0.999$ & $0.999$ & $0.995$ \\
    Loss (CE : MSE)
      & $0.25{:}1$ & $0.25{:}1$ & $0.25{:}1$ & $0.25{:}1$ \\
    Timestep shift
      & $1.0$ & $4.0$ & $4.0$ & $4.0$ \\
    \midrule
    \multicolumn{5}{@{}l}{\textbf{Resolution}} \\
    Context length
      & $16$K & $32$K & $64$K & $64$K \\
    ViT resolution
      & $224$--$560$ & $\leq 896$ & $\leq 1344$ & $\leq 1344$ \\
    Gen.\ pixels
      & $512^{2}$ & $1024^{2}$ & $1536^{2}$ & $1536^{2}$ \\
    Gen.\ long edge
      & $512$ & $1024$ & $2048$ & $2048$ \\
    \midrule
    \multicolumn{5}{@{}l}{\textbf{Parallelism}} \\
    FSDP shard
      & $32$ & $32$ & $16$ & $16$ \\
    Context parallel
      & $1$ & $1$ & $2$ & $2$ \\
    Expert parallel
      & $1$ & $1$ & $1$ & $1$ \\
    Activation checkpoint
      & $22$ & $22$ & $27$ & $27$ \\
    \midrule
    \multicolumn{5}{@{}l}{\textbf{Data mixture}} \\
    Understanding
      & $20\%$ & $20\%$ & -- & -- \\
    Text-to-image
      & $50\%$ & $51\%$ & $67\%$ & $60\%$ \\
    Editing
      & $30\%$ & $25\%$ & $25\%$ & $30\%$ \\
    Multi-reference
      & -- & $4\%$ & $8\%$ & $10\%$ \\
    \bottomrule
  \end{tabularx}
  \caption{Training recipe of \ModelName{}. Generation pixels denotes the
  native-image area budget, and generation long edge gives the additional side cap.
  The product of FSDP shard and context-parallel size is held at $32$ so that
  the parameter-shard width is unchanged when long-context training is
  enabled. Mixture percentages are normalized over understanding,
  text-to-image, editing, and multi-reference data.}
  \label{tab:training-recipe}
\end{table}

\subsection{Post-training}
\label{sec:posttraining}

After SFT, we optimize text-to-image generation (T2I) and image-to-image editing
(I2I) as two separate reinforcement-learning branches initialized from the same
checkpoint, with independent training data, rewards, rollouts, and LoRA
adapters~\cite{hu2022lora}. In both branches, the pretrained backbone stays frozen and training
updates only generation-side LoRA modules. In generation attention, LoRA is
applied to the query, key--value, and output projections. In the dense and
shared-expert feed-forward networks, it is applied to the gate, up, and down
projections. The routed generation experts receive LoRA on the fused gate--up
and down projections. Our ablations on the
adapter scope show that jointly applying LoRA to the MoE layers consistently
improves performance for both T2I generation and image editing. The
understanding pathway, embeddings, normalization layers, and MoE router remain
frozen. Each branch is
optimized with DiffusionNFT~\cite{zheng2026diffusionnft}, which converts
reward-scored rollouts from an EMA policy into optimality supervision for a
forward-process objective over implicit positive and negative velocity fields.
After the two RL branches converge, on-policy distillation (OPD) consolidates
their capabilities into a single student checkpoint, eliminating the need to
switch between separate T2I and I2I adapters at inference and preparing the
merged policy for subsequent deployment distillation.

\label{subsec:rl-posttraining}

% Included from sections/05_training.tex under Post-training.

\subsubsection{Text-to-Image Reinforcement Learning}

For text-to-image generation, four independently trained VLM rewards measure image-text alignment, aesthetics, text rendering, and visual naturalness. We use GDPO~\cite{liu2026gdpo} to normalize the heterogeneous reward dimensions separately within each rollout group before aggregation.

\textbf{Image-text alignment reward.}
Rather than regressing a single opaque alignment score, we adopt a rubric-based
reward in the spirit of RubricRL~\cite{feng2025rubricrl}. Before rollout, each
training prompt is decomposed into a structured rubric, that is, a set of atomic
and visually verifiable criteria, and the reward is obtained by judging the
candidate image against these criteria and aggregating the item-level outcomes.
Rubric construction is deliberately image-blind, and the resulting rubric is
reused for every rollout of the prompt, so that all candidates within a group are
scored under an identical standard.

Rubric generation is anchored on a curated criterion pool rather than produced
free-form. The pool is organized around the dimensions of image--text
consistency that matter in practice, including object presence and count,
rendered text, attributes, spatial relations, actions, negation, and style or
layout constraints. Given a prompt, the generator identifies the most relevant
dimensions, instantiates criteria accordingly, applies prompt-adaptive weighting
that emphasizes the salient dimensions, and caps the total number of criteria per
prompt. Each prompt and each criterion carries a dimension tag. Because the
entire construction pipeline runs offline, these tags enable category-aware
sampling and allow us to systematically resample the offline training corpus
toward failure modes observed in the current model. Since pass rates are tracked
separately per dimension, gains in frequent categories cannot mask persistent
failures in long-tail ones.

During RL training, we present all rollouts of the same prompt to the judge as a
group rather than scoring each candidate independently. Group-wise scoring
provides a shared comparison context, reduces cross-call calibration variance,
and makes fine-grained differences in rubric satisfaction easier to distinguish.
The judge returns an item-level decision with supporting evidence for every
criterion and rollout, and we aggregate these decisions according to item
assessability and importance. This itemized formulation yields both a scalar
training signal and a fine-grained failure diagnosis.

\textbf{Aesthetic reward.} We first train a pairwise VLM to compare two same-prompt images in terms of
composition, lighting, color harmony, texture, and overall visual appeal. Unlike
scalar preference models such as PickScore, which learn a single score from
pairwise choices~\cite{kirstain2023pickapic}, our pairwise model learns from
structured comparison supervision comprising the preference verdict, visual
evidence, per-image dimension scores, and dimension-importance weights. We then
apply this teacher to multiple same-prompt opponents and aggregate its pairwise
outputs into a stable absolute target for each image. Distilling these
multi-opponent targets into a pointwise reward model preserves the reliability
and interpretability of comparative reasoning while allowing each rollout image
to be scored independently with a single forward pass during RL.

As a dedicated component of the aesthetic reward, we additionally train a
visual-deformation detector. General-purpose VLMs remain relatively weak at
recognizing malformed structures directly. We find, however, that conditioning
a VLM on the candidate image together with a coarse deformation category elicits
high-confidence chain-of-thought rationales that localize and explain the
corresponding defect. These category-conditioned rationales provide dense
supervision beyond the original coarse labels, allowing us to train an accurate
deformation reward model with only a modest amount of annotated data.

\textbf{Text-rendering reward.}
All prompts are pre-annotated according to whether rendered text may reasonably
appear in the image. Under an elastic text constraint, rendered text is
optional: if text appears, the evaluator checks whether it is semantically
incorrect, structurally anomalous, illegible, or composed of pseudo-characters,
while correct and legible text incurs no penalty. If a prompt is labeled as
incompatible with rendered text, any visible text is penalized. Because
open-source VLMs remain relatively weak at perceiving fine-grained glyph-level
structural anomalies, particularly in rendered Chinese characters, we train a
dedicated structural-anomaly perception model. It robustly detects glyph
deformation, missing or redundant strokes, blurriness, and misalignment in both
Chinese and English across challenging settings, including dense text, small
text, rare Chinese characters, and artistic typography. The resulting
structured judgments are aggregated into the text-rendering reward.

\textbf{Visual-naturalness reward.} The fourth VLM directly evaluates visual naturalness, focusing on recognizable synthetic-image cues such as plastic or over-smoothed textures, repeated micro-patterns, incoherent local details, and implausibly uniform lighting. This reward encourages natural textures, coherent local details, and physically plausible illumination.

% \paragraph{Reward-specific benchmarks and closed-loop iteration.} We build a versioned internal Bench for every reward dimension. The image-text alignment Bench measures checklist decisions and reports category-level accuracy; the aesthetic Bench evaluates pairwise preference agreement; the text Bench covers text-permissive and text-forbidden prompts as well as erroneous, pseudo-character, or garbled text; and the visual-naturalness Bench evaluates recognizable synthetic-image cues. Together, these Benches connect reward construction, reward-model training, policy optimization, effect evaluation, failure attribution, and data or strategy updates in one iterative loop.

\textbf{Prompt curation and category-aware reward routing.}
We construct the T2I RL prompt pool with balanced sampling across semantic
categories and short-, medium-, and long-form prompts. Before online training,
we perform offline rollouts with the base policy and estimate the mean and
variance of the composite reward for each prompt. We retain prompts whose mean
reward lies within a target range and whose reward variance exceeds a predefined
threshold. This removes prompts that are either already solved or consistently
fail across rollouts, concentrating RL updates on learnable cases that provide
informative within-group contrasts. We further assign each prompt one or more
category labels and use them to determine its reward-weight vector offline. For
example, typography prompts place greater emphasis on text rendering, whereas
portrait prompts assign more weight to aesthetic and deformation-related
signals. This offline routing aligns the optimization objective with the quality
requirements of each category while avoiding a single reward dimension
dominating the heterogeneous training mixture.

\textbf{Mixed original/PE rollout groups.} Prompt enhancement (PE) is applied inside the rollout group rather than as a separate training stream. For every canonical prompt, part of the group is generated from the original prompt and the remainder from its semantically constrained enhanced form. Mixing the two prompt forms in one relative group broadens visual coverage, increases within-group reward variance, and provides more informative positive and negative samples than repeatedly sampling only one formulation.

\subsubsection{Image Editing Reinforcement Learning}

\textbf{Training data and action balancing.}
We categorize image editing instructions into 27 editing action types and use this taxonomy to filter the training data. First, we use multiple open-source reward models to evaluate and rank the difficulty of the instructions offline. During instruction sampling, we set the target ratio of easy, medium, and hard samples to $3\!:\!5\!:\!2$. Within each editing action category, we further exclude samples at the extremes of the difficulty spectrum, including those with near-zero success rates (excessively difficult) and those with near-saturated success rates (overly easy). This filtering retains more informative samples that provide stronger learning signals for policy optimization.

% We organize editing instructions into 27 action categories and use this taxonomy
% for reward-aware data curation. We first rank the categories by their mean
% rewards over four-seed offline rollouts, treating the lowest 18 categories as
% the lower-reward group and the highest 9 as the higher-reward group. We allocate
% 70\% of the filtered training-set budget to the lower-reward group and 30\% to
% the higher-reward group, using approximately uniform category quotas within
% each group. Within each action category, we retain prompts whose per-prompt mean
% rollout reward falls between the 40th and 70th percentiles, avoiding both nearly
% impossible and already saturated cases. During online training, we preserve the
% action marginal of the filtered dataset without additional action-level
% weighting, while sampling instruction-level difficulty with an
% easy/medium/hard target ratio of \(3{:}5{:}2\).

\textbf{Checklist-based editing reward.}
We propose a category-aware, hierarchical checklist-based reward for evaluating image editing. Before training, an offline checklist generator built on a VLM constructs an instruction-specific evaluation checklist. The generator conditions on the corresponding action label as task-level context, enabling it to define evaluation dimensions and criteria tailored to the specific editing category. During training, each instruction consistently reuses its precomputed checklist, ensuring consistent reward evaluation. Each checklist covers three dimensions: prompt following, consistency, and visual
quality. The VLM-based evaluator
assigns each criterion \(i\) a status
\(o_i\in\{\mathrm{pass},\mathrm{fail},\mathrm{not\mbox{-}assessable}\}\) for each candidate edited image.
Failed criteria additionally receive a mild, medium, or severe failure level.
Within each dimension, failed criteria incur penalties determined jointly by
criterion importance and failure severity. We subtract the accumulated penalties
from a fixed score budget and normalize the result to obtain
\(r_P,r_C,r_Q\in[0,1]\) for prompt following, consistency, and visual quality,
respectively. Pass and not-assessable outcomes incur no penalty.
We use prompt following as a multiplicative gate:
\begin{equation}
  r_{\mathrm{edit}}
  =
  r_P\left(0.4+0.4r_C+0.2r_Q\right).
  \label{eq:edit-raw-reward}
\end{equation}
This prevents high consistency or visual quality from compensating for failure
to perform the editing instruction.

% For each editing instruction, an offline VLM-based checklist generator constructs
% an instruction-specific checklist before rollout and stores it in the training
% metadata. The action label provides high-level task context but does not
% prescribe a checklist shared by all instructions in the category. The same
% checklist is reused for all rollouts of the instruction.

% For each instruction, we generate \(K=12\) candidates and normalize the scalar
% \(r_{\mathrm{edit}}\), rather than its three components independently, using
% cumulative running per-prompt statistics with \(\epsilon=10^{-4}\). Following
% DiffusionNFT, we clip the normalized score to \([-3,3]\) and linearly map it to
% the positive- and negative-branch mixing coefficients.

\textbf{Editing-specific training strategies.} We use moderate classifier-free guidance (CFG)~\cite{ho2022classifierfree} to rollout samples while excluding the unconditional branch from the training objective, thereby improving the reliability of rewards and training stability. To accelerate sampling, we use half the standard number of inference steps during rollout, while retaining the complete trajectories for training. In contrast to RL for T2I task, we find that training on complete trajectories yields improved performance for image editing tasks.
% \paragraph{Editing-specific training strategies.}
% \begin{itemize}
%   \item \textbf{Hybrid CFG.}
%   We generate rollout candidates using moderate CFG, while training objective excludes unconditional branches, which ensures the reliability of the reward and the stability of training.

%   \item \textbf{Full noise-level training window.}
%   Each rollout uses 20 sampling points, corresponding to 19 flow transitions.
%   We retain all 19 noise levels for DiffusionNFT optimization. At each level,
%   the rollout endpoint is independently forward-noised to construct the NFT
%   training state, and the losses across all levels are averaged before a single
%   optimizer update. These states cover all noise levels of the rollout schedule,
%   but are constructed by independently forward-noising the rollout endpoint
%   rather than reusing the intermediate latent states of the sampled trajectory.
% \end{itemize}

\subsubsection{On-policy Distillation}

Applying RL independently to T2I task and image editing task yields strong task-specific experts (Teacher), while training on mixed-task data provides balanced generalization but may not reach the performance ceiling of each domain. We therefore employ On-policy Distillation (OPD)~\cite{zhou2026danceopd,li2026rethinking} to integrate the specialized capabilities of the two experts into a unified model. 

% We initialize the student model based on an SFT model and route it to the corresponding T2I or editing RL teacher model based on the task category of the sampled data. We retain CFG when generating student rollout trajectories to consolidate the expert capabilities. During training, we query both the conditional and unconditional branches of the student and teacher, and supervise the conditional prediction together with the corresponding guidance direction. We further introduce noise-aware OPD. The PE target improves semantic and structural planning at high noise levels, while the standard RL target is retained at low noise levels to refine visual details.

We initialize the student model from an SFT checkpoint. During rollout, we employ CFG sampling to generate and record the student’s denoising trajectories. During training, the intermediate states along each trajectory are routed to the corresponding T2I or image-editing RL teacher according to the task category of each sample. Our supervision objective aligns both the student’s positive predictions and its conditional guidance directions with those produced by the corresponding teacher.

To further enhance the model’s semantic and structural planning capabilities, we introduce a PE teacher in addition to the RL teacher and propose a noise-aware OPD. This mechanism selects the supervision target according to the learning requirements at different stages of the denoising process. At high noise levels, the objective derived from the PE teacher guides the student in learning global semantics and content layout. At low noise levels, we retain the standard RL teacher to refine local attributes and visual quality. Let \(q\in\{\mathrm{T2I},\mathrm{Edit}\}\) denote the task and
\(\mathbf{v}_{T_q}(\mathbf{x}_t,t,c)\) the conditional velocity predicted by
the corresponding teacher. The noise-aware target is
\begin{equation}
  \widehat{\mathbf{v}}_{T_q}(\mathbf{x}_t,t,c)
  = \mathbf{v}_{T_q}(\mathbf{x}_t,t,c)
  + h_t\left[
      \mathbf{v}_{T_q}(\mathbf{x}_t,t,c_{\mathrm{PE}})
      - \mathbf{v}_{T_q}(\mathbf{x}_t,t,c)
    \right],
  \label{eq:opd-pe-target}
\end{equation}
where \(h_t=1\) at high noise levels and \(h_t=0\) at low noise levels, and
\(c\) and \(c_{\mathrm{PE}}\) denote the original condition and its
semantically constrained prompt-enhanced version, respectively.

\subsubsection{Few-step Distillation}

To improve inference efficiency, we introduce a scalable few-step distillation framework based on the average-velocity formulation of MeanFlow~\cite{geng2025meanflow} in the final training stage. It compresses a full-step model with CFG into a guidance-free 8/4 step model. By transferring the teacher's long-trajectory generative capability to a fixed few-step schedule and optimizing the student's global transport behavior under large sampling steps, our framework preserves prompt adherence, global composition, and visual fidelity while substantially accelerating inference.

Our framework combines average-velocity MeanFlow consistency distillation with DMD-style online trajectory alignment~\cite{yin2024dmd}. Given a clean latent predicted by the student, an auxiliary predictor and the teacher independently estimate the mean flow over the same noise interval. Their discrepancy quantifies the transport gap between the teacher and student
trajectories and serves as the optimization signal for the student.

To align training with the 8/4-step sampler, we introduce local trajectory anchoring. For each interval sampled from the inference schedule, fine-grained teacher integration provides a target mean flow that directly supervises the student’s initial velocity. Complementing random-interval alignment, this objective reduces Euler discretization errors and mitigates structural drift, texture loss, and detail degradation. Together, these objectives enable efficient few-step generation while preserving text rendering, semantic understanding, composition, and visual fidelity.

\subsection{Training Infrastructure}
\label{sec:training-infra}
\label{sec:distributed-training}

The unified interface of \ModelName{} also simplifies its training
infrastructure. Understanding, text-to-image generation, editing, and
multi-reference synthesis are all represented as interleaved sequences of
text tokens, semantic image tokens, and image latent tokens. Position IDs,
attention masks, and loss masks describe the structure and supervision of
each example. Although text and image targets use different objectives,
the tasks share one Transformer training loop. Multi-task training therefore
becomes a sequence-training problem that can reuse mature LLM infrastructure.

We apply several established LLM training techniques to this shared interface.
Token packing~\cite{krell2021packing} combines variable-length examples within
a common token budget, reducing padding while keeping examples separate
through attention masks. Fully sharded data parallelism
(FSDP)~\cite{zhao2023fsdp} distributes parameters, gradients, and optimizer
states across accelerators to reduce memory use per device. For high-resolution
and multi-reference training, Ulysses context parallelism
(CP)~\cite{jacobs2024ulysses} splits a long sequence across devices while
preserving attention within each example, supporting packed contexts of up
to 64K tokens. BF16 mixed precision and activation checkpointing further
reduce memory requirements.

The MoE backbone also offers a natural path to expert parallelism (EP),
which distributes experts across devices and sends tokens to their selected
experts. The reported training recipe uses FSDP and CP, with EP size set to
1 (Table~\ref{tab:training-recipe}). These choices can be made at the training
system level while retaining the same multi-task input interface.

\section{Efficiency}
\label{sec:efficiency}

\begin{table*}[t!]
    \centering
    \fontsize{8pt}{10pt}\selectfont
    \renewcommand{\arraystretch}{1.0}
    \setlength{\tabcolsep}{1.4mm}
    \begin{tabular}{l|c|ccc|cccc}
        \toprule
        \multirow{2}{*}{\textbf{Model Name}}
        & \multirow{2}{*}{\textbf{Steps}}
        & \multicolumn{3}{c|}{\textbf{Text-to-Image}}
        & \multicolumn{4}{c}{\textbf{Image Editing} (\# input images)} \\
        \cmidrule(lr){3-5}\cmidrule(lr){6-9}
        & & $1024\times576$ & $1024\times768$ & $1024\times1024$
        & 1 img & 2 imgs & 3 imgs & 4 imgs \\
        \midrule
        Qwen-Image-2512-Lightning~\cite{wu2025qwenimage}
        & 4 & 1.79 & 2.33 & 3.21 & -- & -- & -- & -- \\
        Qwen-Image-2512~\cite{wu2025qwenimage}
        & 50 & 35.82 & 47.81 & 67.13 & -- & -- & -- & -- \\
        Qwen-Image-Edit-2511-Lightning~\cite{wu2025qwenimage}
        & 4 & -- & -- & -- & 7.52 & 12.82 & 19.56 & 27.35 \\
        Qwen-Image-Edit-2511~\cite{wu2025qwenimage}
        & 40 & -- & -- & -- & 133.04 & 233.24 & 360.18 & 510.64 \\
        Z-Image~\cite{cai2025z}
        & 50 & 26.67 & 35.82 & 49.43 & -- & -- & -- & -- \\
        SenseNova-U1-8B-MoT~\cite{diao2026sensenova}
        & 50 & 10.62 & 12.95 & 16.99 & 18.30 & 19.91 & 21.55 & 23.44 \\
        HunyuanImage-3.0-Instruct-Distil$^{\dagger}$~\cite{hunyuanimage2025}
        & 8 & -- & -- & 12.20 & 16.79 & 21.97 & 27.65 & -- \\
        \midrule
        \textbf{\ModelName{}}
        & 50 & 32.89 & 43.26 & 56.99 & 67.71 & 78.43 & 89.24 & 100.18 \\
        \textbf{\ModelNameTurboeight{}}
        & 8 & 2.90 & 3.80 & 4.98 & 6.78 & 8.78 & 10.92 & 13.17 \\
        \bottomrule
    \end{tabular}
    \caption{End-to-end inference latency (seconds, lower is better) on NVIDIA H20 GPUs. Each entry is the mean wall-clock time over three timed runs following one warm-up run, and excludes model loading. Steps denotes the number of sampling steps. Editing cases are indexed by the number of input images, so one input image corresponds to instruction-guided single-image editing and two to four input images to multi-reference composition. All cases produce a $1024\times1024$ output. Lightning denotes the 4-step distillation LoRA of the respective base model. The Lightning configurations and \ModelNameTurbo{} are sampled guidance-free with \texttt{true\_cfg\_scale}$=1$, so they perform one forward pass per step rather than the two required under classifier-free guidance. $^{\dagger}$Every configuration is served on a single GPU except HunyuanImage-3.0-Instruct-Distil, which \textbf{requires a 4-way tensor-parallel instance spanning four GPUs}. Its per-instance latency therefore corresponds to four times the accelerator cost of the other entries. It is run with \texttt{bot\_task}$=$\texttt{image} and without external chain-of-thought prompt rewriting.}
    \label{tab:latency}
\end{table*}

Inference cost determines whether a unified generation-and-editing model can be deployed at scale. Editing is the critical case because the visual context grows with the number of input images, and attention over the concatenated token sequence can make latency grow super-linearly. We therefore measure end-to-end latency on text-to-image generation and on instruction-guided editing with one to four input images, treating single-image editing and multi-reference composition as one family parameterized by the number of inputs. Table~\ref{tab:latency} compares the end-to-end inference latency of \ModelName{} against popular models.

\paragraph{Protocol.}
All measurements use NVIDIA H20 GPUs, a fixed random seed, and an identical software stack (PyTorch 2.9.1, Diffusers 0.39.0, Transformers 5.2.0). Each configuration is served on a single GPU, except HunyuanImage-3.0-Instruct-Distil, which requires a 4-way tensor-parallel instance spanning four GPUs. For every model--case pair we execute one warm-up pass followed by three timed passes, and report the mean wall-clock time from the prompt and reference images to the decoded output, excluding model loading. Run-to-run variation is negligible, staying below $0.5\%$ of the mean throughout.

We evaluate \ModelName{} at 50 sampling steps and its distilled variant \ModelNameTurbo{} at 8 steps. The baselines include Qwen-Image-2512 and Qwen-Image-Edit-2511, each in its released standard setting and with its official 4-step Lightning LoRA. We also include Z-Image, SenseNova-U1-8B-MoT, and HunyuanImage-3.0-Instruct-Distil. The latter, like our models and SenseNova-U1-8B-MoT, serves both generation and editing from one checkpoint. One sampling detail is important when reading the table. The Lightning configurations and \ModelNameTurbo{} are guidance-free (\texttt{true\_cfg\_scale}$=1$) and evaluate the network once per step, whereas the standard configurations use classifier-free guidance and evaluate it twice. Step counts alone therefore understate the compute gap between the two regimes by a factor of two. Because the systems also differ in size, sampling schedule, and model parallelism, these results characterize the released inference settings rather than isolating architectural efficiency. Our models run through PyTorch scaled dot-product attention without kernel-level acceleration, quantization, step caching, or graph compilation, so their figures are an unoptimized upper bound.

\paragraph{Text-to-image latency.}
At a matched 50-step budget, \ModelName{} is consistently faster than Qwen-Image-2512, and its margin widens with resolution, from $8\%$ at $1024\times576$ to $15\%$ at $1024\times1024$. The difference reflects their scaling behaviour. Over this range, the pixel count grows by $1.78\times$ while the latency of \ModelName{} grows by only $1.73\times$, against $1.87\times$ for Qwen-Image-2512 and $1.85\times$ for Z-Image. Cost is thus close to linear in the number of generated tokens, indicating that quadratic attention has not yet become the dominant term at megapixel resolution. It also depends on the token count rather than the spatial layout. Transposed portrait resolutions differ from their landscape counterparts by at most $0.3\%$, and we omit them from the table. Z-Image remains faster in absolute terms, as expected from its smaller backbone, while the few-step configurations belong to a separate regime discussed below.

\paragraph{Editing latency and input scaling.}
The advantage is largest on editing, where the input images dominate the sequence length. With one input image \ModelName{} is $1.96\times$ faster than Qwen-Image-Edit-2511 despite using more sampling steps ($50$ versus $40$), and the gap widens to $5.10\times$ with four inputs. The marginal cost of an additional input stays near $10.8$\,s for \ModelName{} but rises from $100$\,s to $150$\,s for Qwen-Image-Edit-2511. Going from one to four inputs therefore costs $1.48\times$ for our model against $3.84\times$ for theirs. This near-linear growth in the visual context is what makes multi-reference composition practical at serving time, as reflected in the multi-image editing latencies reported in Table~\ref{tab:latency}. SenseNova-U1-8B-MoT is faster in absolute terms with an even flatter marginal cost, consistent with its much smaller $8$B backbone.

\paragraph{Few-step regime.}
\ModelNameTurbo{} moves our model into the distilled operating point. It generates a $1024\times1024$ image in $4.98$\,s, an $11.4\times$ reduction over \ModelName{}, and is $7.6$ to $10.0\times$ faster on the editing cases. The gain exceeds the $6.25\times$ reduction in sampling steps because guidance-free sampling also cuts network evaluations per image from $100$ to $8$. Against HunyuanImage-3.0-Instruct-Distil, the matched 8-step unified baseline, \ModelNameTurbo{} is roughly $2.5\times$ faster on both generation and editing while running on one GPU instead of four, a tenfold reduction in accelerator time. It likewise beats Qwen-Image-Edit-2511-Lightning at every input count despite twice the step budget, with the margin growing from $1.11\times$ to $2.08\times$ as inputs are added, and it inherits the flat marginal cost of the base model at roughly $2.1$\,s per additional input. Qwen-Image-2512-Lightning is faster in absolute terms on text-to-image generation, as expected at half the step count, though \ModelNameTurbo{} is the cheaper of the two per sampling step ($0.62$\,s versus $0.80$\,s).

\section{Evaluation}
\label{sec:evaluation}
\subsection{Quantitative Results}

\subsubsection{Text-to-Image \& Text Rendering}

To evaluate the foundational generation quality and the specific capability of rendering complex visual text, we benchmark \ModelName{} against a suite of leading open-source and proprietary text-to-image models, including Qwen-Image~\cite{wu2025qwenimage}, Qwen-Image-2512~\cite{wu2025qwenimage}, Z-Image-Turbo~\cite{cai2025z}, HunyuanImage-3.0~\cite{hunyuanimage2025}, ERNIE-Image-Turbo~\cite{liu2026ernie}, GLM-Image~\cite{glm_image}, HiDream-O1-Image~\cite{cai2026hidreamo1}, SenseNova-U1-8B-MoT~\cite{diao2026sensenova}, Boogu-Image-0.1~\cite{chen2026booguimage01} and FLUX.2-dev~\cite{flux-2-2025}. We report results on multiple widely-used benchmarks: GenEval~\cite{ghosh2023geneval}, PRISM-Bench~\cite{fang2026flux}, OneIG-Bench~\cite{chang2025oneig}, TIIF-Bench~\cite{wei2025tiif}, DPG-Bench~\cite{hu2024ella}, Qwen-Image-Bench~\cite{li2026qwen}, WeGenBench~\cite{liang2026wegenbench}, LongText-Bench~\cite{geng2025x}, and CVTG-2K~\cite{du2025textcrafter},  which together cover compositional generation, long/complex text rendering, general image quality, and instruction-following ability. Table~\ref{tab:overall_results_summary} summarizes the overall quantitative results. We can see that \ModelName{} ranks among the top-tier open-source models on both mainstream and emerging evaluation benchmarks.

\begin{table*}[h!]
\centering
\fontsize{8pt}{10pt}\selectfont
\renewcommand{\arraystretch}{1.0}
\setlength{\tabcolsep}{1.8mm}
\begin{tabularx}{\linewidth}{l|cccccc}
\toprule
\textbf{Model Name} & \textbf{GenEval}$\uparrow$ & \textbf{OneIG}$\uparrow$ & \textbf{PRISM}$\uparrow$ & \textbf{LongText}$\uparrow$ & \textbf{Qwen-Image-Bench}$\uparrow$ & \textbf{WeGenBench}$\uparrow$  \\
\midrule
GPT-Image-2 & 0.886 & 0.515 & 8.039 & 0.972 & 63.165 & 0.969 \\
Nano-Banana-Pro & 0.781 & 0.568 & 7.679 & 0.914 & 59.803 & 0.959 \\
Qwen-Image-3.0
& 0.857 & 0.549 & 7.685 & 0.980 & 61.692 & 0.962 \\
Seedream-5.0-Pro
& 0.864 & 0.532 & 7.421 & 0.947 & 59.591 & 0.968 \\
\midrule
Qwen-Image~\cite{wu2025qwenimage}  & \hlthird{0.871} & 0.530 & 7.072 & 0.934 & 46.325 & 0.909 \\
Qwen-Image-2512~\cite{wu2025qwenimage} & 0.747 & 0.521 & 6.888 & 0.954 & 49.444 & 0.897 \\
Z-Image-Turbo~\cite{cai2025z}   & 0.748 & 0.510 & 6.993 & 0.928 & 49.094 & 0.897 \\
HunyuanImage-3.0-Instruct-Distil~\cite{hunyuanimage2025} & 0.844  & 0.498 & 6.959 & 0.886 & 50.087 & 0.931 \\
HunyuanImage-3.0-Instruct-Distil-CoT~\cite{hunyuanimage2025} & 0.800  & 0.523 & 7.074 & 0.903 & 51.690 & \hlsecond{0.938} \\
ERNIE-Image-Turbo~\cite{liu2026ernie} & 0.845 & 0.521 & 7.111 & \hlsecond{0.965} & 53.335 & 0.923\\
ERNIE-Image-Turbo-PE~\cite{liu2026ernie} & 0.838 & \hlbest{0.554} & 7.163 & \hlbest{0.967} & \hlsecond{55.309} & 0.920 \\
GLM-Image~\cite{glm_image}  & 0.721 & 0.516 & 6.524 & 0.942 & 45.186 & 0.854\\
HiDream-O1-Image~\cite{cai2026hidreamo1}        & \hlsecond{0.881} & 0.504 & 6.830 & \hlsecond{0.965} & 44.750 & 0.868 \\
SenseNova-U1-8B-MoT~\cite{diao2026sensenova}   & 0.857 & 0.524 & 6.424 & 0.859 & 47.719 & 0.932 \\
SenseNova-U1-8B-MoT-CoT~\cite{diao2026sensenova}   & \hlbest{0.882} & 0.531 & 6.462 & 0.885 & 47.605 & 0.922 \\
Boogu-Image-0.1-Base~\cite{chen2026booguimage01} & 0.812 & \hlsecond{0.547}  & \hlsecond{7.500} & \hlthird{0.961} & 52.591 & 0.930 \\
Boogu-Image-0.1-Turbo~\cite{chen2026booguimage01} & 0.779 & 0.529 & 7.246 & 0.958 & 51.349 & \hlthird{0.934} \\
FLUX.2-dev~\cite{flux-2-2025} & 0.853 & 0.531 & 7.385 & 0.851 & 53.048 & 0.928 \\
\midrule
\textbf{\ModelName{}}  & 0.851 & \hlthird{0.534} & \hlbest{7.555} & 0.931  & \hlthird{54.155} & \hlbest{0.948} \\
\textbf{\ModelNameTurboeight} & 0.836 & 0.520 & \hlthird{7.482} & 0.933  & \hlbest{55.992} & \hlbest{0.948} \\
% \textbf{\ModelNameTurbofour} & 0.817 & 0.521 & 7.450 & 0.879  & 54.138 & - \\
\bottomrule
  \end{tabularx}
  \caption{Quantitative evaluation results on Text-to-Image generation and Text Rendering benchmarks. Closed-source models including GPT-Image-2, Nano-Banana-Pro, Qwen-Image-3.0, and Seedream-5.0-Pro are excluded from rank highlighting. Darker cell colors indicate better performance, with the three highlighted shades denoting the 1st, 2nd, and 3rd best results, respectively.}
  \label{tab:overall_results_summary}
\end{table*}

\begin{table}[h!]
\fontsize{8pt}{10pt}\selectfont
\renewcommand{\arraystretch}{1.0}
\setlength{\tabcolsep}{2.5mm}
\begin{tabular}{l|ccccccc}
\toprule
\textbf{Model Name} & \textbf{Single Obj.} & \textbf{Two Obj.} & \textbf{Counting} & \textbf{Colors} & \textbf{Position} & \textbf{Attr Bind.} & \textbf{Overall}$\uparrow$ \\
\midrule
GPT-Image-2 & 0.9969 & 0.9621 & 0.8844 & 0.9043 & 0.8325 & 0.7375 & 0.8863 \\
Nano-Banana-Pro & 0.9844 & 0.8914 & 0.7219 & 0.7606 & 0.8250 & 0.5000 & 0.7806 \\
Qwen-Image-3.0
& 0.9969 & 0.9444 & 0.8250 & 0.9309 & 0.7550 & 0.6900 & 0.8570 \\
Seedream-5.0-Pro
& 0.9938 & 0.9444 & 0.7469 & 0.9149 & 0.8775 & 0.7050 & 0.8637 \\
\midrule
Qwen-Image           & 0.9906 & 0.9545 & \hlbest{0.8875} & 0.8989 & 0.7675 & 0.7275 & \hlthird{0.8711} \\
Qwen-Image-2512      & 0.9844 & 0.9242 & 0.6188 & 0.8750 & 0.4750 & 0.6075 & 0.7475 \\
Z-Image-Turbo        & \hlthird{0.9938} & 0.8990 & 0.7375 & 0.8404 & 0.4475 & 0.5675 & 0.7476 \\
HunyuanImage-3.0-Instruct-Distil  & \hlsecond{0.9969} & \hlthird{0.9621} & 0.7375 & 0.8697 & 0.7575 & 0.7400 & 0.8439 \\
HunyuanImage-3.0-Instruct-Distil-CoT & 0.9875 & 0.9318 & 0.7031 & 0.8431 & 0.6950 & 0.6400 & 0.8001\\
ERNIE-Image-Turbo & \hlbest{1.0000} & 0.9520 & 0.7844 & 0.8670 & 0.7825 & 0.6850 & 0.8452 \\
ERNIE-Image-Turbo-PE & 0.9906 & 0.9470 & \hlthird{0.8156} & 0.8404 & 0.7850 & 0.6500 & 0.8381 \\
GLM-Image            & \hlsecond{0.9969} & 0.8939 & 0.6906 & 0.6755 & 0.6200 & 0.4475 & 0.7207 \\
HiDream-O1-Image           & \hlbest{1.0000} & \hlsecond{0.9672} & 0.7438 & \hlthird{0.9096} & \hlbest{0.9150} & \hlsecond{0.7500} & \hlsecond{0.8809} \\
SenseNova-U1-8B-MoT  & \hlsecond{0.9969} & 0.9369 & 0.7812 & \hlsecond{0.9309} & 0.8175 & 0.6775 & 0.8568 \\
SenseNova-U1-8B-MoT-CoT         & \hlbest{1.0000} & 0.9571 & \hlsecond{0.8625} & 0.9016 & \hlthird{0.8800} & 0.6900 & \hlbest{0.8819} \\
Boogu-Image-0.1-Base & \hlthird{0.9938} & 0.9419 & 0.6312 & 0.8404 & 0.8250 & 0.6375 & 0.8116\\
Boogu-Image-0.1-Turbo & \hlbest{1.0000} & \hlbest{0.9747} &  0.5531 & 0.7420 & \hlsecond{0.8875} & 0.5175 & 0.7791\\
FLUX.2-dev & \hlsecond{0.9969} & 0.9192 & 0.7875 & \hlbest{0.9468} & 0.7150 & \hlbest{0.7550} & 0.8534 \\
\midrule
\textbf{\ModelName{}}    & \hlsecond{0.9969} & 0.9520 & 0.7594 & 0.9069 & 0.7475 & \hlthird{0.7425} & 0.8509 \\
\textbf{\ModelNameTurboeight{}} & \hlbest{1.0000} & 0.9444 & 0.7125 & 0.8670 & 0.8000 & 0.6900 & 0.8357 \\
% \textbf{\ModelNameTurbofour{}} & 0.9969 & 0.9217 & 0.6656 & 0.8803 & 0.7650 & 0.6725 & 0.8170 \\
\bottomrule
\end{tabular}
\caption{Detailed evaluation results on GenEval across different object-centric compositional skills. Closed-source models are excluded from rank highlighting. Darker cell colors indicate better performance, with the three highlighted shades denoting the 1st, 2nd, and 3rd best results, respectively.}
\label{tab:geneval_detailed}
\end{table}

\paragraph{GenEval}
GenEval~\cite{ghosh2023geneval} is designed to rigorously evaluate object-centric generation capabilities. Unlike benchmarks that rely on holistic image-text alignment scores, GenEval isolates specific compositional skills. It tests whether a model can accurately render single or multiple distinct objects, correctly bind attributes (e.g., colors) to specific entities, adhere to precise object counts, and understand complex spatial relations (e.g., ``left of'', ``below''). This fine-grained evaluation is crucial for determining if a model truly understands object composition rather than merely memorizing co-occurrences. Table~\ref{tab:geneval_detailed} shows that the \textbf{\ModelName{}} family performs strongly on GenEval: WeLM-Image-Turbo-8step achieves a perfect score (tied for first) on single-object generation, while WeLM-Image attains the second-highest score on single-object generation and ranks third on attribute binding among open-source models, reflecting reliable compositional understanding and competitive object-level fidelity.

\begin{table*}[h!]
\centering
\fontsize{8pt}{10pt}\selectfont
\renewcommand{\arraystretch}{1.0}
\setlength{\tabcolsep}{2.5mm}
\begin{tabular}{l|ccccc}
\toprule
\textbf{Model Name} & \textbf{Imagination} & \textbf{Entity} & \textbf{Style} & \textbf{Composition} & \textbf{Overall}$\uparrow$ \\
\midrule
GPT-Image-2 & 7.48 & 8.20 & 8.08 & 8.54 & 8.04 \\
Nano-Banana-Pro & 7.06 & 7.78 & 7.74 & 8.11 & 7.68 \\
Qwen-Image-3.0
& 6.95 & 7.80 & 7.64 & 8.18 & 7.68 \\
Seedream-5.0-Pro
& 7.24 & 7.82 & 7.94 & 8.48 & 7.42 \\
\midrule
Qwen-Image & 6.33 & 7.15 & 7.00 & 7.97 & 7.07 \\
Qwen-Image-2512  & 6.47 & 7.04 & 7.02 & 8.10 & 6.89 \\
Z-Image-Turbo  & 5.90 & 6.92 & 6.72 & 7.78 & 6.99 \\
HunyuanImage-3.0-Instruct-Distil & 6.57 & 6.94 & 7.04 & 8.18 & 6.96 \\
HunyuanImage-3.0-Instruct-Distil-CoT & 6.40 & 7.05 & 6.91 & 7.92 & 7.07 \\
ERNIE-Image-Turbo & 6.48 & 6.58 & 7.14 & 8.04 & 7.11 \\
ERNIE-Image-Turbo-PE & 6.79 & 6.72 & 7.23 & 7.73 & 7.16 \\
GLM-Image   & 5.71 & 6.68 & 6.31 & 7.50 & 6.52 \\
HiDream-O1-Image & 6.89 & 6.83 & 6.96 & 7.86 & 6.83 \\
SenseNova-U1-8B-MoT  & 6.66 & 6.64 & 7.02 & 7.65 & 6.42 \\
SenseNova-U1-8B-MoT-CoT  & 6.71 & 6.57 & 6.72 & 7.65 & 6.46 \\
Boogu-Image-0.1-Base & 6.74 & \hlbest{7.51} & 7.32 & \hlsecond{8.37} & \hlsecond{7.50} \\
Boogu-Image-0.1-Turbo & \hlsecond{6.99} & 7.23 & 7.28 & 8.23 & 7.25 \\
FLUX.2-dev & 6.64 & \hlsecond{7.38} & \hlthird{7.37} & \hlthird{8.36} & 7.39 \\
\midrule
\textbf{\ModelName{}} & \hlbest{7.10} & \hlthird{7.35} & \hlbest{7.57} & \hlbest{8.40} & \hlbest{7.56} \\
\textbf{\ModelNameTurboeight{}} & \hlthird{6.94} & 7.32 & \hlsecond{7.40} & 8.31 & \hlthird{7.48} \\
% \textbf{\ModelNameTurbofour{}} & 6.94 & 7.29 & 7.34 & 8.35 & 7.45 \\
\bottomrule
\end{tabular}
\caption{Detailed evaluation results on PRISM-Bench across reasoning and aesthetic dimensions. Closed-source models are excluded from rank highlighting. Darker cell colors indicate better performance, with the three highlighted shades denoting the 1st, 2nd, and 3rd best results, respectively.}
\label{tab:prism_detailed}
\end{table*}

\paragraph{PRISM-Bench}
PRISM-Bench~\cite{fang2026flux} is a comprehensive VLM-powered benchmark that goes beyond simple text-image matching. It evaluates models across diverse axes including Imagination (creative interpretation of abstract concepts), Entity (accurate rendering of specific subjects), Style (adherence to requested artistic mediums), and Composition (overall visual layout). By leveraging advanced Vision-Language Models as evaluators, PRISM-Bench offers a holistic assessment that jointly considers the logical reasoning behind the prompt and the aesthetic quality of the generated image. As shown in Table~\ref{tab:prism_detailed}, \textbf{\ModelName{}} ranks first among open-source models in Overall score as well as across Imagination, Style, and Composition, while placing third in Entity (with WeLM-Image-Turbo-8step also ranking second in Style and third in Overall), demonstrating strong and well-rounded performance across PRISM-Bench's fine-grained dimensions.

\begin{table}[h!]
\centering
\fontsize{8pt}{10pt}\selectfont
\renewcommand{\arraystretch}{1.0}
\setlength{\tabcolsep}{3mm}
\begin{tabular}{l|cccccc}
\toprule
\textbf{Model Name} & \textbf{Alignment} & \textbf{Text} & \textbf{Reasoning} & \textbf{Style} & \textbf{Diversity} & \textbf{Overall}$\uparrow$ \\
\midrule
GPT-Image-2 & 0.8716 & 0.8402 & 0.3354 & 0.3633 & 0.1661 & 0.5153 \\
Nano-Banana-Pro & 0.8741 & 0.9406 & 0.3225 & 0.4589 & 0.2416 & 0.5675 \\
Qwen-Image-3.0
& 0.8634 & 0.9419 & 0.3193 & 0.3998 & 0.2210 & 0.5491 \\
Seedream-5.0-Pro
& 0.8735 & 0.8361 & 0.3216 & 0.4267 & 0.2036 & 0.5323 \\
\midrule
Qwen-Image           & 0.8569 & 0.9215 & 0.2826 & \hlsecond{0.4180} & 0.1719 & 0.5302 \\
Qwen-Image-2512      & 0.8542 & \hlbest{0.9864} & 0.2818 & 0.3433 & 0.1391 & 0.5210 \\
Z-Image-Turbo        & 0.8127 & \hlsecond{0.9743} & 0.2905 & 0.3261 & 0.1471 & 0.5101 \\
HunyuanImage-3.0-Instruct-Distil     & 0.8362 & 0.8777 & 0.2248 & 0.3526 & 0.1964 & 0.4975 \\
HunyuanImage-3.0-Instruct-Distil-CoT    & 0.8291 & 0.8547 & \hlsecond{0.3108} & 0.3483 & \hlbest{0.2741} & 0.5234 \\
ERNIE-Image-Turbo & 0.8494 & 0.9668 & 0.2698 & 0.3981 & 0.1220 & 0.5212 \\
ERNIE-Image-Turbo-PE & 0.8256 & 0.9699 & \hlbest{0.3303} & \hlthird{0.4037} & \hlsecond{0.2392} & \hlbest{0.5538} \\
GLM-Image   & 0.7785 & 0.9578 & 0.2897 & 0.3304 & \hlthird{0.2225} & 0.5158 \\
HiDream-O1-Image  & 0.8152 & 0.9685 & 0.2905 & 0.3097 & 0.1359 & 0.5040 \\
SenseNova-U1-8B-MoT & 0.8423 & 0.9085 & 0.3059 & 0.3961 & 0.1674 & 0.5240 \\
SenseNova-U1-8B-MoT-CoT & 0.8419 & 0.9416 & \hlthird{0.3106} & 0.3737 & 0.1897 & 0.5315 \\
Boogu-Image-0.1-Base & \hlthird{0.8590} & \hlthird{0.9733} & 0.2957 & 0.4018 & 0.2069 & \hlsecond{0.5473} \\
Boogu-Image-0.1-Turbo & \hlbest{0.8737} & 0.9421 & 0.2989 & 0.3896 & 0.1428 & 0.5294 \\
FLUX.2-dev & 0.8583 & 0.9105 & 0.3084 & \hlbest{0.4293} & 0.1505 & 0.5314 \\
\midrule
\textbf{\ModelName{}} & \hlsecond{0.8632} & 0.9647 & 0.2833 & 0.3651 & 0.1922 & \hlthird{0.5337} \\
\textbf{\ModelNameTurboeight{}}  & 0.8542 & 0.9466 & 0.2759 & 0.3123 & 0.2089 & 0.5196 \\
% \textbf{\ModelNameTurbofour{}}  & 0.8515 & 0.9447 & 0.2734 & 0.3455 & 0.1883 & 0.5207 \\
\bottomrule
\end{tabular}
\caption{Detailed evaluation results on OneIG-Bench across fine-grained alignment dimensions. Closed-source models are excluded from rank highlighting. Darker cell colors indicate better performance, with the three highlighted shades denoting the 1st, 2nd, and 3rd best results, respectively.}
\label{tab:oneig_detailed}
\end{table}

\paragraph{OneIG-Bench}
OneIG-Bench~\cite{chang2025oneig} provides a comprehensive evaluation of fine-grained text-to-image alignment. It assesses models across multiple dimensions, including semantic alignment, text rendering accuracy, reasoning capabilities, stylistic consistency, and generation diversity. This benchmark is particularly effective at identifying whether a model can precisely follow complex, multi-faceted instructions without hallucinating or omitting details. As reported in Table~\ref{tab:oneig_detailed}, \ModelName{} secures the second-best Alignment score and ranks third overall on OneIG-Bench among open-source models, balancing prompt fidelity with varied visual generation behavior, as reflected in its competitive Text and Diversity scores.

\begin{table}[h!]
\centering
\fontsize{8pt}{10pt}\selectfont
\renewcommand{\arraystretch}{1.0}
\setlength{\tabcolsep}{2.5mm}
\begin{tabular}{l|cccccccc}
\toprule
\multirow{2}{*}{\textbf{Model Name}}
& \multicolumn{2}{c}{\textbf{Basic}} & \multicolumn{2}{c}{\textbf{Advanced}} & \multicolumn{2}{c}{\textbf{Designer}} & \multicolumn{2}{c}{\textbf{Overall}} \\
\cmidrule(lr){2-3} \cmidrule(lr){4-5} \cmidrule(lr){6-7} \cmidrule(lr){8-9}
& short$\uparrow$   & long$\uparrow$ & short$\uparrow$   & long$\uparrow$  & short$\uparrow$   & long$\uparrow$ & short$\uparrow$   & long$\uparrow$  \\
\midrule
GPT-Image-2 & 0.8631 & 0.8568 & 0.8353 & 0.8347 & 0.9262 & 0.9139 & 0.8547 & 0.8509 \\
Nano-Banana-Pro & 0.8325 & 0.8473 & 0.7970 & 0.7994 & 0.8918 & 0.8769 & 0.8194 & 0.8240 \\
Qwen-Image-3.0 & 0.8745 & 0.8417 & 0.8112 & 0.7847 & 0.8715 & 0.9157 & 0.8390 & 0.8183 \\
Seedream-5.0-Pro & 0.8410 & 0.8569 & 0.8057 & 0.8328 & 0.9067 & 0.9478 & 0.8287 & 0.8536 \\
\midrule
Qwen-Image           & 0.8217 & 0.8239 & \hlthird{0.7995} & \hlbest{0.8226} & 0.8694 & 0.8545 & \hlthird{0.8146} & \hlbest{0.8266} \\
Qwen-Image-2512      & 0.7792 & 0.8034 & 0.7451 & 0.7656 & 0.8795 & \hlthird{0.8769} & 0.7714 & 0.7906 \\
Z-Image-Turbo        & 0.7503 & 0.7798 & 0.7552 & 0.7404 & 0.8358 & 0.8694 & 0.7625 & 0.7679 \\
HunyuanImage-3.0-Instruct-Distil     & 0.8134 & 0.8248 & 0.7583 & 0.7363 & \hlthird{0.8806} & 0.8284 & 0.7903 & 0.7760 \\
HunyuanImage-3.0-Instruct-Distil-CoT& 0.8202 & 0.8132 & \hlsecond{0.8054} & 0.7739 & 0.8694 & \hlthird{0.8769} & \hlsecond{0.8175} & 0.7985 \\
ERNIE-Image-Turbo & 0.8206 & 0.7844 & 0.7659 & 0.7735 & \hlbest{0.9036} & \hlsecond{0.8806} & 0.7994 & 0.7890 \\
ERNIE-Image-Turbo-PE & 0.8018 & 0.7848 & 0.7681 & \hlthird{0.7975} & 0.8582 & \hlbest{0.9030} & 0.7894 & 0.8050 \\
GLM-Image   & 0.8060 & 0.7930 & 0.7082 & 0.7456 & 0.8209 & 0.7948 & 0.7533 & 0.7668 \\
HiDream-O1-Image  & 0.8348 & \hlbest{0.8435} & 0.7896 & 0.7856 & 0.8545 & 0.8433 & 0.8119 & 0.8113 \\
SenseNova-U1-8B-MoT & \hlbest{0.8520} & 0.8154 & \hlbest{0.8248} & 0.7544 & 0.8694 & 0.8694 & \hlbest{0.8388} & 0.7875 \\
SenseNova-U1-8B-MoT-CoT & \hlsecond{0.8404} & \hlthird{0.8264} & 0.7802 & 0.7708 & 0.8694 & 0.8433 & 0.8102 & 0.7974 \\
Boogu-Image-0.1-Base & 0.7939 & \hlsecond{0.8423} & 0.7695 & 0.7915 & 0.8731 & 0.8358 & 0.7892 & \hlthird{0.8134} \\
Boogu-Image-0.1-Turbo & 0.8009 & 0.8226 & 0.7704 & 0.7783 & \hlthird{0.8806} & \hlsecond{0.8806} & 0.7928 & 0.8045 \\
FLUX.2-dev & 0.8081 & 0.8152 & 0.7912 & \hlsecond{0.8087} & 0.8582 & \hlsecond{0.8806} & 0.8042 & \hlsecond{0.8188} \\
\midrule
\textbf{\ModelName{}}  & \hlthird{0.8363} & 0.8151 & 0.7769 & 0.7714 & 0.8769 & 0.8619 & 0.8078 & 0.7960 \\
\textbf{\ModelNameTurboeight{}} & 0.8328 & 0.8047 & 0.7439 & 0.7661 & \hlsecond{0.8993} & 0.8657 & 0.7908 & 0.7901 \\
% \textbf{\ModelNameTurbofour{}} & 0.8088 & 0.8621 & 0.7320 & 0.7644 & 0.8470 & 0.8246 & 0.7704 & 0.8037 \\
\bottomrule
\end{tabular}
\caption{Detailed evaluation results on TIIF-Bench. Closed-source models are excluded from rank highlighting. Darker cell colors indicate better performance, with the three highlighted shades denoting the 1st, 2nd, and 3rd best results, respectively.}
\label{tab:tiff_detailed}
\end{table}

\begin{table}[h!]
\centering
\fontsize{8pt}{10pt}\selectfont
\renewcommand{\arraystretch}{1.0}
\setlength{\tabcolsep}{3mm}
\begin{tabular}{l|cccccc}
\toprule
\textbf{Model Name} & \textbf{Global} & \textbf{Entity} & \textbf{Attribute} & \textbf{Relation} & \textbf{Other} & \textbf{Overall}$\uparrow$ \\
\midrule
GPT-Image-2 & 0.8344 & 0.9290 & 0.8966 & 0.9522 & 0.8554 & 0.8709 \\
Nano-Banana-Pro & 0.8359 & 0.9252 & 0.8939 & 0.9496 & 0.8640 & 0.8684 \\
Qwen-Image-3.0 & 0.8511 & 0.9237 & 0.8944 & 0.9524 & 0.8629 & 0.8656 \\
Seedream-5.0-Pro & 0.8426 & 0.9308 & 0.8997 & 0.9497 & 0.9120 & 0.8732 \\
\midrule
Qwen-Image           & 0.8389 & \hlbest{0.9383} & \hlbest{0.9113} & 0.9516 & \hlbest{0.8880} & \hlbest{0.8881} \\
Qwen-Image-2512      & 0.8480 & 0.9307 & 0.8928 & 0.9485 & \hlbest{0.8880} & 0.8705 \\
Z-Image-Turbo        & 0.8055 & 0.9152 & 0.8830 & 0.9354 & 0.8600 & 0.8509 \\
HunyuanImage-3.0-Instruct-Distil     & 0.8024 & 0.9134 & 0.8894 & 0.9447 & 0.8200 & 0.8485 \\
HunyuanImage-3.0-Instruct-Distil-CoT & 0.8237 & 0.9061 & 0.8793 & 0.9350 & 0.8480 & 0.8432 \\
ERNIE-Image-Turbo & 0.8450 & 0.9250 & 0.8920 & 0.9470 & 0.8640 & 0.8653 \\
ERNIE-Image-Turbo-PE & 0.8389 & 0.9207 & 0.8962 & 0.9408 & 0.8480 & 0.8604 \\
GLM-Image   & 0.8176 & 0.9132 & 0.8677 & 0.9315 & \hlthird{0.8680} & 0.8424 \\
HiDream-O1-Image  & 0.8450 & \hlthird{0.9326} & 0.9042 & \hlbest{0.9571} & 0.8440 & \hlsecond{0.8827} \\
SenseNova-U1-8B-MoT & 0.8389 & 0.9247 & 0.8972 & 0.9489 & \hlsecond{0.8720} & 0.8638 \\
SenseNova-U1-8B-MoT-CoT & \hlthird{0.8511} & 0.9300 & 0.8930 & 0.9458 & 0.8600 & 0.8713 \\
Boogu-Image-0.1-Base & \hlsecond{0.8571} & 0.9284 & \hlsecond{0.9066} & 0.9493 & \hlthird{0.8680} & 0.8733 \\
Boogu-Image-0.1-Turbo & \hlbest{0.8632} & \hlsecond{0.9372} & 0.9030 & \hlthird{0.9524} & 0.8560 & \hlthird{0.8821} \\
FLUX.2-dev & 0.8328 & 0.9260 & 0.9044 & 0.9505 & \hlthird{0.8680} & 0.8750 \\
\midrule
\textbf{\ModelName{}} & 0.8480 & 0.9258 & \hlthird{0.9048} & 0.9513 & 0.8640 & 0.8748 \\
\textbf{\ModelNameTurboeight{}} & \hlsecond{0.8571} & 0.9244 & 0.8978 & \hlsecond{0.9567} & 0.8480 & 0.8659 \\
% \textbf{\ModelNameTurbofour{}} & 0.8571 & 0.9213 & 0.8976 & 0.9540 & 0.8400 & 0.8600 \\
\bottomrule
\end{tabular}
\caption{Detailed evaluation results on DPG-Bench. Closed-source models are excluded from rank highlighting. Darker cell colors indicate better performance, with the three highlighted shades denoting the 1st, 2nd, and 3rd best results, respectively.}
\label{tab:dpg_detailed}
\end{table}

\paragraph{TIIF-Bench} TIIF-Bench~\cite{wei2025tiif} systematically evaluates T2I models' ability to interpret and follow intricate textual instructions, comprising 5,000 prompts across three difficulty levels, each provided in short and long versions with identical core semantics to test robustness to prompt length. It further introduces text rendering and style control as dedicated evaluation dimensions, along with 100 designer-level prompts covering diverse real-world scenarios, and proposes the Global Normalized Edit Distance (GNED) metric for robust typographic accuracy assessment, providing a fine-grained testbed for instruction-following evaluation in text-to-image generation. Table~\ref{tab:tiff_detailed} shows that \ModelName{} reaches the top three on Basic short prompts while maintaining solid Designer and Overall performance, indicating reliable instruction following under varied prompt lengths. Its balanced short- and long-prompt scores suggest that \ModelName{} generalizes well across both simple and complex TIIF-Bench settings.

\begin{table}[t!]
\centering
\fontsize{8pt}{10pt}\selectfont
\renewcommand{\arraystretch}{1.0}
\setlength{\tabcolsep}{3mm}
\begin{tabular}{l|ccccc}
\toprule
\textbf{Model Name} & \textbf{Quality} & \textbf{Aesthetics} & \textbf{Alignment} & \textbf{Creative Gen.} & \textbf{Overall}$\uparrow$ \\
\midrule
GPT-Image-2 & 57.06 & 66.31 & 64.40 & 72.94 & 63.17 \\
Nano-Banana-Pro & 56.05 & 61.55 & 60.49 & 65.97 & 59.80 \\
Qwen-Image-3.0 & 56.96 & 63.80 & 63.08 & 69.41 & 61.69 \\
Seedream-5.0-Pro & 55.57 & 61.93 & 61.10 & 64.98 & 59.59 \\
\midrule
Qwen-Image    & 47.39 & 48.61 & 47.28 & 41.01 & 46.33 \\
Qwen-Image-2512  & 51.68 & 51.85 & 49.71 & 43.60 & 49.44 \\
Z-Image-Turbo   & 51.09 & 51.46 & 48.31 & 44.80 & 49.09 \\
HunyuanImage-3.0-Instruct-Distil  & 50.30 & 52.41 & 51.02 & 47.41 & 50.09 \\
HunyuanImage-3.0-Instruct-Distil-CoT & 50.14 & 53.74 & 52.69 & 53.21 & 51.69 \\
ERNIE-Image-Turbo & \hlthird{53.24} & 55.08 & 53.78 & 52.35 & 53.33 \\
ERNIE-Image-Turbo-PE & \hlsecond{53.76} & \hlsecond{57.20} & \hlsecond{55.61} & \hlbest{59.00} & \hlsecond{55.31} \\
GLM-Image  & 47.99 & 47.20 & 44.68 & 38.81 & 45.19 \\
HiDream-O1-Image  & 46.21 & 47.69 & 45.92 & 37.48 & 44.75 \\
SenseNova-U1-8B-MoT & 46.25 & 49.81 & 50.65 & 45.21 & 47.72 \\
SenseNova-U1-8B-MoT-CoT & 46.78 & 50.33 & 49.50 & 44.10 & 47.61 \\
Boogu-Image-0.1-Base &52.73 & 54.72 & 53.80 & 49.55 & 52.59 \\
Boogu-Image-0.1-Turbo &51.19 & 54.09 & 53.02 & 47.92 & 51.35 \\
FLUX.2-dev & 52.09 & 55.19 & 54.14 & 52.95 & 53.05 \\
\midrule
\textbf{\ModelName{}} & 52.78 & \hlthird{57.08} & \hlthird{54.29} & \hlthird{54.01} & \hlthird{54.15} \\
\textbf{\ModelNameTurboeight{}} & \hlbest{55.19} & \hlbest{58.24} & \hlbest{56.33} & \hlsecond{57.08} & \hlbest{55.99} \\
% \textbf{\ModelNameTurbofour{}} & 53.33 & 56.90 & 55.26 & 53.63 & 54.14 \\
\bottomrule
\end{tabular}
\caption{Detailed evaluation results on Qwen-Image-Bench. Closed-source models are excluded from rank highlighting. Darker cell colors indicate better performance, with the three highlighted shades denoting the 1st, 2nd, and 3rd best results, respectively.}
\label{tab:qwenbench_detailed}
\end{table}

\begin{table}[h!]
\centering
\fontsize{8pt}{10pt}\selectfont
\renewcommand{\arraystretch}{1.0}
\setlength{\tabcolsep}{2mm}
\begin{tabular}{l|cccccccc}
\toprule
\multirow{2}{*}{\textbf{Model Name}}
& \multicolumn{2}{c}{\textbf{Alignments}} & \textbf{Aesthetics} & \multicolumn{5}{c}{\textbf{Visual Text Rendering}} \\
\cmidrule(lr){2-3} \cmidrule(lr){4-4} \cmidrule(lr){5-9}
& QA$\uparrow$ & COT$\uparrow$ & AM$\uparrow$ & $\text{Sim}_{\text{edit}}$$\uparrow$ & $\text{Acc}_{\text{sen}}$$\uparrow$ & $\text{Score}_{\text{qua}}$$\uparrow$ & $\text{GNED}$$\uparrow$ & $\text{Char}_{\text{F1}}$$\uparrow$ \\
\midrule
GPT-Image-2 & 0.97 & 8.95 & 2.84 & 0.50 & 0.57 & 0.88 & 0.52 & 0.56 \\
Nano-Banana-Pro & 0.97 & 8.71 & 2.77 & 0.38 & 0.55 & 0.86 & 0.45 & 0.49 \\
Qwen-Image-3.0 & 0.96 & 8.76 & 2.76 & 0.51 & 0.46 & 0.92 & 0.55 & 0.58 \\
Seedream-5.0-Pro & 0.97 & 8.87 & 2.57 & - & - & - & - & - \\
\midrule
Qwen-Image~ & 0.91 & 7.44 & 2.56 & 0.45 & 0.33 & 0.61 & 0.43 & 0.46 \\
Qwen-Image-2512  & 0.90 & 7.43 & \hlbest{2.70} & 0.62 & 0.43 & 0.72 & 0.57 & 0.59 \\
Z-Image-Turbo  & 0.90 & 7.26 & \hlthird{2.67} & 0.67 & 0.50 & 0.87 & 0.69 & 0.73 \\
HunyuanImage-3.0-Instruct-Distil & \hlthird{0.93} & 8.00 & 2.62 & 0.66 & 0.50 & 0.85 & 0.67 & 0.70 \\
HunyuanImage-3.0-Instruct-Distil-CoT& \hlsecond{0.94} & \hlbest{8.08} & \hlbest{2.70} & 0.57 & 0.54 & \hlthird{0.89} & 0.57 & 0.63 \\
ERNIE-Image-Turbo & 0.92 & 7.87 & 2.65 & \hlthird{0.70} & \hlthird{0.56} & \hlthird{0.89} & \hlsecond{0.71} & \hlsecond{0.75} \\
ERNIE-Image-Turbo-PE & 0.92 & 7.72 & 2.59 & 0.60 & \hlthird{0.56} & \hlsecond{0.90} & 0.61 & 0.65 \\
GLM-Image & 0.85 & 6.23 & 2.58 & \hlsecond{0.73} & \hlthird{0.56} & \hlbest{0.92} & \hlbest{0.74} & \hlbest{0.78} \\
HiDream-O1-Image & 0.87 & 6.11 & 2.30 & 0.49 & 0.54 & 0.79 & 0.56 & 0.57 \\
SenseNova-U1-8B-MoT & \hlthird{0.93} & 7.72 & 2.41 & 0.68 & 0.51 & 0.88 & \hlthird{0.70} & 0.72 \\
SenseNova-U1-8B-MoT-CoT & 0.92 & 7.81 & 2.38 & 0.68 & 0.52 & \hlsecond{0.90} & \hlthird{0.70} & \hlthird{0.74} \\
Boogu-Image-0.1-Base & \hlthird{0.93} & 7.93 & 2.64 & 0.64 & 0.53 & 0.84 & 0.69 & 0.72 \\
Boogu-Image-0.1-Turbo & \hlthird{0.93} & \hlthird{8.01} & 2.54 & 0.55 & 0.51 & 0.86 & 0.65 & 0.70 \\
FLUX.2-dev  & \hlthird{0.93} & 7.91 & \hlsecond{2.69} & 0.51 & 0.38 & 0.76 & 0.55 & 0.58 \\
\midrule
\textbf{\ModelName{}} & \hlbest{0.95} & 7.96 & 2.65 & \hlbest{0.76} & \hlbest{0.64} & 0.84 & \hlbest{0.74} & 0.73 \\
\textbf{\ModelNameTurboeight{}} & \hlbest{0.95} & \hlsecond{8.02} & \hlsecond{2.69} & \hlthird{0.70} & \hlsecond{0.58} & 0.86 & 0.61 & 0.64 \\
% \textbf{\ModelNameTurbofour{}} & 0.95 & 7.93 & 2.67 & 0.68 & 0.56 & 0.80 & 0.61 & 0.63 \\
\bottomrule
\end{tabular}
\caption{Detailed evaluation results on WeGenBench. Closed-source models are excluded from rank highlighting. Darker cell colors indicate better performance, with the three highlighted shades denoting the 1st, 2nd, and 3rd best results, respectively.}
\label{tab:wegenbench_detailed}
\end{table}

\paragraph{DPG-Bench} DPG-Bench (Dense Prompt Graph Benchmark)~\cite{hu2024ella} is introduced for evaluating text-to-image models' ability to follow dense, complex prompts, consisting of 1,000 dense prompts covering multiple objects, detailed attributes, complex relationships, and long-text alignment. Compared to prompts used in earlier text-to-image benchmarks, DPG-Bench prompts are notably longer and more information-rich, making it particularly effective at evaluating multiple object compositions involving diverse attributes and relationships, and serving as a widely-adopted testbed for assessing dense prompt-following capability in modern text-to-image generation models. Table~\ref{tab:dpg_detailed} shows that \ModelName{} achieves a top-three Attribute score among open-source models, reflecting strong fine-grained attribute understanding. Its consistently high Global, Relation, and Overall scores further demonstrate that \ModelName{} maintains competitive dense-prompt alignment across DPG-Bench.

\begin{table}[h!]
\centering
\fontsize{8pt}{10pt}\selectfont
\renewcommand{\arraystretch}{1.0}
\setlength{\tabcolsep}{2.8mm}
\begin{tabular}{l|ccc|ccc}
\toprule
\multirow{2}{*}{\textbf{Model Name}} & \multicolumn{3}{c|}{\textbf{LongText-Bench}} & \multicolumn{3}{c}{\textbf{CVTG-2K}} \\
\cmidrule(lr){2-4}
\cmidrule(lr){5-7}
& English & Chinese & Overall &  NED & CLIPScore & Word Accuracy  \\
\midrule
GPT-Image-2 & 0.981 & 0.963 & 0.972 & 0.950 & 0.775 & 0.901 \\
Nano-Banana-Pro & 0.975 & 0.854 & 0.914 & 0.875 & 0.751 & 0.779 \\
Qwen-Image-3.0 & 0.981 & 0.979 & 0.980 & 0.951 & 0.759 & 0.907 \\
Seedream-5.0-Pro & 0.943 & 0.951 & 0.947 & 0.860 & 0.761 & 0.762 \\
\midrule
Qwen-Image  & 0.936 & 0.932 & 0.934 & 0.915 & \hlbest{0.813} & 0.834 \\
Qwen-Image-2512 & 0.946 & 0.961 & 0.954 & 0.917 & 0.785 & 0.835 \\
Z-Image-Turbo  & 0.920 & 0.935 & 0.928 & 0.921 & 0.802 & 0.846 \\
HunyuanImage-3.0-Instruct-Distil & 0.874 & 0.899 & 0.886 & 0.869 & 0.799 & 0.743 \\
HunyuanImage-3.0-Instruct-Distil-CoT& 0.914 & 0.893 & 0.903 & 0.918 & 0.792 & 0.857 \\
ERNIE-Image-Turbo & \hlsecond{0.964} & \hlthird{0.966} & \hlsecond{0.965} & 0.927 & 0.775 & 0.852 \\
ERNIE-Image-Turbo-PE & \hlbest{0.970} & 0.963 & \hlbest{0.967} & 0.922 & 0.766 & 0.857 \\
GLM-Image  & 0.912 & \hlbest{0.972} & 0.942 & 0.943 & 0.791 & 0.886 \\
HiDream-O1-Image  & 0.958 & \hlsecond{0.971} & \hlsecond{0.965} & 0.937 & 0.789 & 0.877 \\
SenseNova-U1-8B-MoT & \hlthird{0.962} & 0.756 & 0.859 & \hlsecond{0.950} & \hlsecond{0.812} & \hlthird{0.895} \\
SenseNova-U1-8B-MoT-CoT & 0.949 & 0.822 & 0.885 & \hlbest{0.953} & \hlthird{0.811} & \hlbest{0.902} \\
Boogu-Image-0.1-Base & 0.951 & \hlsecond{0.971} & \hlthird{0.961} & 0.931 & 0.801 & 0.868 \\
Boogu-Image-0.1-Turbo & 0.944 & \hlsecond{0.971} & 0.958 & 0.905 & 0.789 & 0.810 \\
FLUX.2-dev  & 0.954 & 0.748 & 0.851 & 0.921 & 0.796 & 0.849 \\
\midrule
\textbf{\ModelName{}}  & 0.921 & 0.940 & 0.931 & \hlthird{0.948} & 0.760 & \hlsecond{0.899} \\
\textbf{\ModelNameTurboeight{}}  & 0.912 & 0.955 & 0.933 & 0.943 & 0.776 & 0.872 \\
% \textbf{\ModelNameTurbofour{}}  & 0.853 & 0.905 & 0.879 & 0.874 & 0.748 & 0.741 \\
\bottomrule
\end{tabular}
\caption{Detailed evaluation results on LongText-Bench and CVTG-2K. Closed-source models are excluded from rank highlighting. Darker cell colors indicate better performance, with the three highlighted shades denoting the 1st, 2nd, and 3rd best results, respectively.}
\label{tab:longtext_cvtg2k_detailed}
\end{table}

\paragraph{Qwen-Image-Bench}
Qwen-Image-Bench~\cite{li2026qwen} is designed to evaluate comprehensive text-to-image generation capabilities, with a strong emphasis on bilingual (English and Chinese) prompt understanding and culturally relevant rendering. It tests the model's ability to handle diverse scenarios, ranging from everyday objects to complex abstract concepts and culturally specific elements. Table~\ref{tab:qwenbench_detailed} reveals that the best-performing \ModelName{} variant, \ModelNameTurboeight{}, achieves the top open-source Overall score on Qwen-Image-Bench, leading across Quality, Aesthetics, and Alignment while remaining highly competitive in Creative Generation. 

\paragraph{WeGenBench} WeGenBench~\cite{liang2026wegenbench} is a multidimensional diagnostic benchmark for fine-grained, multi-perspective evaluation of text-to-image generation. It comprises 4,000 prompts balanced between Chinese and English across two primary categories, each annotated with multi-dimensional, language-specific tags for sub-category evaluation. Combined with novel Vision-Language-Model-based metrics, this cross-dimensional mechanism enables precise diagnosis of model shortcomings, guiding text-to-image model optimization. Table~\ref{tab:wegenbench_detailed} shows the \ModelName{} family leading in both QA score and sentence-level text accuracy, demonstrating strong performance on both alignment-oriented evaluation and visual text rendering.

\paragraph{LongText-Bench}
LongText-Bench~\cite{geng2025x} is a benchmark from the X-Omni project for evaluating long-text rendering in text-to-image generation, covering both English and Chinese. It comprises 320 prompts spanning eight content categories (e.g., road signs, posters, slides, dialogues) and two length tiers—short (10–30 English words / 20–40 Chinese characters) and long (30–50 English words / 60+ Chinese characters)—penalizing spelling errors, missing characters, and layout collapse in extended text sequences. Table~\ref{tab:longtext_cvtg2k_detailed} shows that \ModelName{} maintains balanced performance across English and Chinese long-text generation. Although the benchmark is highly competitive, its stable scores indicate reliable text rendering ability across languages.

\paragraph{CVTG-2K} CVTG-2K~\cite{du2025textcrafter} is a benchmark for Complex Visual Text Generation (CVTG), evaluating models' ability to render multiple text regions within a single image. It contains 2,000 diverse prompts (e.g., posters, book covers, logos) with substantially longer text than prior benchmarks (avg. 8.10 words, 39.47 characters), and is the first to vary text region count (2–5 per image) while randomly assigning size, color, and font attributes to half the samples. Table~\ref{tab:longtext_cvtg2k_detailed} further demonstrates the strength of \ModelName{} on CVTG-2K, where it ranks among the top models in NED and Word Accuracy. These results suggest that \ModelName{} can generate visually grounded text with strong character-level correctness and semantic consistency.

\subsubsection{Image Editing}

Image editing requires a model to faithfully execute user instructions while preserving subjects, structures, and visual details that are not intended to change. To evaluate the image editing capability of \ModelName{}, we benchmark it against leading proprietary models, including GPT-Image-2, Nano-Banana-Pro, Qwen-Image-3.0, and Seedream-5.0-Pro, as well as a broad set of recent open-source editing models, such as HunyuanImage-3.0-Instruct-Distil~\cite{hunyuanimage2025},  FireRed-Image-Edit-1.1~\cite{firered2026rededit}, Qwen-Image-Edit-2511~\cite{wu2025qwenimage}, JoyAI-Image-Edit~\cite{song2026joyai}, Boogu-Image-0.1-Edit~\cite{chen2026booguimage01}, LongCat-Image-Edit~\cite{team2025longcat}, SenseNova-U1-8B-MoT~\cite{diao2026sensenova}, HiDream-O1-Image~\cite{cai2026hidreamo1}, and FLUX.2-dev~\cite{flux-2-2025}. 

\paragraph{Single-image Editing} We report results on four complementary benchmarks, including GEdit~\cite{liu2025step1x}, ImgEdit~\cite{ye2025imgedit}, UnicBench~\cite{ye2026unicedit}, and our internally developed WeGenEditBench, which together assess general editing quality, instruction following, and in-domain editing robustness. As summarized in Table~\ref{tab:alleditbench}, \ModelName{} achieves the best performance among open-source models on both UnicBench and WeGenEditBench, while ranking among the top three on GEdit and ImgEdit, demonstrating consistently strong and balanced editing capability.

\begin{table}[h!]
\centering
\fontsize{8pt}{10pt}\selectfont
\renewcommand{\arraystretch}{1.0}
\setlength{\tabcolsep}{3mm}
\begin{tabular}{l|cccc}
\toprule
\textbf{Model Name} & \textbf{GEdit} $\uparrow$ & \textbf{ImgEdit} $\uparrow$ & \textbf{UnicBench} $\uparrow$ & \textbf{WeGenEditBench} $\uparrow$ \\
\midrule
GPT-Image-2 & 8.580 & 4.665 & 9.079 & 4.104 \\
Nano-Banana-Pro & 8.264 & 4.455 & 8.660 & 3.999 \\
Qwen-Image-3.0 & 8.521 & 4.597 & 8.903 & 3.961 \\
Seedream-5.0-Pro & 8.460 & 4.650 & 8.906 & 4.160 \\
\midrule
HunyuanImage-3.0-Instruct-Distil~\cite{hunyuanimage2025} & 8.101 & 4.552 & 8.481 & 3.791 \\
HunyuanImage-3.0-Instruct-Distil-CoT~\cite{hunyuanimage2025} & \hlbest{8.385} & 4.523 & 8.577 & \hlthird{3.872} \\
FireRed-Image-Edit-1.1~\cite{firered2026rededit} & 8.130 & \hlbest{4.616} & \hlsecond{8.723} & 3.580 \\
Qwen-Image-Edit-2511~\cite{wu2025qwenimage} & 7.924 & 4.524 & 8.472 & 3.508 \\
JoyAI-Image-Edit~\cite{song2026joyai} & 8.159 & \hlsecond{4.594} & 8.353 & 3.648 \\
Boogu-Image-0.1-Edit~\cite{chen2026booguimage01} & 7.532 & 4.351 & 7.885 & 3.523 \\
LongCat-Image-Edit~\cite{team2025longcat} & 7.855 & 4.449 & 8.064 & 3.156 \\
SenseNova-U1-8B-MoT-CoT~\cite{diao2026sensenova} & 7.600 & 4.162 & 8.030 & 3.649 \\
SenseNova-U1-8B-MoT~\cite{diao2026sensenova} & 7.524 & 4.050 & 7.926 & 3.643 \\
HiDream-O1-Image~\cite{cai2026hidreamo1} & 7.256 & 4.084 & 6.646 & 2.659 \\
FLUX.2-dev~\cite{flux-2-2025} & 4.347 & 2.890 & 3.156 & 1.984 \\
\midrule
\textbf{\ModelName{}} & \hlsecond{8.381} & \hlthird{4.581} & \hlbest{8.992} & \hlbest{4.005} \\
\textbf{\ModelNameTurboeight{}} & \hlthird{8.314} & 4.537 & \hlthird{8.664} & \hlsecond{3.947} \\
% \textbf{\ModelNameTurbofour{}} & 7.999 & 4.430 & 8.552 & 3.992 \\
\bottomrule
\end{tabular}
\caption{Quantitative comparison across single-image editing benchmarks. GEdit reports the average of the English and Chinese overall scores. ImgEdit and UnicBench report benchmark-level overall scores. WeGenEditBench reports the average of the text-editing and single-image editing subset scores. Closed-source models are excluded from rank highlighting. Darker cell colors indicate better performance, with the three highlighted shades denoting the 1st, 2nd, and 3rd best results, respectively.}
\label{tab:alleditbench}
\end{table}

\begin{table*}[h]
\centering
\fontsize{8pt}{12pt}\selectfont
\renewcommand{\arraystretch}{1.0}
\setlength{\tabcolsep}{0.9mm}
\begin{tabular}{l|cccc|ccc|cccc}
\toprule
\multirow{2}{*}{\textbf{Model Name}} & \multicolumn{4}{c|}{\textbf{MICo}} & \multicolumn{3}{c|}{\textbf{OmniContext}} & \multicolumn{4}{c}{\textbf{WeGenEditBench-Multi}} \\
\cmidrule(lr){2-5}\cmidrule(lr){6-8}\cmidrule(lr){9-12}
& W $\uparrow$ & SC $\uparrow$ & PQ $\uparrow$ & Overall $\uparrow$ & PF $\uparrow$ & SC $\uparrow$ & Overall $\uparrow$ & EG $\uparrow$ & PR $\uparrow$ & VQ $\uparrow$ & Overall $\uparrow$ \\
\midrule
GPT-Image-2 & 0.945 & 8.36 & 8.01 & 63.32 & 8.92 & 7.86 & 8.35 & 4.358 & 4.134 & 3.707 & 4.138 \\
Nano-Banana-Pro & 0.902 & 8.35 & 7.82 & 59.14 & 8.77 & 7.58 & 8.13 & 4.031 & 3.993 & 3.340 & 3.877 \\
Qwen-Image-3.0 & - & - & - & - & 8.76 & 7.76 & 8.23 & - & - & - & - \\
Seedream-5.0-Pro & 0.938 & 8.19 & 7.86 & 60.56 & 8.74 & 7.62 & 8.13 & 4.270 & 4.093 & 3.750 & 4.095 \\
\midrule
HunyuanImage-3.0-Instruct-Distil & \hlthird{0.827} & \hlthird{8.01} & \hlthird{7.64} & \hlthird{51.23} & \hlbest{8.50} & \hlthird{7.59} & \hlsecond{7.99} & \hlthird{3.279} & \hlthird{3.493} & 3.091 & \hlthird{3.327} \\
FireRed-Image-Edit-1.1 & 0.823 & 7.76 & 7.40 & 47.60 & 8.35 & 7.03 & 7.62 & 2.714 & 2.977 & 3.055 & 2.887 \\
Qwen-Image-Edit-2511 & 0.727 & 6.63 & 6.97 & 36.49 & \hlsecond{8.48} & 7.07 & 7.69 & 2.589 & 2.884 & \hlthird{3.093} & 2.808 \\
\midrule
\textbf{\ModelName{}} & \hlsecond{0.853} & \hlbest{8.22} & \hlbest{7.84} & \hlsecond{55.35} & \hlthird{8.43} & \hlsecond{7.64} & \hlthird{7.97} & \hlbest{4.004} & \hlbest{4.004} & \hlsecond{3.511} & \hlbest{3.905} \\
\textbf{\ModelNameTurboeight{}} & \hlbest{0.926} & \hlsecond{8.11} & \hlsecond{7.71} & \hlbest{58.18} & 8.23 & \hlbest{7.84} & \hlbest{8.00} & \hlsecond{3.756} & \hlsecond{3.930} & \hlbest{3.629} & \hlsecond{3.800} \\
% \textbf{\ModelNameTurbofour{}} & 0.907 & 7.93 & 7.04 & 51.12 & 8.04 & 7.52 & 7.73 & 3.470 & 3.724 & 3.164 & 3.510 \\
\bottomrule
\end{tabular}
\caption{Quantitative comparison across three multi-image editing benchmarks: MICo, OmniContext, and WeGenEditBench-Multi. Closed-source models are included for reference only and excluded from rank highlighting. Darker cell colors indicate better performance, with the three highlighted shades denoting the 1st, 2nd, and 3rd best results, respectively. ``--'' denotes unavailable results due to input image limits.}
\label{tab:multi-benchmark-comparison}
\end{table*}

\paragraph{Multi-image Editing} To provide a comprehensive assessment, we evaluate the models on MICo, OmniContext, and the multi-image subset of our WeGenEditBench (WeGenEditBench-Multi), covering multi-image reasoning, contextual consistency, reference grounding, and instruction-following quality.
From Table~\ref{tab:multi-benchmark-comparison}, the \ModelName{} family ranks first among open-source models on all MICo metrics, with \ModelName{} also achieving the best EG, PR, and Overall scores on WeGenEditBench-Multi.
When the strongest variant is considered for each metric, \ModelNameTurboeight{} further obtains the top VQ score on WeGenEditBench-Multi, highlighting robust multi-image editing quality across complementary settings.

\begin{table*}[t!]
\centering
\fontsize{8pt}{12pt}\selectfont
\renewcommand{\arraystretch}{1.0}
\setlength{\tabcolsep}{3mm}
\begin{tabular}{l|ccc|ccc}
\toprule
\multirow{2}{*}{\textbf{Model Name}} & \multicolumn{3}{c|}{\textbf{English}} & \multicolumn{3}{c}{\textbf{Chinese}} \\
\cmidrule(lr){2-4}\cmidrule(lr){5-7}
& SC $\uparrow$ & PQ $\uparrow$ & Overall $\uparrow$ & SC $\uparrow$ & PQ $\uparrow$ & Overall $\uparrow$ \\
\midrule
GPT-Image-2 & 9.310 & 8.296 & 8.657 & 9.134 & 8.330 & 8.503 \\
Nano-Banana-Pro & 8.927 & 8.129 & 8.253 & 8.760 & 8.331 & 8.276 \\
Qwen-Image-3.0 & 9.181 & 8.204 & 8.536 & 9.207 & 8.163 & 8.506 \\
Seedream-5.0-Pro & 9.068 & 8.274 & 8.426 & 9.109 & 8.299 & 8.494 \\
\midrule
HunyuanImage-3.0-Instruct-Distil & 8.908 & \hlthird{8.052} & 8.169 & 8.692 & \hlsecond{8.105} & 8.033 \\
HunyuanImage-3.0-Instruct-Distil-CoT & \hlbest{9.223} & \hlsecond{8.095} & \hlbest{8.466} & \hlsecond{9.070} & \hlthird{8.068} & \hlthird{8.304} \\
Qwen-Image-Edit-2511 & 8.768 & 7.631 & 7.903 & 8.782 & 7.621 & 7.945 \\
JoyAI-Image-Edit & 8.976 & 7.771 & 8.183 & 8.803 & 7.898 & 8.135 \\
FireRed-Image-Edit-1.1 & \hlthird{9.064} & 7.802 & 8.215 & \hlthird{8.840} & 7.814 & 8.045 \\
LongCat-Image-Edit & 8.652 & 7.749 & 7.943 & 8.584 & 7.539 & 7.767 \\
SenseNova-U1-8B-MoT & 8.497 & 6.845 & 7.353 & 8.687 & 7.212 & 7.695 \\
SenseNova-U1-8B-MoT-CoT & 8.732 & 6.942 & 7.567 & 8.752 & 7.051 & 7.633 \\
Boogu-Image-0.1-Edit & 8.626 & 7.329 & 7.644 & 8.371 & 7.250 & 7.420 \\
HiDream-O1-Image & 8.363 & 7.115 & 7.392 & 7.796 & 7.428 & 7.121 \\
FLUX.2-dev & 4.446 & 7.615 & 4.442 & 4.226 & 7.482 & 4.252 \\
\midrule
\textbf{\ModelName{}} & \hlsecond{9.149} & 8.015 & \hlsecond{8.352} & \hlbest{9.172} & 8.051 & \hlbest{8.410} \\
\textbf{\ModelNameTurboeight{}} & 8.682 & \hlbest{8.408} & \hlthird{8.297} & 8.722 & \hlbest{8.421} & \hlsecond{8.332} \\
% \textbf{\ModelNameTurbofour{}} & 8.432 & 8.064 & 7.983 & 8.506 & 8.067 & 8.014 \\
\bottomrule
\end{tabular}
\caption{Quantitative comparison on GEdit. Closed-source models are included for reference only and excluded from rank highlighting. Darker cell colors indicate better performance, with the three highlighted shades denoting the 1st, 2nd, and 3rd best results, respectively.}
\label{tab:gedit}
\end{table*}

\paragraph{GEdit} GEdit~\cite{liu2025step1x} is a real-world instruction-based image editing benchmark that addresses the limitations of prior benchmarks relying on synthetic instructions. It contains 606 test samples (1,212 total across English and Chinese subsets) collected from genuine user editing requests, spanning 11 editing task categories, enabling authentic and comprehensive evaluation of image editing models. Table~\ref{tab:gedit} shows that \ModelName{} consistently ranks among the top open-source models across both English and Chinese evaluations, even leading on the Chinese subset, highlighting its strong and balanced bilingual editing capability.

\begin{table*}[t!]
\centering
\fontsize{8pt}{12pt}\selectfont
\renewcommand{\arraystretch}{1.0}
\setlength{\tabcolsep}{1.1mm}
\begin{tabular}{l|ccccccccc|c}
\toprule
\textbf{Model Name} & Action & Add & Adjust & BG & Compose & Extract & Remove & Replace & Style & Overall \\
\midrule
% Closed-source models: no color
GPT-Image-2 & 4.89 & 4.87 & 4.75 & 4.70 & 3.88 & 4.40 & 4.83 & 4.76 & 4.89 & 4.665 \\
Nano-Banana-Pro & 4.72 & 4.62 & 4.64 & 4.40 & 3.78 & 3.79 & 4.73 & 4.53 & 4.87 & 4.455 \\
Qwen-Image-3.0 & 4.90 & 4.68 & 4.71 & 4.50 & 4.14 & 4.19 & 4.78 & 4.75 & 4.73 & 4.597 \\
Seedream-5.0-Pro & 4.78 & 4.79 & 4.78 & 4.73 & 4.43 & 3.79 & 4.89 & 4.76 & 4.90 & 4.650 \\
\midrule
HunyuanImage-3.0-Instruct-Distil & \hlthird{4.81} & 4.70 & \hlsecond{4.74} & 4.51 & 3.74 & 3.97 & \hlsecond{4.83} & \hlsecond{4.78} & \hlsecond{4.88} & 4.552 \\
HunyuanImage-3.0-Instruct-Distil-CoT & 4.63 & 4.76 & \hlbest{4.84} & \hlthird{4.53} & 3.42 & 4.01 & \hlbest{4.87} & \hlthird{4.77} & \hlthird{4.87} & 4.523 \\
Qwen-Image-Edit-2511 & \hlsecond{4.86} & 4.73 & 4.59 & 4.43 & 3.96 & \hlthird{4.29} & 4.71 & 4.44 & 4.70 & 4.524 \\
JoyAI-Image-Edit & \hlbest{4.87} & \hlsecond{4.80} & 4.70 & \hlsecond{4.54} & 3.68 & \hlsecond{4.46} & \hlthird{4.80} & 4.67 & 4.81 & \hlsecond{4.594} \\
FireRed-Image-Edit-1.1 & 4.67 & 4.75 & \hlsecond{4.74} & 4.49 & \hlthird{4.00} & \hlbest{4.60} & 4.68 & 4.73 & \hlsecond{4.88} & \hlbest{4.616} \\
LongCat-Image-Edit & 4.75 & 4.62 & 4.60 & 4.37 & 3.77 & 4.00 & 4.55 & 4.53 & 4.85 & 4.449 \\
SenseNova-U1-8B-MoT & 4.55 & 4.05 & 4.38 & 4.13 & 3.38 & 3.19 & 3.47 & 4.56 & 4.75 & 4.050 \\
SenseNova-U1-8B-MoT-CoT & 4.63 & 4.12 & 4.35 & 4.14 & 3.59 & 3.29 & 3.94 & 4.63 & 4.76 & 4.162 \\
Boogu-Image-0.1-Edit & 4.55 & 4.68 & 4.45 & 4.19 & 3.36 & 4.06 & 4.49 & 4.62 & 4.77 & 4.351 \\
HiDream-O1-Image & 4.56 & 4.52 & 4.28 & 4.32 & 3.35 & 2.19 & 4.16 & 4.57 & 4.81 & 4.084 \\
FLUX.2-dev & 3.31 & 3.52 & 2.60 & 3.10 & 2.32 & 1.90 & 2.28 & 3.02 & 3.98 & 2.890 \\
\midrule
\textbf{\ModelName{}} & 4.36 & \hlbest{4.89} & \hlthird{4.73} & \hlthird{4.53} & \hlbest{4.20} & 4.28 & 4.63 & 4.73 & \hlthird{4.87} & \hlthird{4.581} \\
\textbf{\ModelNameTurboeight{}} & 4.28 & \hlthird{4.78} & 4.67 & \hlbest{4.60} & \hlsecond{4.07} & 4.22 & 4.52 & \hlbest{4.80} & \hlbest{4.89} & 4.537 \\
% \textbf{\ModelNameTurbofour{}} & 4.08 & 4.72 & 4.80 & 4.52 & 3.86 & 4.28 & 4.36 & 4.63 & 4.62 & 4.430 \\
\bottomrule
\end{tabular}
\caption{Quantitative comparison on ImgEdit. The nine subtask scores are reported with two decimal places, while Overall is reported with three decimal places. Closed-source models are included for reference only and excluded from rank highlighting. Darker cell colors indicate better performance, with the three highlighted shades denoting the 1st, 2nd, and 3rd best results, respectively.}
\label{tab:imgedit}
\end{table*}

\begin{table*}[h!]
\centering
\fontsize{8pt}{10pt}\selectfont
\renewcommand{\arraystretch}{1.0}
\setlength{\tabcolsep}{4mm}
\begin{tabular}{l|ccccc}
\toprule
\textbf{Model Name} & IF $\uparrow$ & NC $\uparrow$ & VQ $\uparrow$ & RA $\uparrow$ & Overall $\uparrow$ \\
\midrule
% Closed-source models: no color
GPT-Image-2 & 9.626 & 8.979 & 8.877 & 9.000 & 9.079 \\
Nano-Banana-Pro & 9.233 & 8.693 & 8.473 & 8.210 & 8.660 \\
Qwen-Image-3.0 & 9.252 & 9.264 & 8.643 & 8.040 & 8.903 \\
Seedream-5.0-Pro & 9.309 & 9.021 & 8.735 & 8.289 & 8.906 \\
\midrule
HunyuanImage-3.0-Instruct-Distil & 8.783 & 9.085 & 8.261 & 7.264 & 8.481 \\
HunyuanImage-3.0-Instruct-Distil-CoT & 8.934 & \hlthird{9.104} & 8.263 & 7.644 & 8.577 \\
Qwen-Image-Edit-2511 & \hlthird{9.011} & 8.533 & 8.301 & 7.844 & 8.472 \\
JoyAI-Image-Edit & 8.883 & 8.418 & 8.221 & 7.896 & 8.353 \\
FireRed-Image-Edit-1.1 & \hlsecond{9.124} & 9.013 & \hlsecond{8.446} & \hlsecond{8.048} & \hlsecond{8.723} \\
LongCat-Image-Edit & 8.572 & 8.065 & 8.034 & 7.524 & 8.064 \\
SenseNova-U1-8B-MoT & 8.692 & 7.959 & 7.439 & 7.832 & 7.926 \\
SenseNova-U1-8B-MoT-CoT & 8.743 & 8.043 & 7.573 & \hlthird{7.912} & 8.030 \\
Boogu-Image-0.1-Edit & 8.114 & 8.806 & 7.709 & 6.851 & 7.885 \\
HiDream-O1-Image & 7.212 & 7.245 & 6.582 & 5.208 & 6.646 \\
FLUX.2-dev & 2.985 & 5.066 & 4.336 & 1.512 & 3.156 \\
\midrule
\textbf{\ModelName{}} & \hlbest{9.381} & \hlbest{9.310} & \hlbest{8.644} & \hlbest{8.320} & \hlbest{8.992} \\
\textbf{\ModelNameTurboeight{}} & 8.905 & \hlsecond{9.234} & \hlthird{8.388} & 7.760 & \hlthird{8.664} \\
% \textbf{\ModelNameTurbofour{}} & 8.791 & 9.010 & 8.231 & 7.381 & 8.552 \\
\bottomrule
\end{tabular}
\caption{Quantitative comparison on UnicBench. Closed-source models are included for reference only and excluded from rank highlighting. Darker cell colors indicate better performance, with the three highlighted shades denoting the 1st, 2nd, and 3rd best results, respectively.}
\label{tab:unicbench}
\end{table*}

\paragraph{ImgEdit} ImgEdit~\cite{ye2025imgedit} is a comprehensive benchmark comprising 811 test cases across 14 sub-tasks, organized into three suites of increasing difficulty: a Basic-Edit Suite (734 cases) covering nine common editing operations (e.g., addition, removal, replacement, style transfer, background change) across six object categories; a 47-case Understanding-Grounding-Editing Suite targeting complex spatial reasoning and multi-object edits; and a Multi-Turn Suite assessing content memory, content understanding, and version backtracking over three-turn dialogues. Together, these suites enable fine-grained evaluation of instruction adherence, editing quality, and detail preservation. Table~\ref{tab:imgedit} shows that the \ModelName{} family achieves top-three open-source performance on six ImgEdit subtasks and ranks third on the overall score.
Notably, \ModelNameTurboeight{} leads Background, Replace, and Style, while \ModelName{} attains the best Add and Compose scores, demonstrating strong coverage across diverse editing operations.

\paragraph{UnicBench} UnicBench~\cite{ye2026unicedit} is a comprehensive image-editing benchmark covering 1,100 manually reviewed test cases across 22 sub-tasks spanning four major categories: Object, Attribute, Scene, and Reasoning Editing. Beyond conventional evaluation, it introduces Non-edit Consistency and Reasoning Accuracy metrics to enable fine-grained diagnosis of both basic and complex, reasoning-driven editing capabilities. Table~\ref{tab:unicbench} highlights that \ModelName{} ranks first among open-source models across all UnicBench metrics.
Together with \ModelNameTurboeight{} placing second on NC and third on both VQ and Overall, the \ModelName{} family shows strong instruction alignment and visual quality.

\begin{table*}[t!]
\centering
\fontsize{8pt}{10pt}\selectfont
\renewcommand{\arraystretch}{1.0}
\setlength{\tabcolsep}{2.5mm}
\begin{tabular}{l|ccccc}
\toprule
\textbf{Model Name} & EG $\uparrow$ & PR $\uparrow$ & VQ $\uparrow$ & DS $\uparrow$ & \textbf{Overall} $\uparrow$ \\
\midrule
% Closed-source models
GPT-Image-2 & 4.298 & 4.049 & 4.463 & 4.182 & 4.076 \\
Nano-Banana-Pro & 4.159 & 3.752 & 4.370 & 4.050 & 3.882 \\
Qwen-Image-3.0 & 4.120 & 3.469 & 4.331 & 4.126 & 3.813 \\
Seedream-5.0-Pro & 4.207 & 4.084 & 4.434 & 4.206 & 4.029 \\
\midrule
HunyuanImage-3.0-Instruct-Distil & 4.020 & \hlsecond{4.175} & \hlthird{4.280} & 3.952 & 3.841 \\
HunyuanImage-3.0-Instruct-Distil-CoT & \hlthird{4.071} & \hlthird{4.065} & 4.265 & \hlthird{4.008} & \hlthird{3.862} \\
FireRed-Image-Edit-1.1 & \hlsecond{4.091} & 3.550 & 4.227 & 3.902 & 3.736 \\
JoyAI-Image-Edit & 4.042 & 3.547 & 4.169 & 3.794 & 3.667 \\
Qwen-Image-Edit-2511 & 3.963 & 3.567 & 4.095 & 3.708 & 3.588 \\
Boogu-Image-0.1-Edit & 3.850 & 3.718 & 4.092 & 3.768 & 3.564 \\
SenseNova-U1-8B-MoT-CoT & 3.948 & 3.636 & 3.932 & 3.605 & 3.535 \\
SenseNova-U1-8B-MoT & 4.030 & 3.408 & 3.917 & 3.674 & 3.552 \\
LongCat-Image-Edit & 3.761 & 3.439 & 4.003 & 3.643 & 3.410 \\
HiDream-O1-Image & 3.158 & 2.942 & 3.855 & 3.231 & 2.816 \\
FLUX.2-dev & 2.056 & 3.343 & 3.393 & 1.975 & 1.828 \\
\midrule
\textbf{\ModelName{}} & \hlbest{4.216} & 4.043 & \hlsecond{4.374} & \hlsecond{4.116} & \hlbest{3.992} \\
\textbf{\ModelNameTurboeight{}} & 4.033 & \hlbest{4.201} & \hlbest{4.548} & \hlbest{4.176} & \hlsecond{3.963} \\
% \textbf{\ModelNameTurbofour{}} & 4.029 & 4.256 & 4.541 & 4.019 & 3.932 \\
\bottomrule
\end{tabular}
\caption{Quantitative comparison on WeGenEditBench-single. EG, PR, VQ, and DS denote Edit Goal, Preservation, Visual Quality, and Domain-specific Capability, respectively. Closed-source models are included for reference only and excluded from rank highlighting. Darker cell colors indicate better performance, with the three highlighted shades denoting the 1st, 2nd, and 3rd best results, respectively.}
\label{tab:wegen-single}
\end{table*}

\paragraph{WeGenEditBench}
Although public benchmarks provide broad comparisons, they remain limited in their coverage of long instructions, compositional editing goals, fine-grained text modification, and multi-reference image composition. Their VLM-based evaluators also commonly assess an edited image holistically without explicit sample-specific verification criteria, allowing a salient successful edit or high overall visual quality to obscure omitted operations, unintended modifications, or incorrect reference binding.

To address these limitations, we construct WeGenEditBench, comprising 2,740 manually collected and curated samples. It contains 1,008 single-image editing samples covering local manipulation, portrait and aesthetic editing, and reasoning-intensive editing. The benchmark further includes 1,000 text editing samples covering textual-content modification, text-style adjustment, and complex visual layouts, together with 732 multi-image editing samples designed to evaluate identity and product references, style transfer, and multi-reference composition.

For each sample, we generate and freeze a fine-grained checklist before evaluating model outputs. Complex instructions are decomposed into independently verifiable requirements specifying the editing target, operation, attribute, location, and expected change. These requirements are organized into four dimensions: Edit Goal (\textbf{EG}), Preservation (\textbf{PR}), Visual Quality (\textbf{VQ}), and Domain-specific Capability (\textbf{DS}). A fixed VLM judge provides an explicit rationale and deduction for every checklist item before assigning the dimensional scores. All models are evaluated using identical pre-generated checklists, ensuring consistent and traceable evaluation criteria.

For single-image editing, the final score is defined as
\begin{equation}
S_{\mathrm{single}}
=
\operatorname{clip}
\left(
0.6S_{\mathrm{EG}}
+
0.2S_{\mathrm{PR}}
+
0.2S_{\mathrm{VQ}}
-
\frac{1}{|T|}
\sum_{t\in T}
\frac{5-S_{\mathrm{DS},t}}{4},
0,5
\right).
\end{equation}
For text editing, text-side and image-side scores are evaluated separately. When image-side requirements are present, the final score is
\begin{equation}
S_{\mathrm{text\text{-}edit}}
=
0.75S_{\mathrm{text}}
+
0.25S_{\mathrm{image}}.
\end{equation}
For multi-image editing, the final score is
\begin{equation}
S_{\mathrm{multi}}
=
0.4S_{\mathrm{EG}}
+
0.4S_{\mathrm{PR}}
+
0.2S_{\mathrm{VQ}}.
\end{equation}

On the single-image subset (WeGenEditBench-single), \ModelName{} demonstrates strong and balanced complex editing capabilities. Table~\ref{tab:wegen-single} shows that \ModelName{} leads EG and Overall, while the Turbo variant contributes the strongest PR, VQ, and DS results, indicating robust single-image editing quality.
For the text-editing subset of WeGenEditBench, Table~\ref{tab:wegen-text} jointly evaluates textual correctness and natural visual presentation. The \ModelName{} family secures the top two open-source Overall scores on WeGenEditBench-Text, with \ModelName{} delivering the strongest overall result—leading in text-side EG and DS as well as image-side DS—while \ModelNameTurboeight{} ranks second overall and tops both text-side and image-side PR and VQ, together demonstrating strong text-aware editing performance.

\begin{table*}[t!]
\centering
\fontsize{8pt}{12pt}\selectfont
\renewcommand{\arraystretch}{1.0}
\setlength{\tabcolsep}{1.2mm}
\begin{tabular}{l|cccc|cccc|c}
\toprule
\multirow{2}{*}{\textbf{Model Name}} & \multicolumn{4}{c|}{\textbf{Text-side}} & \multicolumn{4}{c|}{\textbf{Image-side}} & \multirow{2}{*}{\textbf{Overall}} \\
\cmidrule(lr){2-5}\cmidrule(lr){6-9}
& EG & PR & VQ & DS & EG & PR & VQ & DS & \\
\midrule
GPT-Image-2 & 4.289 & 4.031 & 4.329 & 4.399 & 4.604 & 3.919 & 4.421 & 4.452 & 4.132 \\
Nano-Banana-Pro & 4.295 & 4.092 & 4.310 & 4.322 & 4.396 & 4.208 & 4.232 & 4.306 & 4.116 \\
Qwen-Image-3.0 & 4.367 & 3.732 & 4.329 & 4.385 & 4.509 & 3.869 & 4.406 & 4.414 & 4.109 \\
Seedream-5.0-Pro & 4.415 & 4.327 & 4.416 & 4.504 & 4.362 & 4.207 & 4.455 & 4.429 & 4.291 \\
\midrule
SenseNova-U1-8B-MoT-CoT & \hlthird{4.039} & \hlthird{3.880} & 3.844 & 3.943 & 4.340 & 4.092 & 3.935 & 4.090 & 3.762 \\
HunyuanImage-3.0-Instruct-Distil & 3.914 & 3.801 & 3.917 & 4.133 & 4.361 & 4.177 & 4.109 & 4.214 & 3.742 \\
HunyuanImage-3.0-Instruct-Distil-CoT & \hlsecond{4.090} & 3.807 & \hlthird{4.102} & \hlthird{4.188} & 4.422 & \hlsecond{4.258} & \hlthird{4.194} & \hlsecond{4.323} & \hlthird{3.882} \\
SenseNova-U1-8B-MoT & 4.013 & 3.809 & 3.902 & 3.924 & 4.408 & 3.830 & 3.895 & 4.099 & 3.733 \\
JoyAI-Image-Edit & 3.900 & 3.617 & 3.751 & 3.975 & \hlbest{4.565} & 3.925 & 4.129 & \hlthird{4.287} & 3.630 \\
FireRed-Image-Edit-1.1 & 3.712 & 3.308 & 3.646 & 3.890 & \hlthird{4.497} & 3.622 & 3.932 & 4.245 & 3.423 \\
Qwen-Image-Edit-2511 & 3.658 & 3.583 & 3.655 & 3.820 & 4.384 & 3.830 & 3.969 & 4.112 & 3.429 \\
Boogu-Image-0.1-Edit & 3.842 & 3.180 & 3.732 & 3.915 & 4.412 & 3.459 & 3.898 & 4.197 & 3.482 \\
LongCat-Image-Edit & 3.027 & 3.560 & 3.385 & 3.320 & 4.187 & 3.639 & 3.789 & 3.949 & 2.901 \\
HiDream-O1-Image & 2.893 & 2.541 & 3.213 & 3.145 & 3.568 & 3.051 & 3.483 & 3.539 & 2.502 \\
FLUX.2-dev & 2.183 & 3.313 & 3.640 & 2.384 & 2.724 & 3.331 & 3.570 & 2.782 & 2.140 \\
\midrule
\textbf{\ModelName{}} & \hlbest{4.194} & \hlsecond{3.975} & \hlsecond{4.105} & \hlbest{4.293} & \hlsecond{4.510} & \hlthird{4.207} & \hlsecond{4.309} & \hlbest{4.374} & \hlbest{4.017} \\
\textbf{\ModelNameTurboeight{}} & 4.027 & \hlbest{4.082} & \hlbest{4.137} & \hlsecond{4.253} & 4.313 & \hlbest{4.313} & \hlbest{4.354} & 4.282 & \hlsecond{3.931} \\
% \textbf{\ModelNameTurbofour{}} & 4.110 & 4.160 & 4.261 & 4.326 & 4.443 & 4.344 & 4.467 & 4.516 & 4.052 \\
\bottomrule
\end{tabular}
\caption{Quantitative comparison on WeGenEditBench-text. Closed-source models are included for reference only and excluded from rank highlighting. Darker cell colors indicate better performance, with the three highlighted shades denoting the 1st, 2nd, and 3rd best results, respectively.}
\label{tab:wegen-text}
\end{table*}

\subsection{Qualitative Results}
To provide a qualitative assessment of WeLM-Image, we present representative results on both text-to-image generation and instruction-guided image editing. These examples cover diverse visual domains, artistic styles, and task complexities, offering insights into the model's ability to translate natural-language instructions into visual outputs. In addition to generating images with rich details and coherent compositions, WeLM-Image supports precise content manipulation and the coordinated execution of multiple editing instructions. The following results illustrate these capabilities across a variety of generation and editing scenarios.

\begin{figure*}[t]
    \centering
    \includegraphics[width=\textwidth]{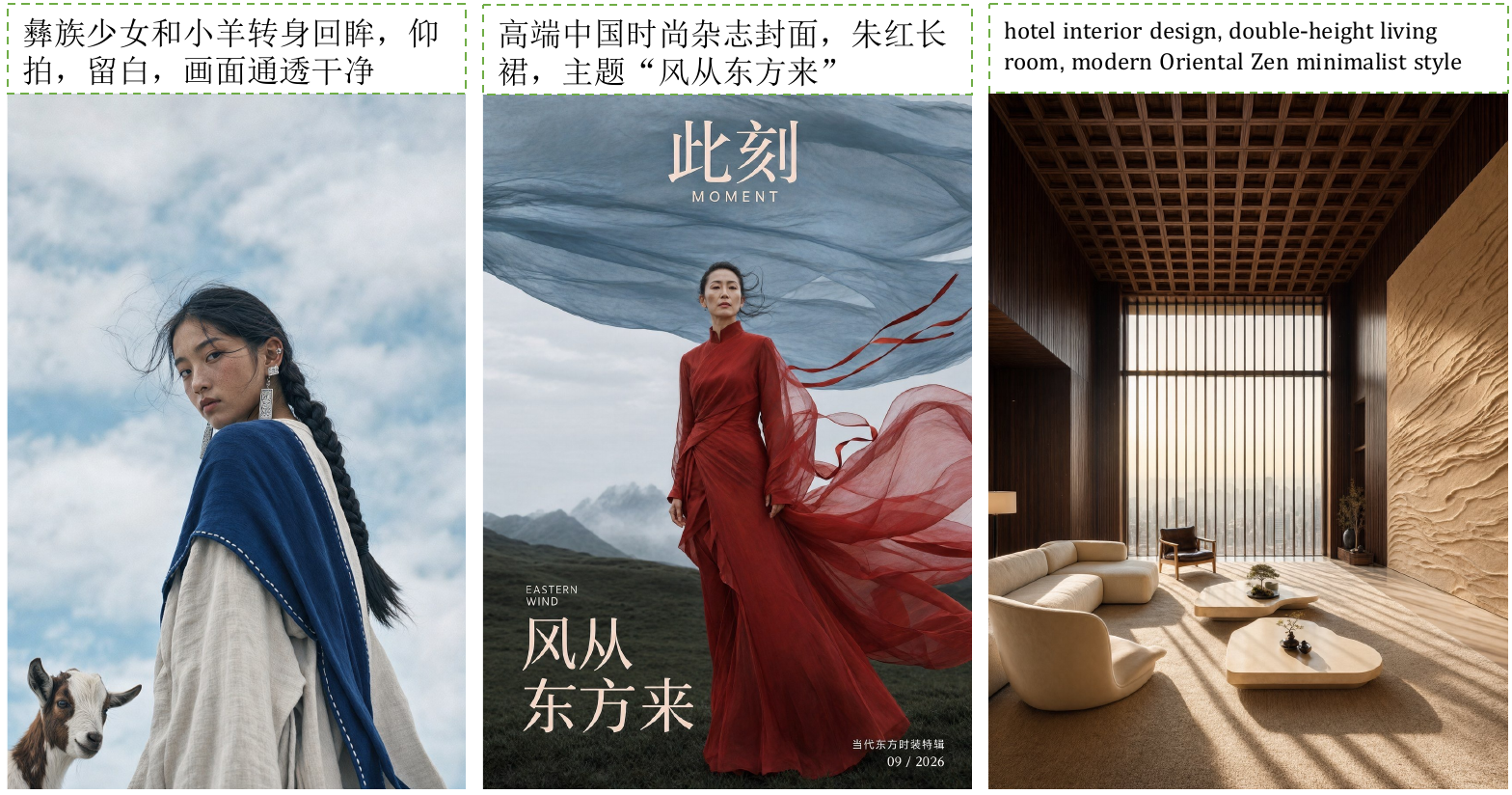}
    \includegraphics[width=\textwidth]{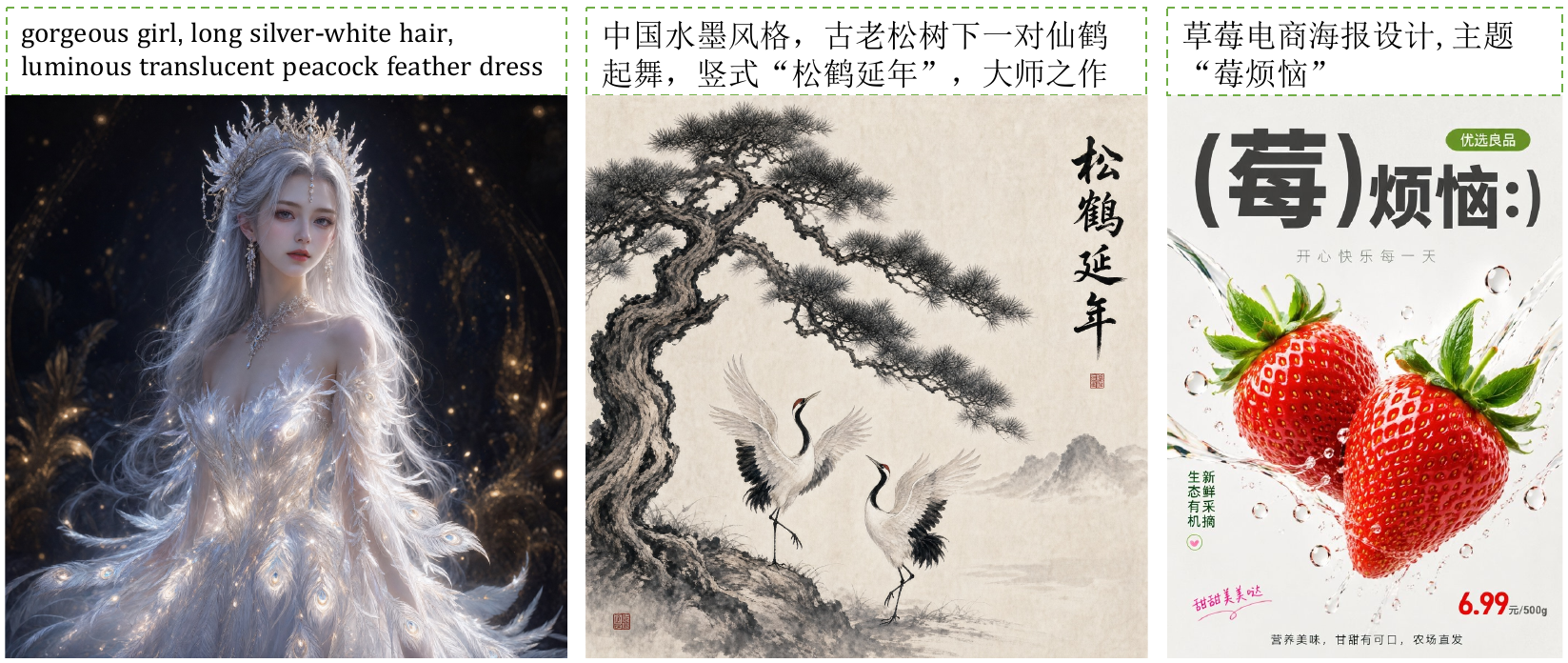}
    \includegraphics[width=\textwidth]{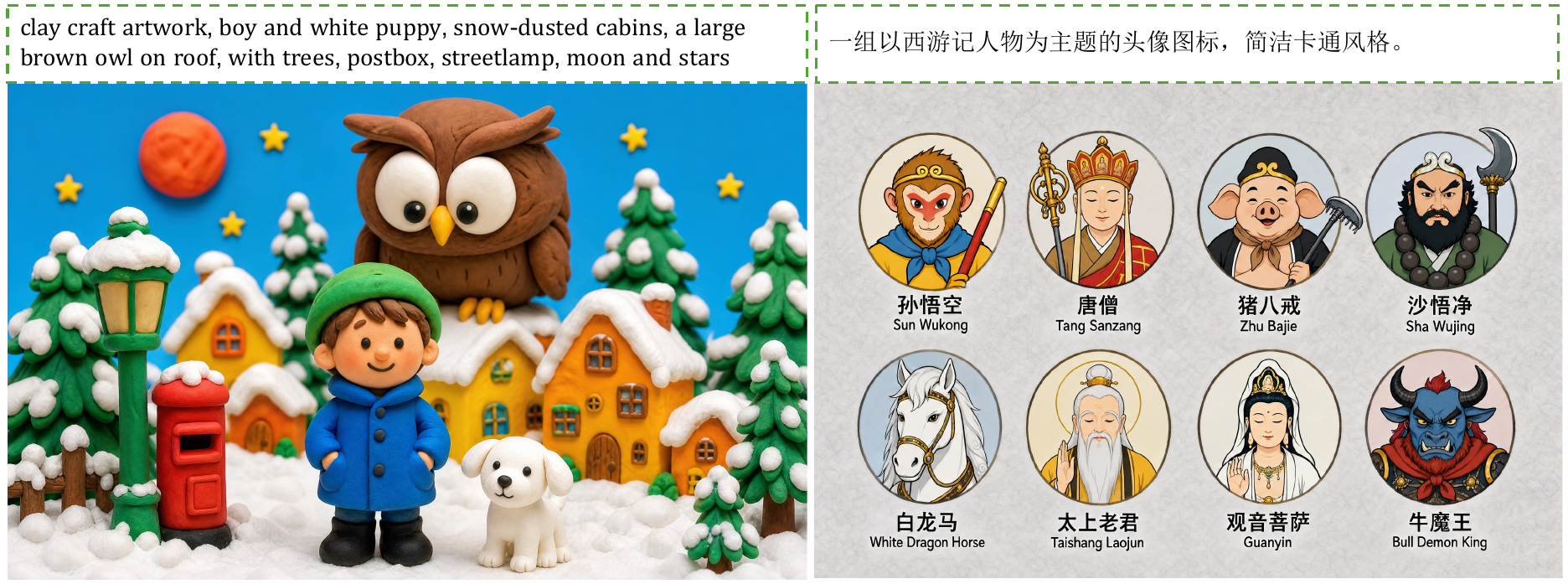}
    \caption{Qualitative results of \ModelName{} on text-to-image generation.}
    \label{fig:t2i_samples}
\end{figure*}

Figure~\ref{fig:t2i_samples} presents qualitative results of WeLM-Image on text-to-image generation across a diverse range of prompts and visual domains. The model demonstrates strong versatility in synthesizing images with distinct artistic styles, ranging from photorealistic portraits and cinematic fashion photography to traditional Chinese ink paintings, fantasy illustrations, 3D clay artwork, and graphic posters. Beyond stylistic diversity, \ModelName{} captures fine-grained visual attributes, including subject appearance, clothing details, scene composition, lighting, and background elements. It also handles text-rich image generation, producing posters and character illustrations that integrate textual content with visual elements in a coherent layout. These results illustrate the model's ability to translate diverse natural-language descriptions into visually rich images while maintaining consistency with the specified content, style, and composition.

\begin{figure*}[t]
    \centering
    \includegraphics[width=\textwidth]{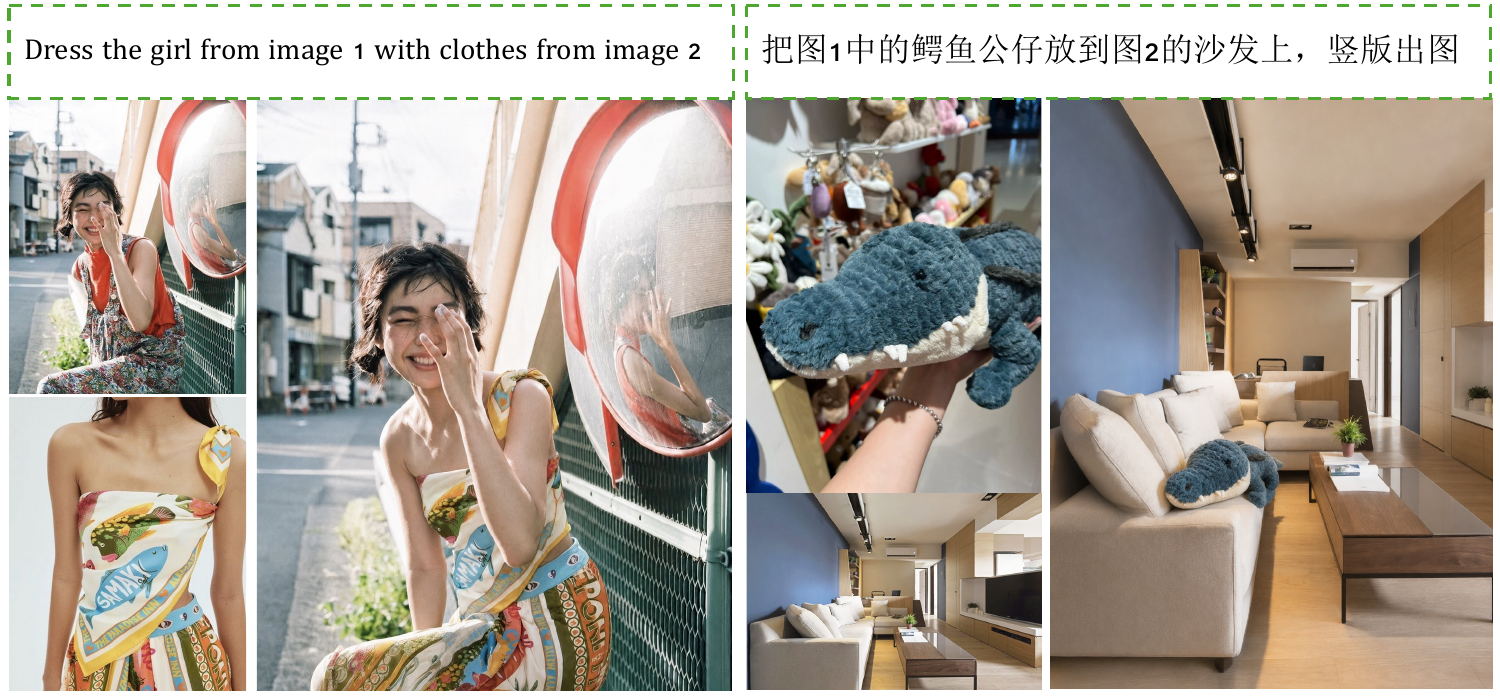}
    \includegraphics[width=\textwidth]{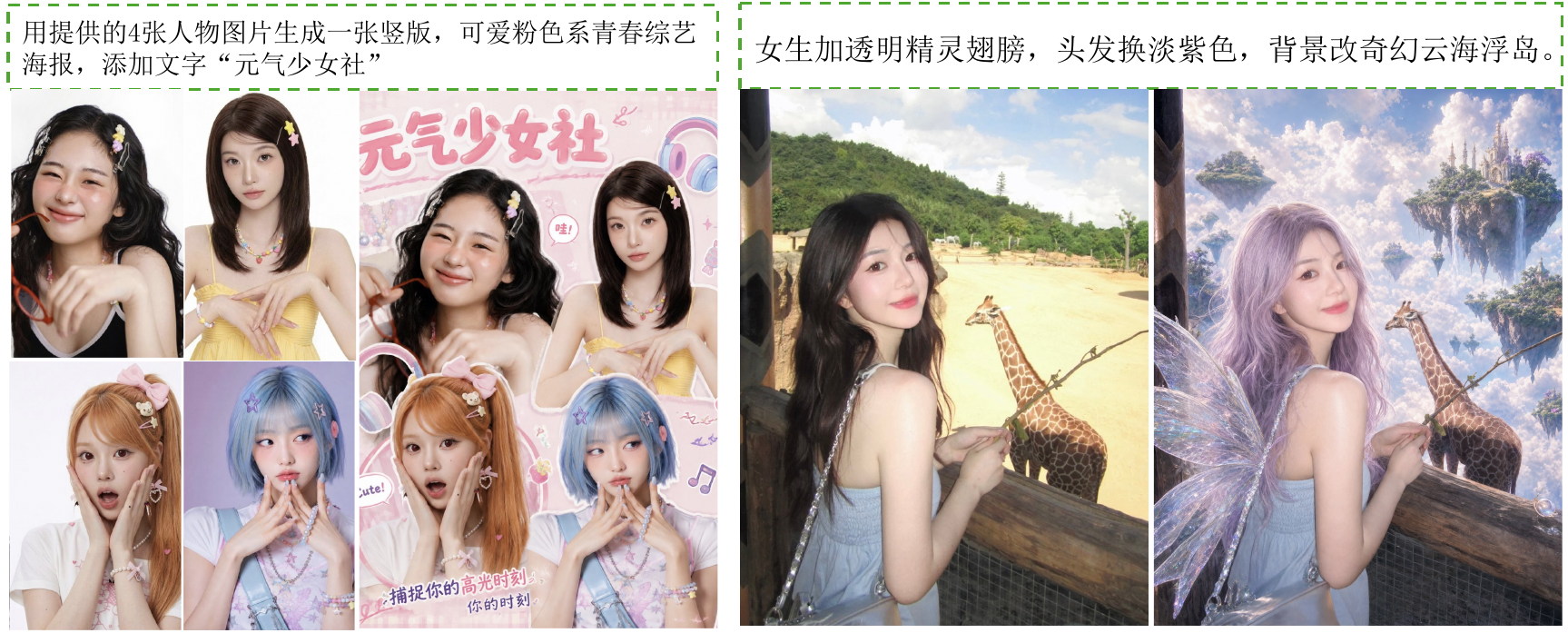}
    \includegraphics[width=\textwidth]{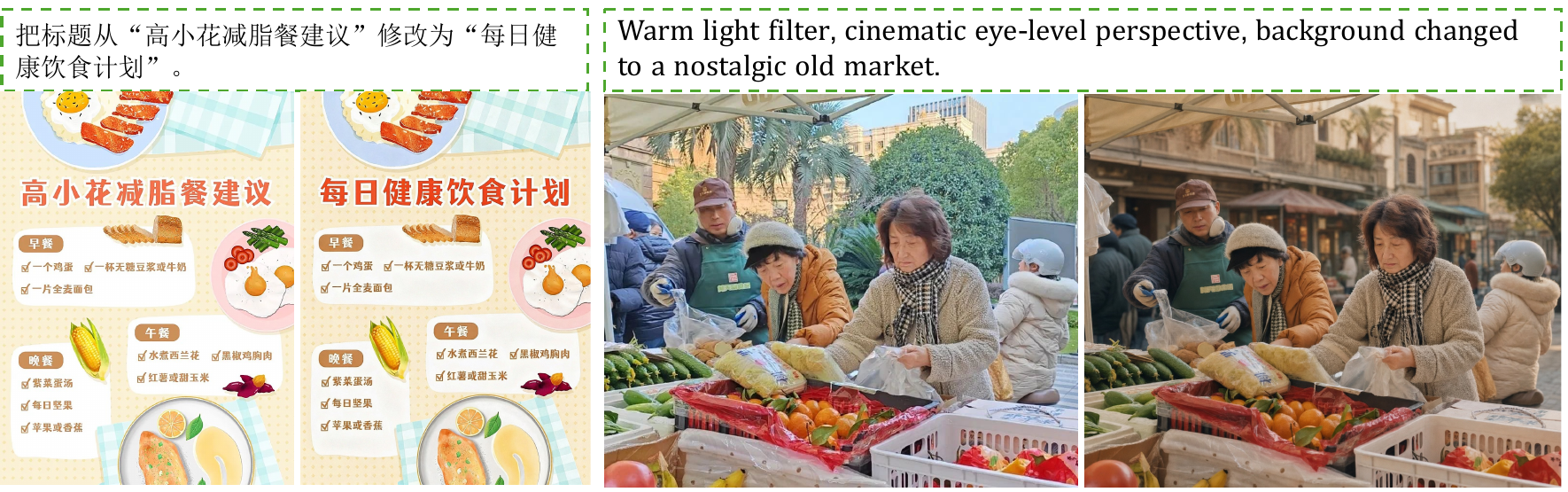}
    \caption{Qualitative results of \ModelName{} on image editing.}
    \label{fig:edit_samples}
\end{figure*}

Figure~\ref{fig:edit_samples} presents qualitative results of \ModelName{} on image editing, covering diverse editing operations and instruction complexities. The examples demonstrate the model's ability to perform a wide range of edits, including attribute transfer, object insertion, multi-image composition, background transformation, and text modification. Beyond individual editing operations, several cases highlight its ability to follow multiple instructions simultaneously, requiring coordinated changes to different image elements while respecting their respective constraints. For example, the model can integrate specified visual elements from multiple reference images, modify designated content, and preserve other aspects of the original image within a single editing process. Such tasks require not only accurate instruction interpretation but also effective coordination among multiple editing objectives. The results illustrate \ModelName{}'s ability to handle compositional editing instructions, incorporate reference-image information, and generate coherent outputs that reflect the requested changes while maintaining consistency with the intended content and structure.

\clearpage
\section{Conclusion}
\label{sec:conclusion}

We have presented \ModelName{}, a unified image foundation model that brings
multimodal understanding into text-to-image generation, instruction-guided
editing, and multi-reference synthesis. Its MoT--MoE backbone combines
specialized understanding and generation parameters with shared attention,
allowing textual instructions and visual references to guide synthesis
throughout the model. Sparse expert activation, a dedicated VAE, and
text-compatible spatial encoding provide the capacity and visual
representation needed for detailed, controllable image creation.

The data and training pipeline develops these capabilities together.
A knowledge-grounded visual vocabulary and large-scale clustering guide
semantic coverage, while bilingual captions, verified editing pairs, and
training-time feedback improve the relevance and quality of supervision.
Progressive pretraining and supervised fine-tuning extend the model to native
1.5K resolution. Task-specific reinforcement learning then refines generation
and editing, and noise-aware on-policy distillation brings their specialized
capabilities back into one checkpoint. This combines joint learning across
tasks with targeted optimization of their distinct quality requirements.

The reported evaluations demonstrate strong performance in bilingual
generation, text rendering, image editing, and multi-reference composition,
including leading overall results among the evaluated open-source baselines
on PRISM-Bench, Qwen-Image-Bench, and UnicBench. Few-step distillation extends
this model family to efficient deployment: \ModelNameTurbo{} supports
generation and editing with eight guidance-free steps and generates a
$1024\times1024$ image in 4.98\,s on a single NVIDIA H20 GPU.

% The model, data pipeline, and training methodology, spanning pretraining,
% reinforcement learning, and distillation, together provide a practical
% foundation for high-quality image creation in WeChat's consumer services.
% Further work will focus on stronger visual reasoning, more reliable control
% under complex instructions, and continued improvements in efficiency.

Together with these results, the model, data pipeline, and training
methodology provide a practical foundation for high-quality image creation
in WeChat's consumer services. Further work will focus on stronger visual
reasoning, more reliable control under complex instructions, and continued
improvements in efficiency.

\section{Authors}
The list is organized by contribution role, with individuals listed alphabetically by
their first name within each category. Authors marked with an asterisk (*) denote individuals who are no longer affiliated with our team.

\textbf{Core Contributors:} 

Canmiao Fu, Chen Li, Guohui Zhang, Hao Lu, Hongyang Wei, Huadong Wu, Huaisong Zhang, Hubery Yin\textsuperscript{*}, Liyuan Cui, Qixin Yan, Tong Shao, Xiaoqiang Zhou, Xinhang Leng, Ying Zhang, Zhenhai Liu, Zhipeng Huang.

\textbf{Contributors:} 

Can Ouyang, Chao Dai, Feng Wang, Fengbin Guan, Hao Liu, Hongrui Li, Jialian Li\textsuperscript{*}, Jikang Cheng\textsuperscript{*}, Jingyang Lin, Jinming Liu, Jun He, Junyu Wu, Lihao Ni, Pujing Yang, Qian Liang\textsuperscript{*}, Qian Wang\textsuperscript{*}, Shaobin Zhuang, Siyu Liu\textsuperscript{*}, Sujie Hu, Tianyi Zhang\textsuperscript{*}, Wei Liu, Wenhao Su, Xiaomin Li, Xiaotao Liu, Yazhen Xie, Zhigeng Xu, Zhonglei Chen.

\textbf{Acknowledgements: }

We would like to specially thank Hao Zhou, Sifan Liu, Leteng Weng and Jing Lyu for sponsoring this project. We would also like to thank
Chong Sun, 
Fangyuan Li, 
Feng Lin, 
Guohui Lin, 
Guomin Chen, 
Huayang Sun, 
Jiaxin Zhang, 
Jing Nie, 
Jingjing Li, 
Junyi Pan, 
Ken Deng\textsuperscript{*}, 
Lei Qian\textsuperscript{*}, 
Pang Yue, 
Piexuan Chen, 
Shaomian Yao, 
Tiantian Cao, 
Wenjing Wang, 
William Ming\textsuperscript{*}, 
Wujie Zheng, 
Xiaodan Zeng, 
Yinan Li\textsuperscript{*}, 
Yuanyuan Zhao, 
Yunhao Bai\textsuperscript{*}, 
Zaizhou Gong, 
Zhendong Yang\textsuperscript{*} and 
Ziyang Ma
for their valuable support, helpful discussions and contributions to this project.

%%\appendix
%%\input{sections/A_ablation_details}
%%\input{sections/B_reproducibility}
%%\input{sections/C_deployment_runbook}

\clearpage
\bibliographystyle{unsrt}
\bibliography{references}

\end{document}